%% file: ICS_Schrouff_arxiv.tex
\documentclass{article}

\usepackage[preprint,nonatbib]{neurips_2021}

\usepackage[utf8]{inputenc} 
\usepackage[T1]{fontenc}    
\usepackage{hyperref}       
\usepackage{url}            
\usepackage{booktabs}       
\usepackage{amsfonts}       
\usepackage{nicefrac}       
\usepackage{microtype}      
\usepackage{xcolor}         
\usepackage{amsmath}
\usepackage{graphicx}
\usepackage{cleveref}
\usepackage[export]{adjustbox}
\usepackage[font=small]{caption}
\usepackage[labelformat = empty,position=top]{subcaption} 

\title{Best of both worlds: local and global explanations with human-understandable concepts}

\author{%
  Jessica Schrouff\thanks{equal contribution} \\
  Google Research\\
  London, United Kingdom\\
  \texttt{schrouff@google.com} \\
   \And
   Sebastien Baur\footnotemark[1] \\
   Google Health \\
   London, UK \\
   \AND
   Shaobo Hou \\
   DeepMind \\
   London, UK \\
   \And
   Diana Mincu \\
   Google Research \\
   London, UK \\
   \And
   Eric Loreaux \\
   Google Health \\
   Palo Alto, CA, USA \\
    \And
   Ralph Blanes\thanks{Now at korgi.ai}\\
   Google Research \\
   Mountain View, CA, USA \\
   \And
   James Wexler \\
   Google Research \\
   Mountain View, CA, USA \\
   \And
   Alan Karthikesalingam \\
   Google Health \\
   London, UK \\
   \And
   Been Kim \\
   Google Research \\
   Mountain View, CA, USA \\
}

\begin{document}
\maketitle

\begin{abstract}
Interpretability techniques aim to provide the rationale behind a model's decision, typically by explaining either an individual prediction (local explanation, e.g. `why is this patient diagnosed with this condition') or a class of predictions (global explanation, e.g. `why is this set of patients diagnosed with this condition in general'). While there are many methods focused on either one, few frameworks can provide both local and global explanations in a consistent manner. In this work, we combine two powerful existing techniques, one local (Integrated Gradients, IG) and one global (Testing with Concept Activation Vectors), to provide \textit{local and global} concept-based explanations. We first sanity check our idea using two synthetic datasets with a known ground truth, and further demonstrate with a benchmark natural image dataset. We test our method with various concepts, target classes, model architectures and IG parameters (e.g. baselines). We show that our method improves global explanations over vanilla TCAV when compared to ground truth, and provides useful local insights. Finally, a user study demonstrates the usefulness of the method compared to no or global explanations only. We hope our work provides a step towards building bridges between many existing local and global methods to get the best of both worlds.
\end{abstract}

\section{Introduction}
\label{introduction}
Interpretability in machine learning (ML) has been deemed a key element for trustworthy models~\cite{Lipton2016, DoshiKim2017Interpretability}, and considered as a core requirement to deploy ML to high-stake domains such as healthcare or self-driving. When stakes are high, explanations are often expected at multiple levels: first at the global/population level to obtain a general understanding of a model's predictions, as well as at the local level (e.g., one patient of interest). Building inherently interpretable models (e.g. using attention~\cite{bahdanau2016neural}, including rule lists~\cite{Angelino2017}, or knowledge~\cite{Chen2019,Koh2020_CBM}) and post-hoc methods are widely explored in the field (e.g.~\cite{Simonyan2013,Ribeiro2016,Lundberg2017_SHAP,Ancona2018,Ancona2019}) to produce explanations. 

Among interpretability methods, post-hoc global or local methods have gained significant interest due to their convenience (see~\cite{Escalante2018} for a review). 
In particular, ``attribution methods'' is a family of methods that provide an importance score for each input feature. For vision applications, these scores are typically per pixel and can be displayed as a heat map, often referred to as an ``attribution map''. Naturally, the importance of a pixel is limited to one image (local), and cannot be simply ``averaged" to learn which pixels are important in general for many pictures in a class (global level).
It was however shown that practitioners are also keen to understand the ``overall reasoning" of the model~\cite{Tonekaboni2019}. How does the model reason to classify a set of patients to a degree of cancer? To answer this, a variety of \textit{global} explanation methods are investigated (e.g.~\cite{Wu2020,Hohman2020}). Despite the need for local and global methods, only few can provide both types of explanations~\cite{Lundberg19, Linden2019GlobalAO}. There are however advantages in having a single method for both, as local and global explanations can then be directly compared, allowing to highlight the specificities of each example compared to the average of the class in a consistent manner.

This work is an attempt to bridge the gap between local and global explanation techniques by combining two popular methods: Integrated Gradients (IG~\cite{Sundararajan2017}) and Testing with Concept Activation Vectors (TCAV~\cite{Kim2017}). IG offers input feature-wise importance based on a set of game-theory-inspired principles, while TCAV provides explanations using intuitive high-level concepts (e.g., red color, stripes). The proposed merging of the two results in a unique method that provides the best of both worlds: local and global explanations, with human-understandable concepts. Our contributions are:

\begin{itemize}
    \item We propose Integrated Conceptual Sensitivity (ICS) to provide local, concept-based explanations.
    \item We propose a set of IG baselines, specific to our formulation.
    \item We derive a closed form solution in simple scenarios.
    \item We empirically show that our formulation leads to more faithful global explanations compared to vanilla TCAV.
    \item We validate our method on two synthetic datasets as well as a natural image dataset, and test our results across multiple baselines, model architectures, and concepts.
    \item We perform a user study that demonstrates the usefulness of the method.
\end{itemize}

\section{Related work and methods}
\label{sec:related_works}
In this section, we take a deep dive on the two methods that we propose to combine: IG and TCAV and describe their limitations before presenting our method. 

\subsection{Notation and setup} 
Let our dataset be the junction of a set of $n$ samples with $d$ features $\mathcal{X} \in \mathbb{R}^{n \times d}$ and associated labels $\mathbf{y} \in \{0,1\}^{n}$. In the multiclass case, $\mathcal{X}_k$ represents the set of inputs whose label is $k$, with $k \in {1, \dots, K}$. We consider a trained neural network $F : \mathcal{X} \rightarrow \{1, ..., K\}$ with $L$ layers. For a given layer $1\leq l \leq L$ and a given class $1 \leq k \leq K$, we can write the $k$-th output of $F$ as $F_k(\mathbf{x}) := h_k(f_l(\mathbf{x}))$ where $f_l(\mathbf{x})$ is the activation vector at the $l$-th layer, which we refer to as $\mathbf{a}_l$. We also define $\mathbf{W}_l \in \mathbb{R}^{d_{l+1}\times d_l}$ to represent the weights of the $l$-th layer, and $b_l \in \mathbb{R}^{d_{l+1}}$ as its bias. We drop the layer index $l$ in the remaining of the paper for compactness. 

\subsection{Gradient-based attributions and integrated gradients}
Of interest in our work is a family of saliency map techniques~\cite{Simonyan2013} that typically use the gradients w.r.t. input to provide a score per input feature. Recent developments in gradient-based techniques include e.g. gradients $\times$ inputs ~\cite{Shrikumar2016}, LRP~\cite{Bach2015}, DeepLIFT~\cite{Shrikumar2017}, smoothGrad~\cite{Smilkov2017-vc}, or integrated gradients~\cite{Sundararajan2017}. In vision applications, these methods give a score to each pixel in one image, which makes them inherently local. While criticism of these methods exists~\cite{jain2019attention, sixt2019explanations, Adebayo2018}, they are considered in real-world applications~\cite{Sayres2019}. 

In particular, the integrated gradients (IG) technique~\cite{Sundararajan2017} computes a path integral between an uninformative input (the baseline) $\mathbf{x'}$ and the observed input $\mathbf{x}$ (Eq.\ref{eq:ig}). This technique verifies completeness (i.e. attributions sum to the difference in predictions ~\cite{Sundararajan2017,Ancona2018}), Sensitivity(a) (i.e. relevant variables obtain a non-null attribution score) and Sensitivity(b) (i.e. irrelevant variables obtain a null attribution score). Formally, IG is defined as
\begin{equation}
    \mathrm{IG}_i^k(\mathbf{x},\mathbf{x'}) := (x_i-x'_i) \int_0^1 \nabla_i F_k(\mathbf{x'}+\alpha(\mathbf{x-x'}))d\alpha
    \label{eq:ig}
\end{equation}
Where $\nabla$ is the gradient operator and $\nabla_i$ is its projection on the $i$-th dimension, $i \in {1,\dots,d}$. $\nabla f(\mathbf{x})$ designates the gradient of $f$ evaluated at $\mathbf{x}$. One of the critical decision in using IG is deciding which baseline to use. Conceptually, the baselines represent ``no information'', and the original paper used a simple baseline (e.g., black/white images). However it was shown that the method is sensitive to the choice of baselines~\cite{Ancona2018,Sturmfels2020,Goh2020}. In our work, we propose and test many baselines across model architectures and datasets.


\subsection{Explanations with human-understandable concepts}
Basing attributions or feature representations on human-understandable concepts has seen recent interest. Concepts can be defined manually by associations (e.g., \cite{Lee2019}), based on existing ontology (e.g., \cite{Panigutti2020}) or through (labelled) examples \cite{bau2017network,Kim2017,Clough2019,Graziani2018,Mincu2021}. Concept scores can then be computed per single neuron, as in network dissection \cite{bau2017network} or per layer \cite{Kim2017}. As we are more interested in providing local and global scores to explain the \emph{prediction} of a model compared to the behavior of single neurons, our work focuses on TCAV~\cite{Kim2017}.

\subsection{Testing with Concept Activation Vectors}
\begin{figure}[!t]
    \centering
    \includegraphics[width=0.8\textwidth]{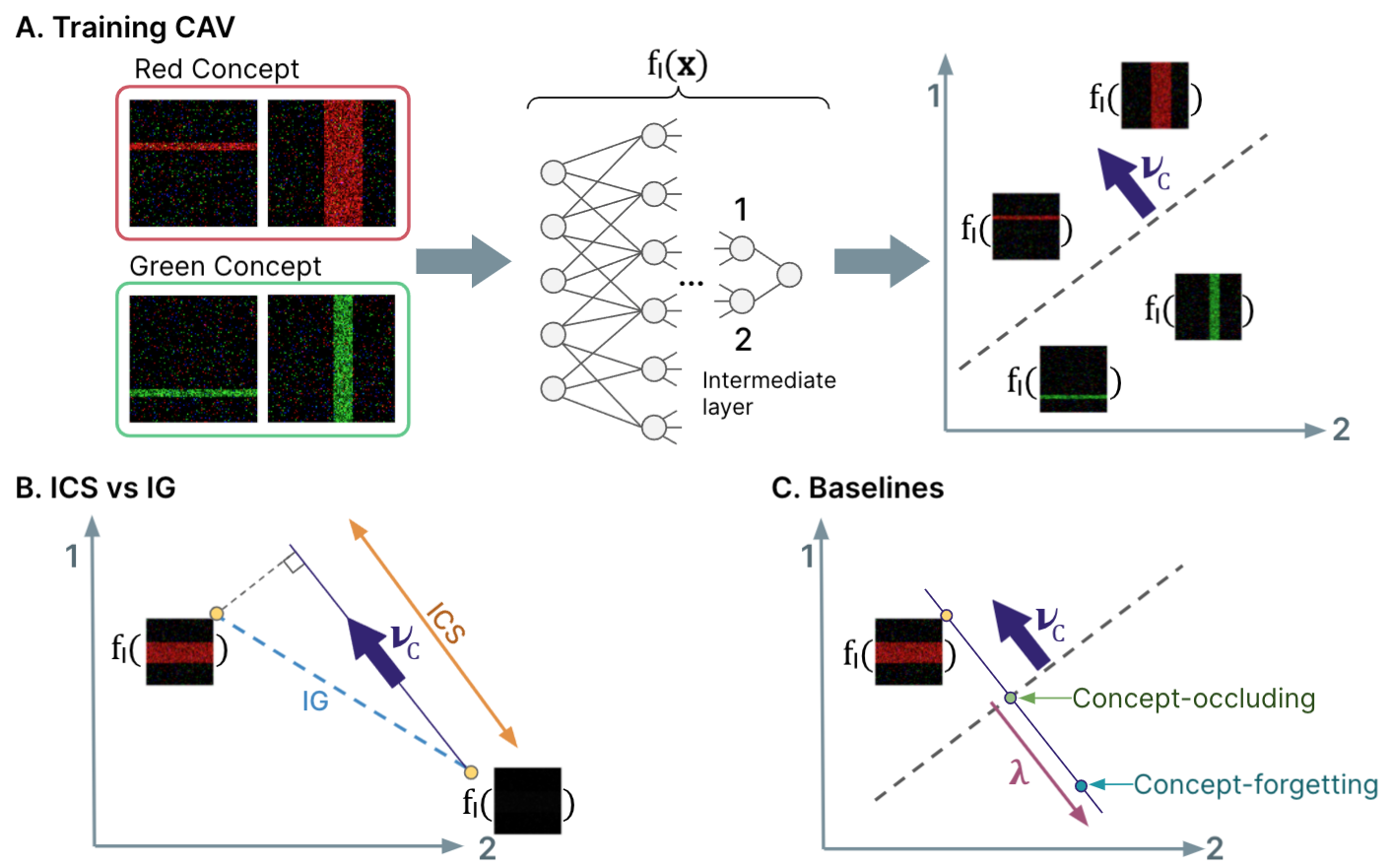}
    \caption{\textbf{Illustration of TCAV and ICS. A} Images of red (concept) and green (control) bars and their activations $f_l$ in a network. A linear classifier (top right) is trained to distinguish concept activations from control images, generating a CAV ($v_C$). \textbf{B} ICS is the projection of integrated gradients (IG) between a baseline (e.g., black image) and a new sample (red horizontal bar) on the direction of the CAV, $v_C$. \textbf{C} Concept-forgetting baselines remove information in the direction of the concept, with a strength $\lambda$. `Concept-occluding' is a particular case of this baseline that projects the activations on the linear model hyperplane.}
    \label{fig:TCAV_illustration}
\end{figure}

TCAV~\cite{Kim2017} uses \textit{concepts} ($C$) instead of features to provide explanations, where concepts are defined as understandable and identifiable by humans (e.g. ``stripes'' or ``pointy ears''). TCAV has been successfully applied to different domains e.g. in healthcare~\cite{Clough2019, Graziani2018, Mincu2021}. Technically, TCAV assumes that users have a set of samples for a concept of interest. To express this concept, they find a Concept Activation Vector (CAV) in a network's activations space $\mathbf{a} \in \mathbb{R}^{d}$ (a layer with $d$ dimension) that points from any direction to these `concept data samples'. Formally, the CAV is obtained by training a linear classifier to distinguish concept activations from random activations, and taking the unit-norm vector $\mathbf{v_C}$ orthogonal to its decision boundary. Intuitively, a CAV represents the direction of the selected concept as encoded in the network. Figure~\ref{fig:TCAV_illustration}a illustrates the CAV building for a red/green concept in a 2-dimensional layer.

To understand the influence of a concept on the model's predictions, Conceptual Sensitivity (CS, Eq.\ref{eq:cs}) is defined as the directional derivative of one of the network's outputs $k$ w.r.t. the CAV for concept $C$: 
\begin{equation}
    \mathrm{CS}_C^k(F, \mathbf{x}) := \frac{\partial h_k(f(\mathbf{x}))}{\partial \mathbf{v_C}} = \nabla h_k(f(\mathbf{x}))^T\mathbf{v_C}
    \label{eq:cs}
\end{equation}

While the raw value of CS can be seen as a local explanation, note that the issue of gradient saturation also happens here as seen in~\cite{Sundararajan2017}: a high directional derivative only means that the gradient aligns strongly with the CAV but does not quantify the role this direction played in the model’s decision. In this sense, the increase in prediction by perturbing in the direction of the concept could be either very small (e.g. an increase from 0.9 to 0.92) or very large (e.g. an increase from 0.1 to 0.9).

CS can however provide global attributions by aggregating the scores over multiple examples to obtain a \textit{TCAV} score between 0 and 1. Since the scale of directional derivative values is a priori unknown (i.e., CS is not a normalized quantity), TCAV circumvents this difficulty by defining the TCAV score as the average \textit{sign} of CS values on a given dataset. 

An important step in TCAV is ensuring that CAVs did not accidentally return a high TCAV score. To do this, the method suggests building many bootstrapped versions of the CAVs and run t-testing between TCAV scores from concept CAVs v.s. random CAVs (CAVs trained with random images). We also discuss how we adapt this test in our approach.

\subsection{TCAV alone overestimates concept importance}
\begin{figure}[!t]
\centering
\begin{subfigure}[t]{0.03\textwidth}
\textbf{a}
\end{subfigure}
\begin{subfigure}[t]{0.4\textwidth}
\includegraphics[width=\linewidth,valign=t]{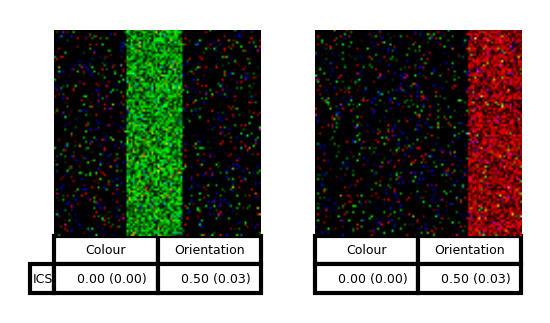} 
\end{subfigure}
\begin{subfigure}[t]{0.03\textwidth}
\textbf{b}
\end{subfigure}
\begin{subfigure}[t]{0.48\textwidth}
\includegraphics[width=\linewidth,valign=t]{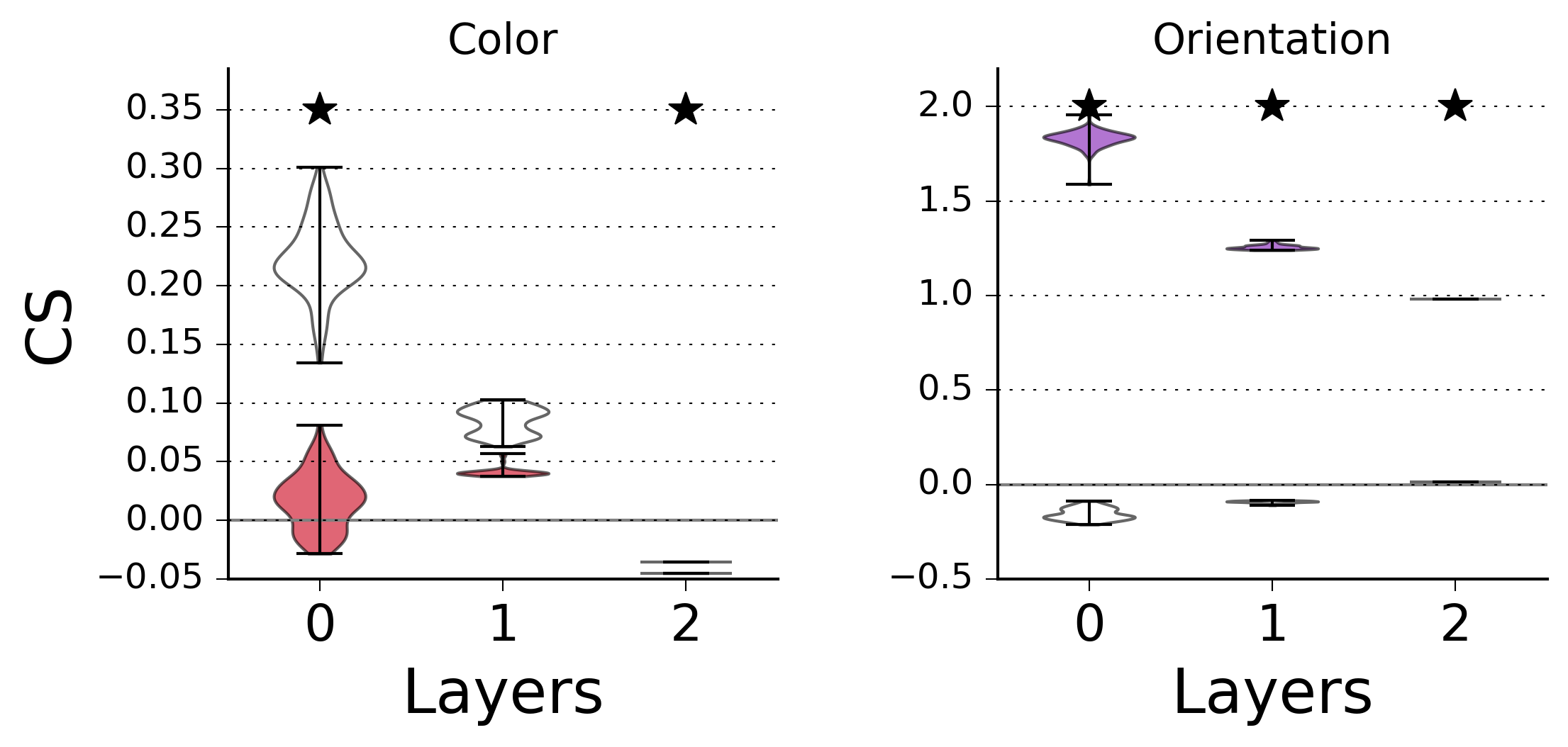}
\end{subfigure}\\
\begin{subfigure}[t]{0.03\textwidth}
\textbf{}
\end{subfigure}
\begin{subfigure}[t]{0.4\textwidth}
\includegraphics[width=\linewidth,valign=t]{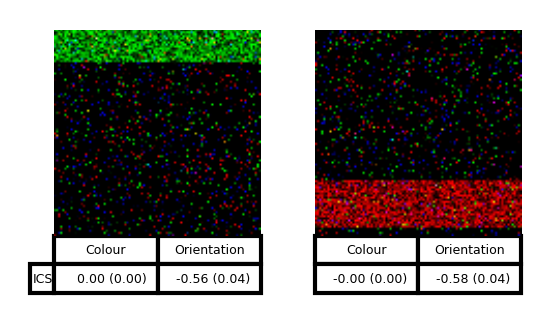}
\end{subfigure}
\begin{subfigure}[t]{0.03\textwidth}
\textbf{c}
\end{subfigure}
\begin{subfigure}[t]{0.48\textwidth}
\includegraphics[width=\linewidth,valign=t]{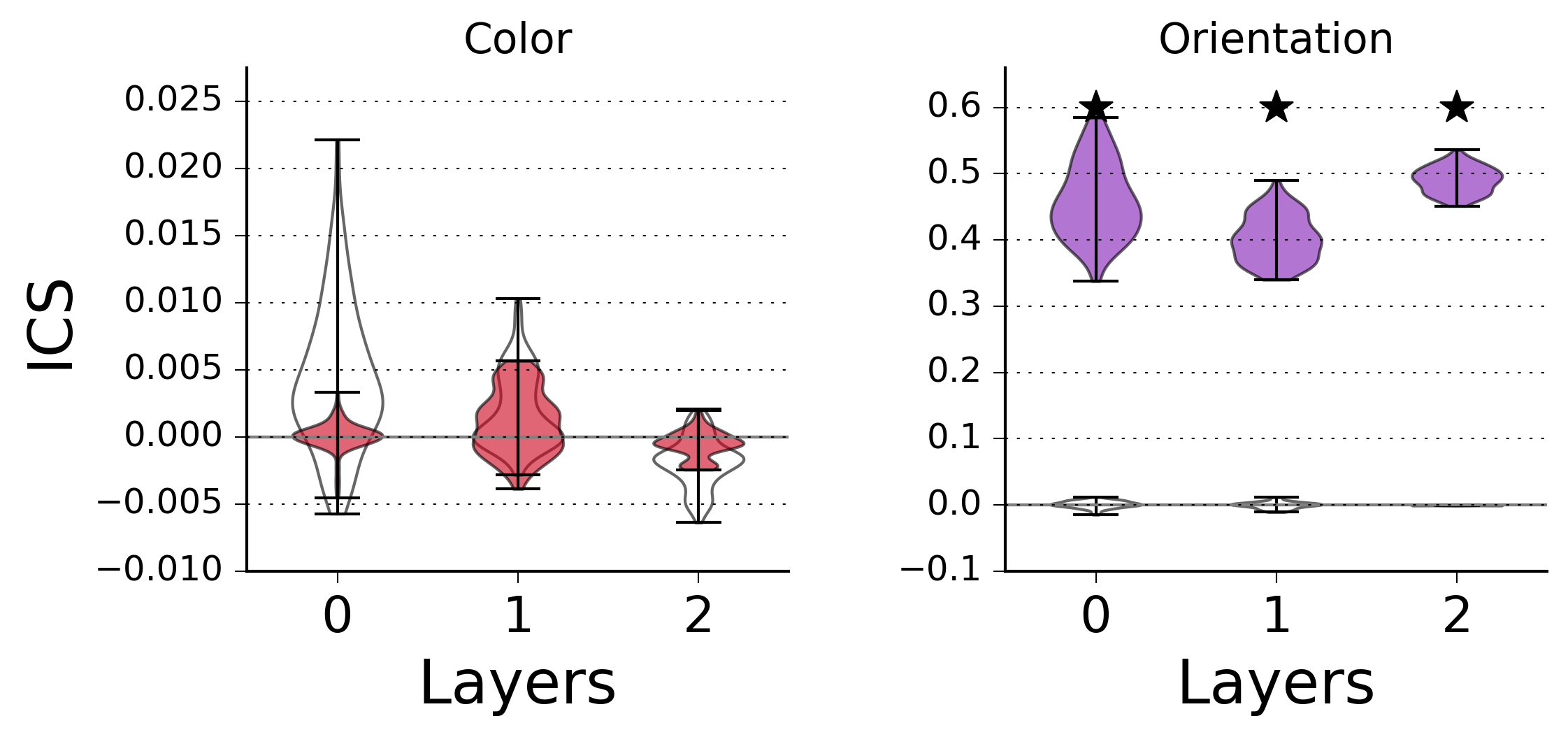}
\end{subfigure}
\caption{\textbf{CS leads to significant scores for irrelevant concepts, while ICS does not. a} Four representative local examples, displaying associated ICS for all concepts, at layer 2 of the model predicting bar orientation $F_o$. \textbf{b} Distribution of CS for target 1 of the orientation model $F_o$, for the color (displayed in red) and orientation (displayed in purple) concepts, for each layer of the model. Null distributions from permuted CAVs are displayed as white violins and significant TCAV$^{sign(CS)}$ scores are highlighted by a star (Bonferroni corrected for layers and concepts). \textbf{c} Distribution for ICS (computed using a black image baseline).}
\label{fig:BARS}
\end{figure}

We have however observed that TCAV scores can be significant even when the concept is not used in the model's predictions. We illustrate this failure mode using the synthetic dataset used in \cite{Goyal2019} (referred to as ``BARS''). This dataset contains noisy black images with vertical and horizontal lines that are either green or red (Figure~\ref{fig:BARS}\textbf{a}). Suppose the orientation of the bar defines the label $y$, while the color is irrelevant (i.e. the color is independent of the orientation). A 3-layered MLP with ReLU activations and 50\% dropout after each fully-connected layer predicts the orientation of the bar ($F_o$, with `o' referring to `orientation') with 99.97\% accuracy in train and 100\% in test time. While this model does not rely on the color of the bar to make a prediction, both the ``orientation'' and ``color'' concepts seem to be encoded in the network's activations and significant CAVs can be built for both concepts (see Supplement~\ref{app:cavs_synth} for details). These represent the direction from red to green (color) and from horizontal to vertical (orientation). We observe that the orientation concept has a positive influence on the predictions in all layers, i.e. CS scores are positive (Figure~\ref{fig:BARS}b). The color concept is also displaying a positive influence for layers 0 and 1 and a negative influence for layer 2. Despite smaller magnitudes of CS for the color concept, both concepts are considered as ``significant'' under the mechanism proposed in TCAV (2 out of 3 layers for color). In this simple controlled setting, we hence display that the aggregation technique is not \textit{sensitive} and can be misleading. A similar phenomenon was observed in \cite{Goyal2019}.

\subsection{Our method: Integrated Conceptual Sensitivity (ICS)}
\label{sec:ics_methods}
We introduce Integrated Conceptual Sensitivity (ICS, Eq.\ref{eq:ics}) as a \textit{local, concept-based attribution technique}, which can also be aggregated into global explanations. For a concept $C$ with unit norm CAV, $\mathbf{v_C}$, we define ICS as follows:
\begin{equation}
    \mathrm{ICS}_C^k(\mathbf{a},\mathbf{a'}) := (f(\mathbf{x})-\mathbf{a'})^T \mathbf{v_C} \int_{[\mathbf{a'}, f(\mathbf{x})]} \nabla_{\mathbf{v_C}}h_k(\mathbf{a}) d\mathbf{a}
    \label{eq:ics}
\end{equation} 
Where $\mathbf{a'}\in \mathbb{R}^{d}$ represents the baseline's activation at some layer and the support of the integral $[\mathbf{a'}, f(\mathbf{x})]$ corresponds to $\{\mathbf{a'} + \alpha(f(\mathbf{x}) - \mathbf{a'}), \alpha \in [0,1] \}$. Intuitively, ICS (Figure~\ref{fig:TCAV_illustration}b) represents how much of the change in predicted probability is explained by the change in the direction of the concept $\mathbf{v_C}$, compared to a (typically uninformative) baseline. While ICS is the projection of the integrated gradients (a bounded quantity) onto the concept, there is no guarantee that this projection is itself bounded\footnote{Despite this fact, we rarely observe values for ICS that are larger than 1 in our experiments, see Supplement~\ref{app:BARS_baseline}.}, meaning that ICS does not satisfy completeness \cite{Sundararajan2017}.
Global concept attributions can then be obtained by averaging ICS over multiple examples \cite{Van_der_Linden2019-vf}. Note that, contrary to the original formulation of TCAV score that averages the signs of $CS$ (referred as $\mathrm{TCAV}^{sign(\mathrm{CS})}$), the ICS scores are aggregated using their average:
$\mathrm{TCAV}^\mathrm{ICS} = \frac{1}{n}\sum_{i=0}^n ICS_{i,C}^k$. We however note that other aggregation techniques could be considered, e.g. based on ranking \cite{Ibrahim2019-hh,Van_der_Linden2019-vf}.

\textbf{Choice of baselines and closed form solutions:} As discussed in \cite{Sundararajan2017,Ancona2018,Kapishnikov_2019_ICCV,Goh2020,Sturmfels2020}, the choice of the baseline affects the obtained explanations. We test with `uninformative' baselines (e.g., black and white image, maximum entropy leading to neutral predictions) and propose `informative' baselines for concepts:
\begin{itemize}
 \item \textit{Concept-forgetting baseline}: $\mathbf{a'} = \mathbf{a} - \lambda\mathbf{v_C}$ for some $\lambda \in \mathbb{R}^*+$, which removes a certain amount of concept-aligned information.
    \item \textit{Concept-occluding baseline}: $\mathbf{a'} = \mathbf{a} - (\mathbf{a}^T\mathbf{v_C}+b)\frac{\mathbf{v_C}}{||\mathbf{v_C}||_2^2}$: the baseline with the concept removed, where $b$ is the bias of the linear model. This particular case of concept-forgetting can be related to the occlusion~\cite{Zeiler2014} of the concept.
\end{itemize}
Naturally, the best baseline for each application will be different. In Section~\ref{sec:results}, we show this variation for different datasets. With entropy-maximizing and concept-forgetting baseline, we can derive analytical formulations of ICS, removing the computational expense related to the integral (see Supplement~\ref{app:closed_form} for derivation):

\textbf{Analytical formulation 1:} For the last layer of a binary classification model with entropy-maximizing baseline (i.e. $h(\mathbf{a'}) = 0.5$), ICS can be written as (Eq.\ref{eq:ics_last_layer_main}):
\begin{equation}
    \mathrm{ICS}_{C}(\mathbf{a},\mathbf{a'}) = \bigg(\frac{\mathbf{v_C}^T\mathbf{w}}{||\mathbf{w}||_2}\bigg)^2\bigg(\sigma(\mathbf{w}^T\mathbf{a}+b) - 0.5\bigg)
\label{eq:ics_last_layer_main}
\end{equation}

\textbf{Analytical formulation 2:} For a multi-class model, with concept-forgetting baseline, we can derive ICS as (Eq.\ref{eq:ics_multi_main}):
\begin{equation}
 \mathrm{ICS}_C^k(\mathbf{a}, \mathbf{a'}) = h_k(\mathbf{a}) - h_k(\mathbf{a'})   
 \label{eq:ics_multi_main}
\end{equation}

\textbf{Statistical testing for TCAV:} \label{sec:statistical_testing}
An important part of TCAV is to confirm the statistical significance of the CAV. We propose a nonparametric permutation test to assess the significance of the out-of-sample classification performance of the CAV, which can be seen as a generalized version of the approach proposed in~\cite{Kim2017}\footnote{Please note a slight inconsistency between the statistical testing in \cite{Kim2017} and the corresponding code at \url{https://github.com/tensorflow/tcav}. After consulting with the authors, our work refers to the code, which tests TCAV scores against scores obtained from `random' CAVs instead of a fixed value of 0.5}. Intuitively, this test estimates whether a concept has been ``significantly'' encoded in each layer and prevents testing with non-significant directions. Concretely, we train CAVs on bootstrapped datasets (n=100 resamples). For each bootstrap sample, we build CAVs from the same data but with permuted `positive' and `negative' concept labels ($n_{perm}=10$). We then compute the number of permuted CAVs leading to higher or similar performance compared to the performance of the non-permuted CAVs and assess a CAV as significant when this proportion is $p<0.05$. Note that when testing multiple concepts, a correction method (e.g., Bonferroni, false discovery rate) should be added. The permuted CAVs can also be used to assess the significance of the scores by comparing the distribution of CS (resp. ICS) across bootstrap samples to the distribution of CS (resp. ICS) as estimated from the permuted CAVs. This test can be performed both at the global and local level. 

\section{Results}
\label{sec:results}

To validate our approach, we compare global ICS results to the original formulation of TCAV on two synthetic datasets with ground truth in Section~\ref{sec:synthetic} and illustrate local explanations using ICS on a benchmark imaging dataset in Section~\ref{sec:results_imagenet}. We test our results across different models with varying complexity.

\subsection{Quantitative evaluation on synthetic datasets}\label{sec:synthetic}
One of the challenges in interpretability methods is to show the ``faithfulness'' of an explanation. To validate our approach, we use a simple synthetic dataset along with semi-natural images that are synthetically generated. With carefully crafted distributions of this data, we can train models with known ground truth of what the explanation ``should be''. We describe the metrics and datasets used to quantitatively illustrate the performance of $\mathrm{TCAV}^\mathrm{ICS}$ over $\mathrm{TCAV}^{sign(\mathrm{CS})}$.

\subsubsection{Datasets and models}

\textbf{BARS:} This dataset from~\cite{Goyal2019} features noisy 100x100 images of red and green, horizontal and vertical bars. We introduce two models (3-layered ReLU MLP with dropout): $F_o^\mathrm{BARS}$ which is trained to predict the orientation of the bars, and $F_c^\mathrm{BARS}$ which is trained to predict their color.

\textbf{BAM\footnote{\url{https://github.com/google-research-datasets/bam}, Apache 2 license}:} This dataset from \cite{Yang2019} features random objects from MSCOCO \cite{Lin2014} pasted onto random scenes from MiniPlaces \cite{Zhou2018}. There is a total of 10 object classes and 10 scene classes. The training dataset contains 90k images, and labels are carefully balanced in a way that prevents confounding. As per \cite{Yang2019}, we introduce two models: $F_o^\mathrm{BAM}$ (resp. $F_s^\mathrm{BAM}$) is trained to identify the object displayed in the image (resp. the background scene). We report results using two state-of-the-art model architectures: EfficientNet-B3 \cite{Tan2020} and ResNet50 \cite{He2015}. We used publicly available \footnote{\url{www.tensorflow.org}, Apache 2 license} models~\cite{tfmodels2020github} that were pre-trained on ImageNet and fine-tuned them on BAM. We use analytical solutions from Eq.~\ref{eq:ics_multi_main} and Eq.~\ref{eq:ics_last_layer_main} to debug and compute ICS whenever possible.\\

\subsubsection{Metrics}

In~\cite{Yang2019}, the authors define the Model Contrast Score (MCS, Eq.~\ref{eq:mcs}) to evaluate global attribution methods. More specifically, MCS contrasts the attributions obtained for $C$ in two models: model $F_1$ that has concept $C$ as one of its targets and model $F_2$ for which concept $C$ is irrelevant. MCS can be thought of as the contrast between Sensitivity(a) and Sensitivity(b) as defined in \cite{Sundararajan2017}, at the global level for concepts. Hence, larger MCS means a more sensitive attribution method. For TCAV, MCS is written as: 
\begin{equation}
    \mathrm{MCS}_{C} := \mathrm{TCAV}_{C, C}(F_1, \mathcal{X}_\mathrm{TP}(F_1)) - \mathrm{max}_k \mathrm{TCAV}_{k, C}(F_2, \mathcal{X}_\mathrm{TP}(F_2))
    \label{eq:mcs}
\end{equation}
where $\mathrm{TCAV}_{k, C}(F, \mathcal{X}_\mathrm{TP}(F))$ is the TCAV score for concept $C$ when predicting target $k$ with model $F$ based on $\mathcal{X}_\mathrm{TP}(F)$. $\mathcal{X}_\mathrm{TP}(F) \subset \mathcal{X}$ is the set of true positives for model $F$. Conditioning on this subset allows to disentangle the errors of the model from those of the attribution technique. For the ``orientation'' concept in BARS, we compute MCS by contrasting the TCAV scores obtained from $F_o$ (orientation model, where the concept is relevant) and from $F_c$ (color model, where the concept is irrelevant). We then perform the same computation for the “color” concept by reversing the contrast. For BAM, we contrast the object and scene models for the object concepts (i.e. $F_1=F_o$ and $F_2=F_s$), and conversely for the scene concepts, reporting each concept separately. For $\mathrm{TCAV}^{sign(\mathrm{CS})}$, the ideal value of MCS is 0.5: for instance, the TCAV score of the ``orientation'' concept should be 1 for $F_o$, but 0.5 for $F_c$ as it is irrelevant for this model. As $\mathrm{TCAV}^{sign(\mathrm{CS})}$ is not bounded in $[0 ,1]$, MCS values should be positive, and as large as possible. MCS is computed for each CAV separately, hence providing a distribution based on bootstrapped directions (see Supplement~\ref{app:bootstrap_mcs} for the detailed procedure).

\subsubsection{Local explanations with ICS}

Using BARS, we showcase the effectiveness of the local explanation aspect of ICS.
Figure~\ref{fig:BARS}a illustrates this on 4 local examples of the BARS dataset for $F_o$. ICS correctly indicates that the color of the bars does not affect its predictions (score close to 0.), while  
the orientation concept (which determines the label) are close to 50\% for vertical bars and -50\% for horizontal bars, reflecting the importance of the presence (vertical) or absence (horizontal) of the concept.

\begin{figure}[!t]
\centering
\begin{subfigure}[t]{0.03\textwidth}
\textbf{a}
\end{subfigure}
\begin{subfigure}[t]{0.42\textwidth}
\includegraphics[width=0.9\linewidth,valign=t]{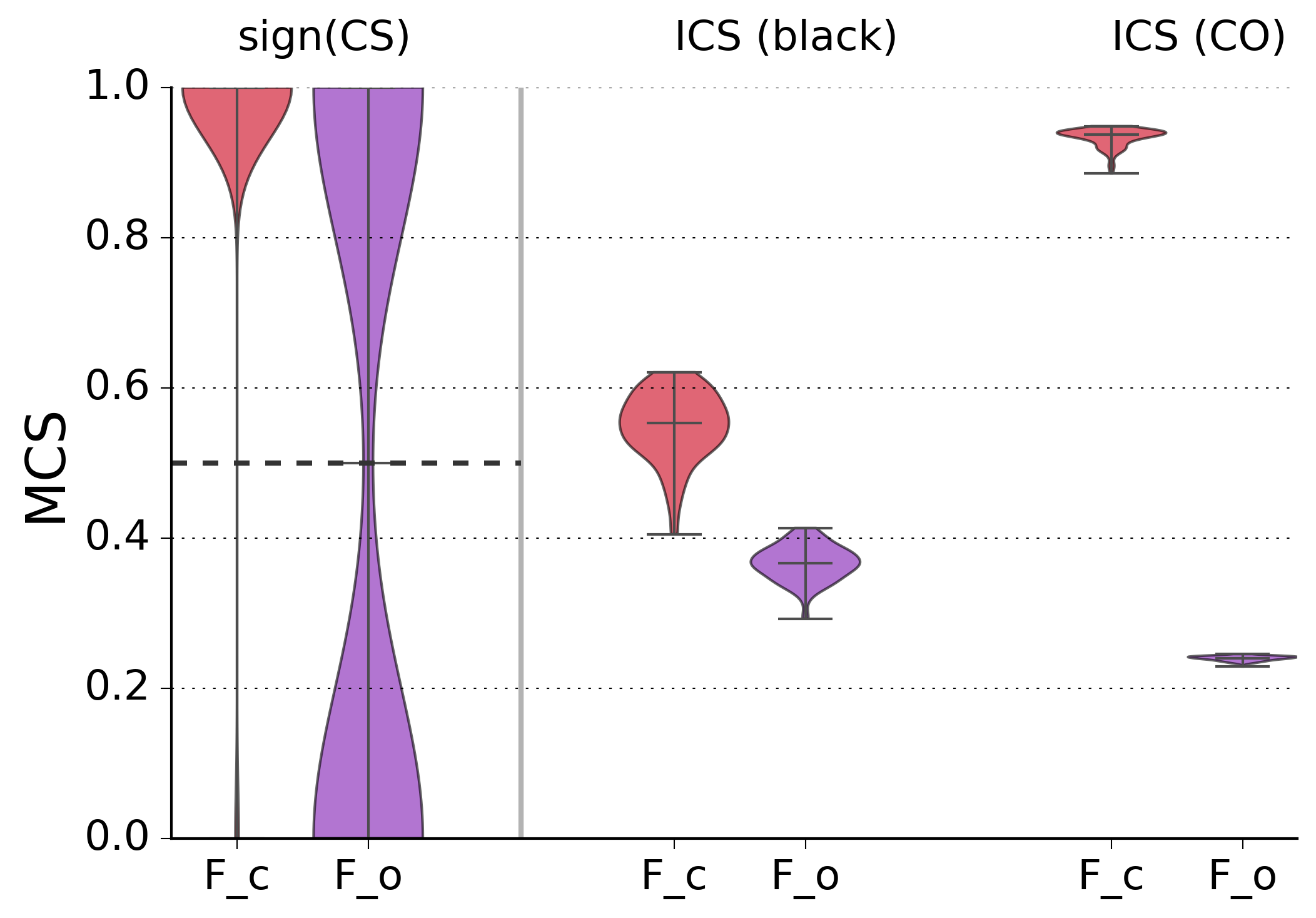} 
\end{subfigure}
\begin{subfigure}[t]{0.03\textwidth}
\textbf{c}
\end{subfigure}
\begin{subfigure}[t]{0.5\textwidth}
\includegraphics[width=\linewidth,valign=t]{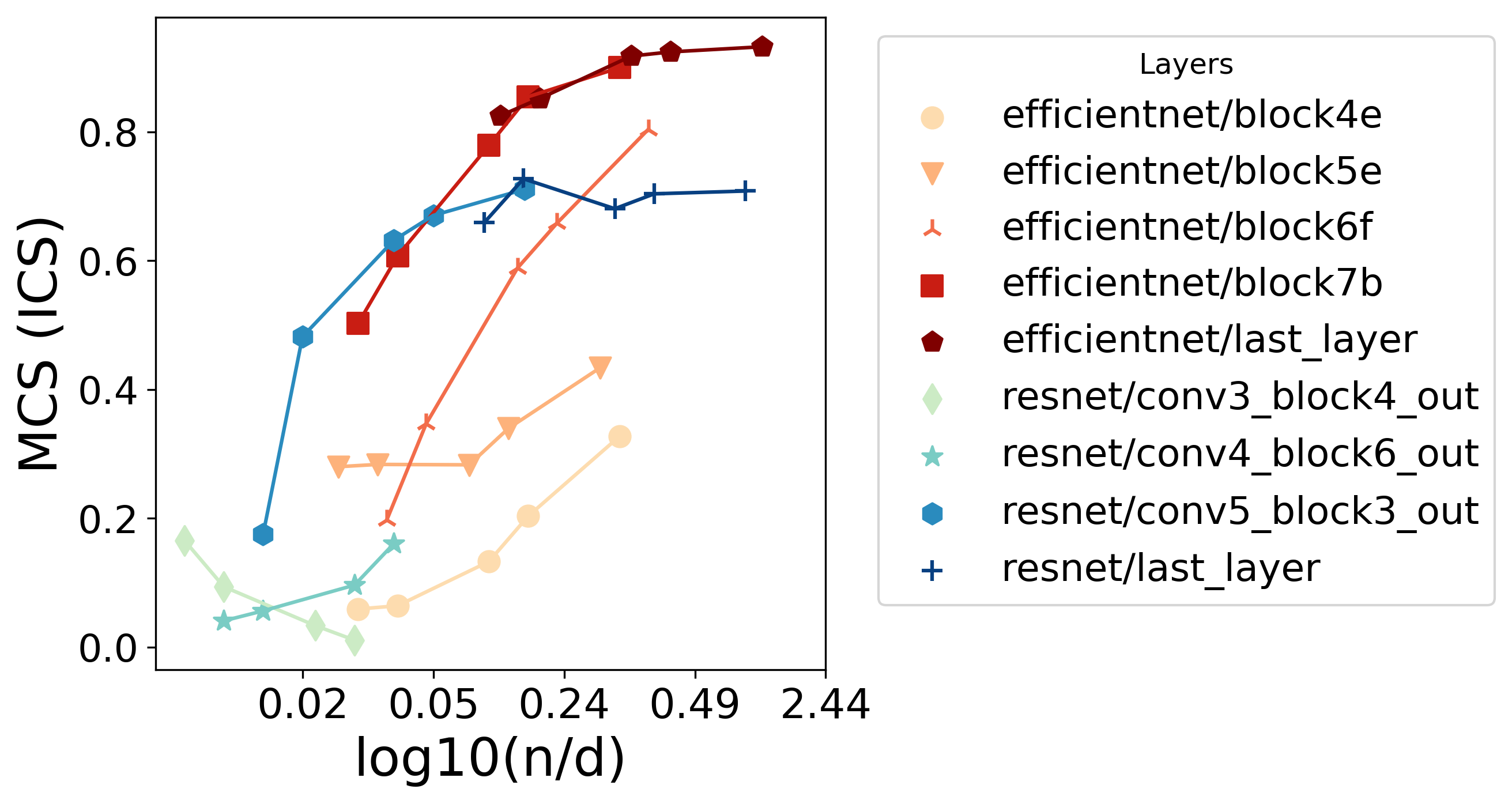}
\end{subfigure}\\
\begin{subfigure}[t]{0.03\textwidth}
\textbf{b}
\end{subfigure}
\begin{subfigure}[t]{0.42\textwidth}
\includegraphics[width=\linewidth,valign=t]{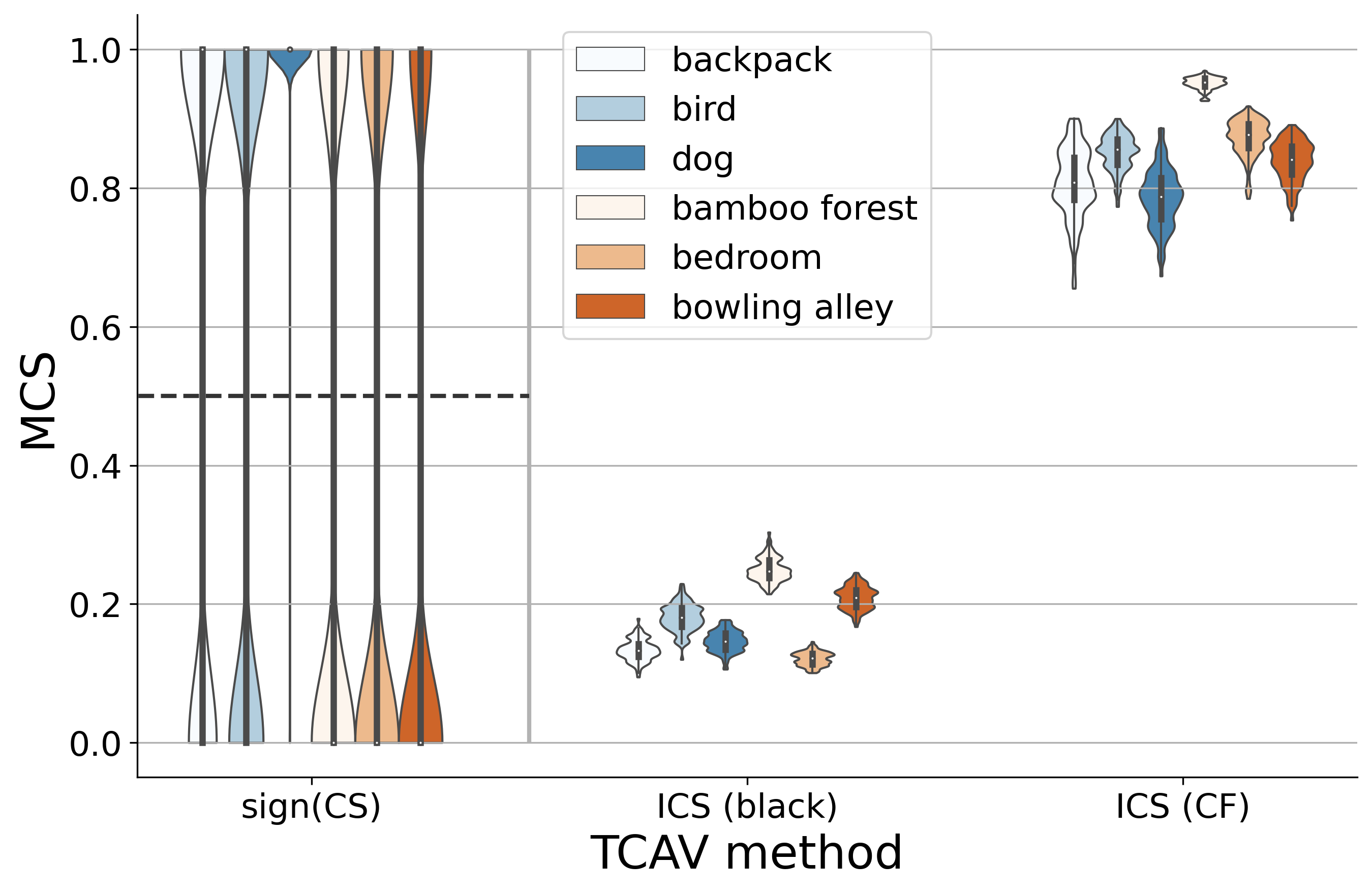}
\end{subfigure}
\begin{subfigure}[t]{0.03\textwidth}
\textbf{d}
\end{subfigure}
\begin{subfigure}[t]{0.5\textwidth}
\includegraphics[width=0.58\linewidth,valign=t]{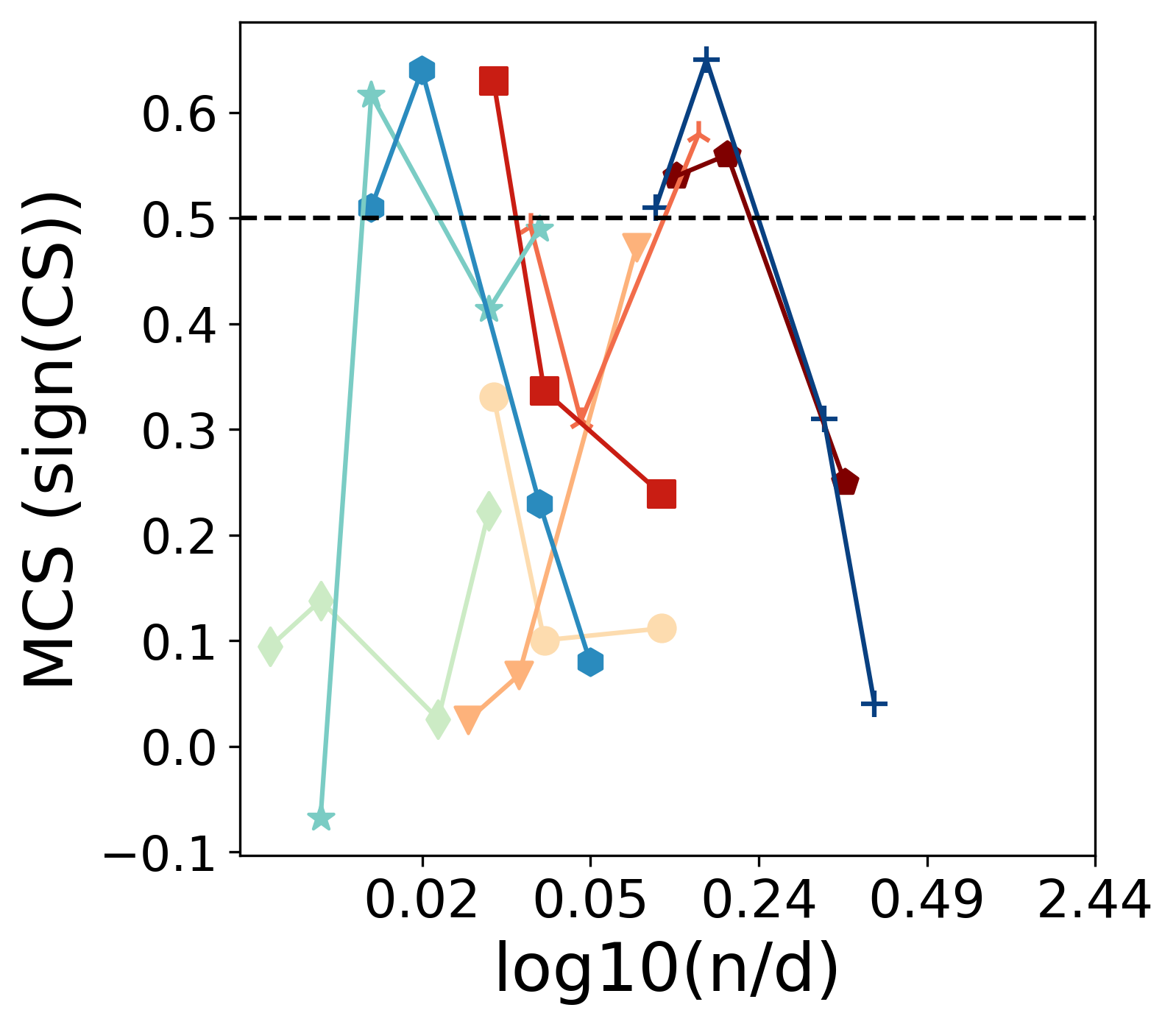}
\end{subfigure}
\caption{\textbf{MCS for $\mathrm{TCAV}^{sign(\mathrm{CS})}$ and $\mathrm{TCAV}^\mathrm{ICS}$. a} MCS distributions for BARS, computed with a black and concept-occluding (CO) baseline for ICS, for $F_c$ (red) and $F_o$ (purple). The dashed line represents the 0.5 ground truth for $\mathrm{TCAV}^{sign(\mathrm{CS})}$. \textbf{b} MCS at the last layer of EfficientNet with a black image and concept-forgetting (CF) baseline, for 3 objects and 3 scenes in BAM. \textbf{c} Varying $n$ for training CAVs across different layers of EfficientNet (orange to red) and ResNet (green to blue) improves MCS for $\mathrm{TCAV}^\mathrm{ICS}$ for a concept-forgetting baseline. \textbf{d} In contrast, MCS($\mathrm{TCAV}^{sign(\mathrm{CS})}$) scores fail to consistently improve.}
\label{fig:MCS_results}
\end{figure}
 
\subsubsection{$\mathrm{TCAV}^\mathrm{ICS}$ outperforms $\mathrm{TCAV}^{sign(\mathrm{CS})}$ in global explanations}

To showcase the global aspect of ICS, we compare  $\mathrm{TCAV}^{sign(\mathrm{CS})}$ and $\mathrm{TCAV}^\mathrm{ICS}$ for both BARS and BAM datasets.
Note that the best MCS score for $\mathrm{TCAV}^{sign(\mathrm{CS})}$ is 0.5, while the larger the score the better for  $\mathrm{TCAV}^\mathrm{ICS}$. We first observe that $\mathrm{TCAV}^{sign(\mathrm{CS})}$ is typically bimodal (either 0 or 1) for irrelevant concepts in a statistically significant manner (i.e., not close to the chance-level 0.5), leading to MCS scores that do not reflect the ground truth of 0.5. On the other hand, $\mathrm{TCAV}^\mathrm{ICS}$ gives irrelevant CAVs a score close to 0 consistently and positive scores to relevant concepts, leading to positive MCS scores. These findings are replicated across multiple concepts and baselines for both BARS (Figure~\ref{fig:MCS_results}a) and BAM (Figure~\ref{fig:MCS_results}b). Our results hence suggest that $\mathrm{TCAV}^\mathrm{ICS}$ is more faithful than $\mathrm{TCAV}^{sign(\mathrm{CS})}$ to estimate concepts' global importance, with the caveats discussed in Section~\ref{sec:curse_of_dimensionality}. Please see Supplements~\ref{app:BARS_baseline} and \ref{app:BAM_baseline} for detailed MCS and ICS scores across datasets, concepts and baselines.

\subsubsection{Improving the effect of the curse of dimensionality on CAVs}
\label{sec:curse_of_dimensionality}
When scaling to larger models with small $n$ (number of concept pictures), naively applying ICS can lead to degraded results (i.e. values <0.1\%).
Given that the degradation is more acute for shallow layers which tend to be wider, we hypothesize that these results are due to the curse of dimensionality, as many different directions can encode a concept. 
To test this hypothesis, we decrease the $\frac{n}{d}$ ratio from 30 to 0.1 to train CAVs on BARS, resulting in MCS scores 3 times smaller (45\% vs 15\%). We are able to mitigate this by augmenting the CAV training data with widely used image transformations such as random flips and changes of contrast, brightness, hue, and saturation. 
With these modifications, MCS scores on BAM can be improved for most layers (more impact on the deeper layers, Figure~\ref{fig:MCS_results}c). This is intuitive; one should get `better' quality of CAVs as you increase $n$. In contrast, $\mathrm{TCAV}^{sign(\mathrm{CS})}$ was not consistently improved by the augmentation, given its formulation based on the sign of CS as shown in Figure~\ref{fig:MCS_results}d.

\subsection{Qualitative illustration}
\label{sec:results_imagenet}
We use the ImageNet dataset~\cite{Deng09} to generate some of the concepts demonstrated in~\cite{Kim2017} and replicate results at the global level. We experiment with multiple network architectures: EfficientNet (presented in the main text, \cite{Tan2020}), ResNet50 \cite{He2015}, Inception-v1 \cite{Szegedy2015}, and MobileNet-V2 \cite{Sandler2018} which produced consistent results. We report results using a white image as the baseline, but a black baseline also produced consistent results. Alongside with global results, we showcase local examples that demonstrate the intuitive nature of ICS scores. 

\begin{figure}[!t]
\centering
\includegraphics[width=\linewidth]{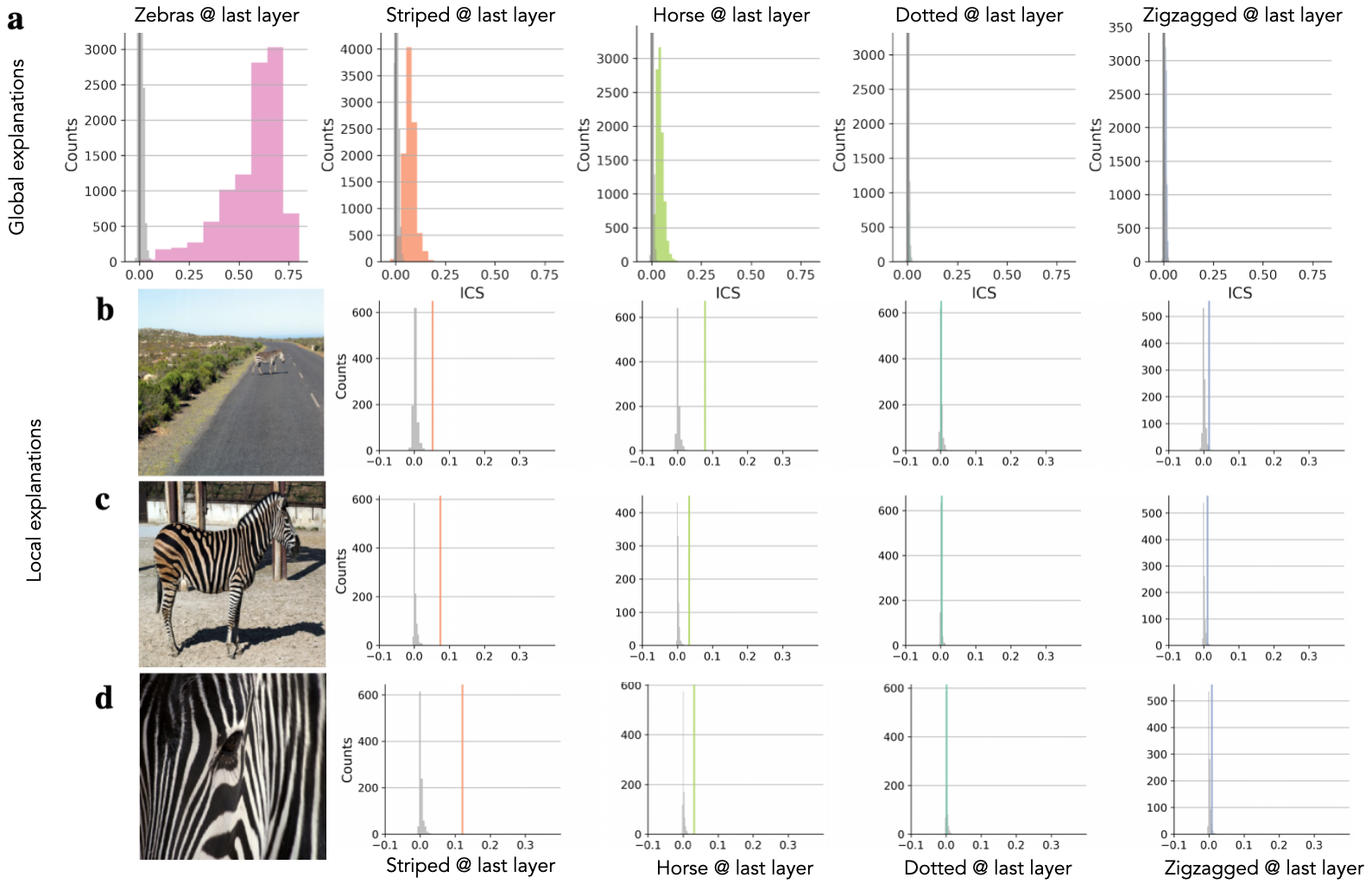}
\caption{\textbf{ICS scores for the ``zebra'' class.} Distribution of ICS scores across 100 pictures of zebra, for 5 concepts (zebra, striped, horse, dotted, zigzagged), computed on the last layer of an EfficientNet-B3 model. Gray histograms represent null distributions obtained by CAV trained on permuted labels. \textbf{a} Global scores. \textbf{b-d} 3 cherry picked samples, representing far and close distance between the camera and the zebra. The zebra becomes more ``stripy" when it is more zoomed-in.}
\label{fig:ics_zebra}
\end{figure}

\textbf{Zebra class: } We first focus on the ``zebra'' class and build CAVs to represent the ``striped'', ``dotted'', ``zigzagged'', ``zebras'', and ``horse'' concepts to draw similarity and contrast with the analyses done in~\cite{Kim2017}). $\mathrm{TCAV}^\mathrm{ICS}$ (Figure \ref{fig:ics_zebra}a) is able to replicate the high $\mathrm{TCAV}^{sign(\mathrm{CS})}$ scores for the striped concept, as well as for the ``zebra'' concept (as a sanity check) and for the ``horse'' concept (more detail on CAV building in Supplement~\ref{app:imagenet_cavs}). Similar to~\cite{Kim2017}, ``zigzagged'' and ``dotted'' have significantly small $\mathrm{TCAV}^{sign(\mathrm{CS})}$ scores (0.0002 and 0.1760, respectively), which reflects their negative influence on the model prediction for the class ``zebra''. We observe one important difference: contrarily to $\mathrm{TCAV}^{sign(\mathrm{CS})}$, the ICS distributions for ``dotted'' and ``zigzagged'' overlap with their associated null distribution (grey histogram in Figure \ref{fig:ics_zebra}) at both global (Figure \ref{fig:ics_zebra}a) and local levels (Figure \ref{fig:ics_zebra}b-d). This potentially hints that ICS is better at separating the meaningful concepts from null. Please see Supplement~\ref{app:imagenet} for more global results for 3 other architectures.

Figure~\ref{fig:ics_zebra}b-d demonstrate the local explanation aspect of ICS. When the zebra is far away from the camera (b), it may look less stripy, more like a horse than if the zebra is close to the camera. This is well-reflected with ICS scores. As we \emph{zoom in} on the zebra ((c) and (d)), stripes are obtaining higher ICS scores, while the ``horse'' concept becomes less important in the model's decision. Using ICS, we can compare local scores with the global distributions, to contrast local compared to global behaviors. Comparing these distributions suggests that the score for `striped' obtained in the zoomed in image is higher than that of the class. Being able to map a complex network's decision to what humans can understand with this level precision is encouraging.
Note that two irrelevant concepts, `zigzagged' and `dotted' have their ICS distributions closer to the null at all camera distances. 

\begin{figure}[!t]
\centering
\includegraphics[width=\linewidth,valign=t]{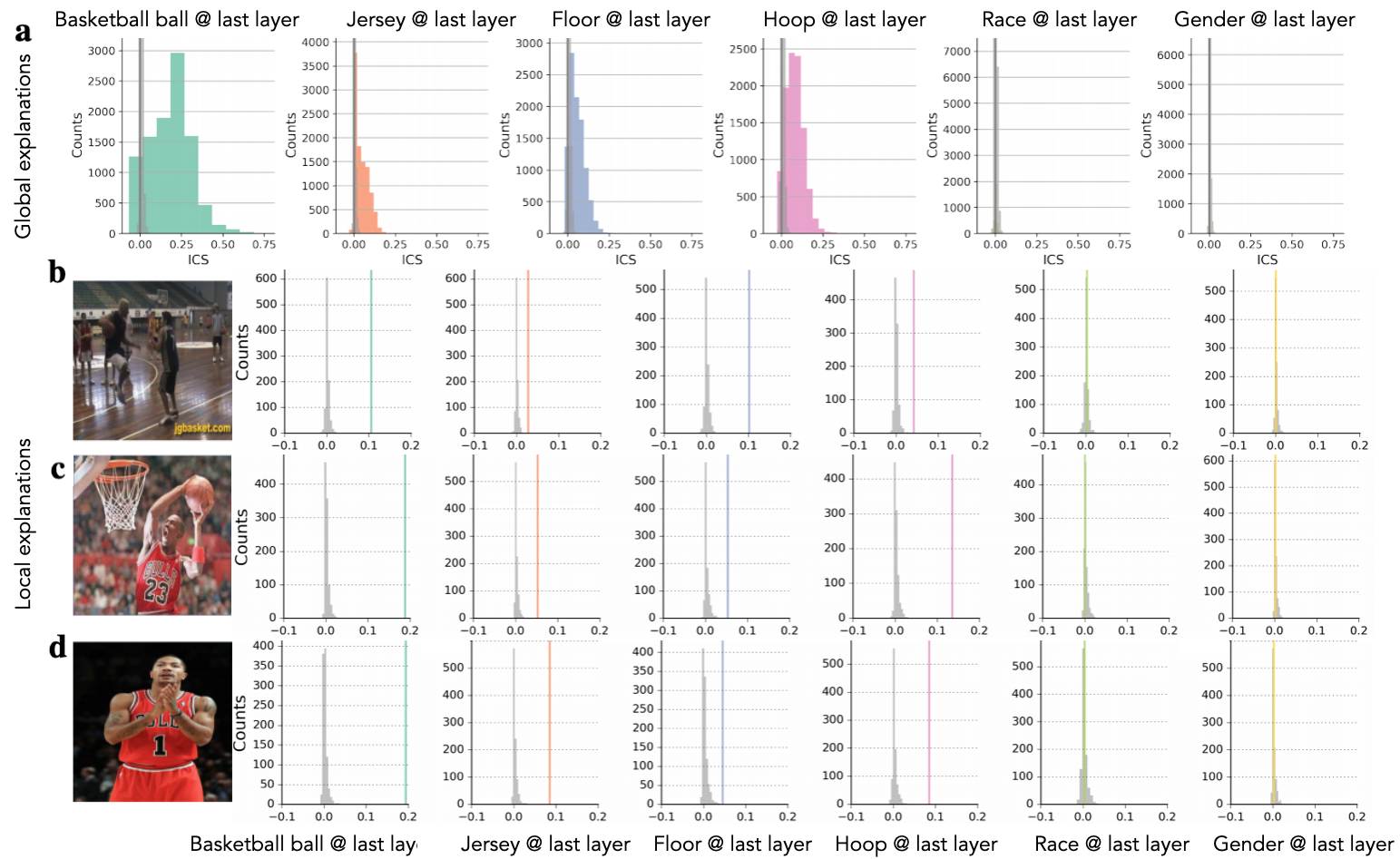}
\caption{\textbf{ICS scores for the ``basketball'' class.} Distribution of ICS scores (white image baseline) across 100 pictures of basketball, for 5 concepts (ball, jersey, floor, hoop, race, gender), computed on the last layer of an EfficientNet-B3 model. Gray histograms represent null distributions obtained by CAV trained on permuted labels. \textbf{a} Global scores. \textbf{b-d} 3 cherry picked samples displaying different elements.}
\label{fig:ics_basketball}
\end{figure}

\textbf{Basketball class: }As a second illustration, we investigate the ``basketball'' game class and define 4 concepts that are directly related to the game: the ball, the jersey, the floor, and the hoop. As racial biases have been highlighted in \cite{Kim2017} for this class, we also build perceived ``gender'' (man vs woman) and perceived ``race'' (darker vs lighter skin) concepts\footnote{See Supplement~\ref{app:imagenet_cavs} for a discussion on how these concepts were obtained and what proxy signal we believe they represent}. All CAVs are estimated as ``significantly encoded'' in the last layer of the model, with performance of the linear classifier $\sim 99\%$ for the game related concepts, 94\% for gender and 75\% for race. $\mathrm{TCAV}^{sign(\mathrm{CS})}$ scores for the game-related concepts are 1, while they are 0.85 and 0.91 for gender and race, respectively. ICS scores however display a ``ranking'' of concepts in terms of global scores: the ball has the largest impact, followed by the hoop, then floor and jersey (Figure \ref{fig:ics_basketball}). On the contrary, ICS distributions for race and gender are indistinguishable from the nulls. We qualitatively tested a few manually selected images with strong minority attributes, and observed that this network is able to correctly predict the ``basketball'' class. Note that this is not a proof that this model is not biased. 

When investigating local examples, we observe that ICS is able to describe fine grain level of local explanations (Figure \ref{fig:ics_basketball}b-d): it separates an image with less visible jersey (b) to more visible jersey (d), less floor (c, d) to more floor (b), and the existence of the hoop (c) correctly. Interestingly, the ICS scores for the ball are high in general, consistent with the global explanation. It turns out that ImageNet images for the  ``basketball'' class include many images that only contain the ball, making the ball and the basketball game highly correlated. In this case again, directly comparing the global and local explanations helped in understanding how an example image contrasts to the general model behavior.

\begin{figure}[!t]
\centering
\begin{subfigure}[t]{0.01\textwidth}
\textbf{a}
\end{subfigure}
\begin{subfigure}[t]{0.45\textwidth}
\includegraphics[width=0.9\linewidth,valign=t]{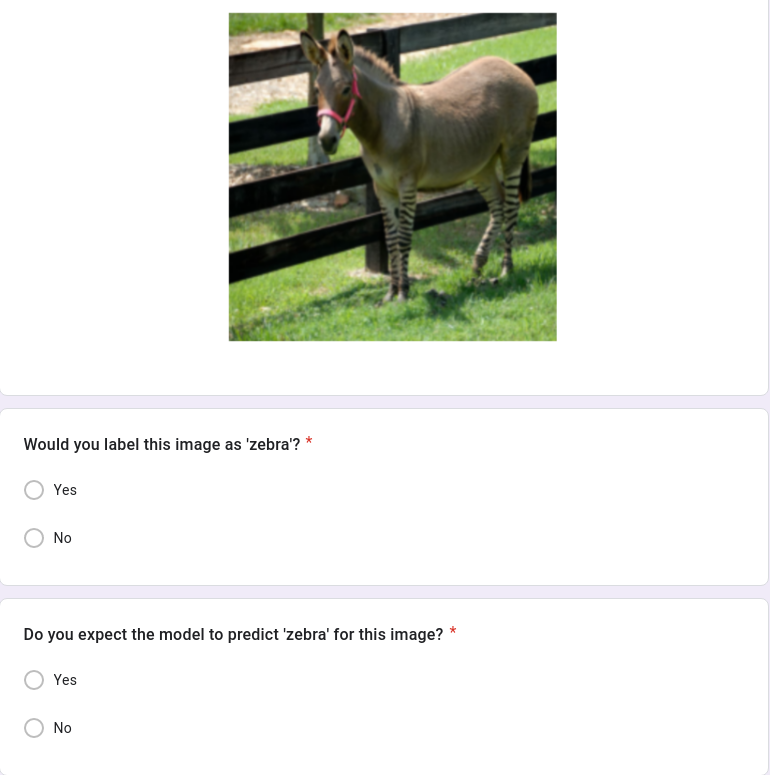} 
\end{subfigure}
\begin{subfigure}[t]{0.45\textwidth}
\includegraphics[width=0.85\linewidth,valign=t]{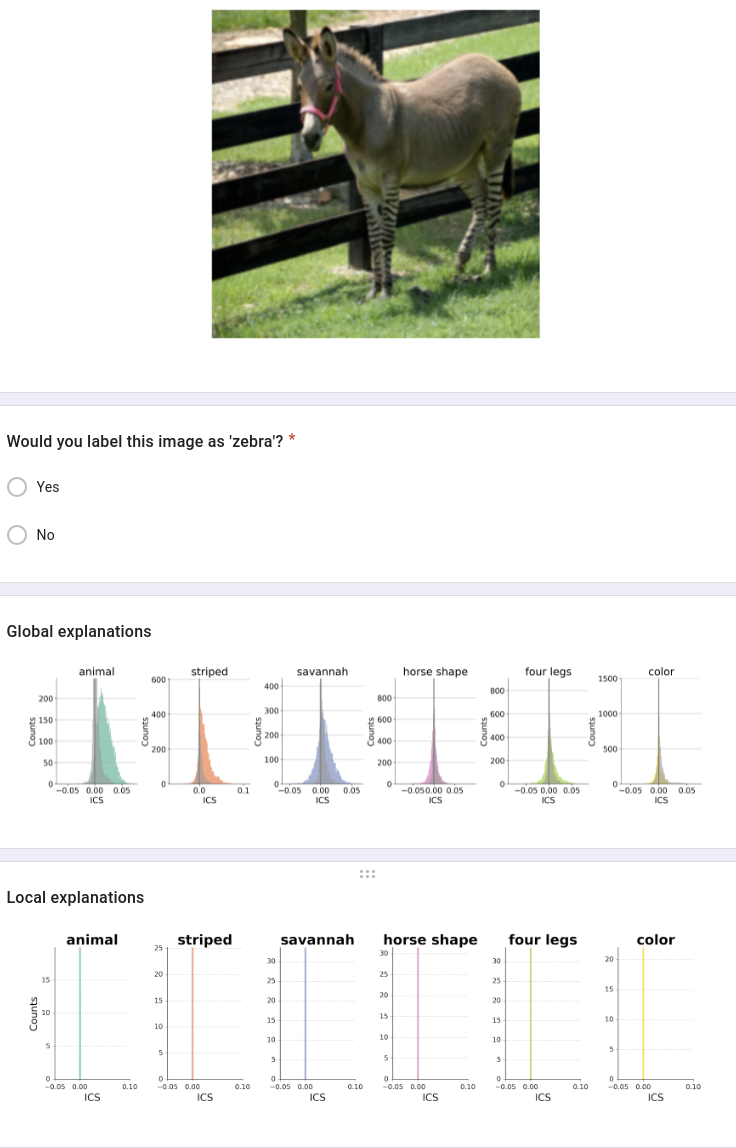} 
\end{subfigure}
\caption{\textbf{User study. a} Example of question for Group 1. \textbf{b} Group 3 sees the concept global explanations as well as the local ICS scores for each image (Note that the second question is out of view on this image only). The animal in the image is a crossing between a donkey and a zebra. The majority human label is `not zebra', and the model prediction is not zebra. Of note, ICS scores are zero for all concepts and overlap the null distributions.}
\label{fig:user_study}
\end{figure}

\subsection{User study}
Finally, we perform a user study to demonstrate the usefulness of the method to users. Based on evidence that machine learning researchers and practitioners are the main consumers of model explanations \cite{Bhatt2020-mz}, we recruited machine learning researchers and practitioners. The task was based on `model simulability' \cite{Fel2021-ce}, i.e. are the stakeholders able to simulate the behavior of the machine learning model? To assess model simulability, we referred to the zebra class as predicted by the Inception architecture trained on ImageNet from section~\ref{sec:results_imagenet} as our target class and asked participants to simulate the model's prediction when given an image. More specifically, each participant had to answer two questions per image (Figure~\ref{fig:user_study}a):
\begin{itemize}
    \item \textit{Would you label this image as `zebra'?} The participants were instructed to answer with their top-1 ground truth. The answer was a binary yes or no.
    \item \textit{Do you expect the model to predict `zebra' for this image?} For this question, the participants had to guess whether the model's top-1 prediction was zebra or not for this image. The answer was a binary yes or no.
\end{itemize}

The images were a set of 40 images manually selected from the internet (see Supplement Figure ~\ref{app:fig_user_study_images}). As these images were mostly unlabelled, the first question aims at obtaining the human label. To answer the second question, all participants were given some basic information on the model, including access to the list of 1,000 classes that the model is trained to predict, some training examples for the zebra class, the model's top-1 performance across all classes, the sensitivity score for the zebra class (i.e. percentage of time predicting zebra when given a zebra-labelled image), and three examples of predictions when given a zebra-labelled image. This basic information aims to provide a similar level of intuition as if presented with a short model card (e.g. \cite{Mitchell2019-ma}). Participants were then divided into three groups:
\begin{enumerate}
    \item \textbf{Group 1}: no model explanation.
    \item \textbf{Group 2}: Global model explanations. For this group, we described the concepts selected and displayed the results of global explanations.
    \item \textbf{Group 3}: Local and global explanations. This group had access to the global explanations as in Group 2 and to local explanations computed with ICS for each image.
\end{enumerate}

To provide model explanations, we built concepts (and thus CAVs) as described in Section~\ref{sec:results_imagenet}, expanding the list of concepts to cover more aspects of the class. We hence built CAVs for `animal', `striped', `savannah', `four legged', `horse shape' and `color' concepts. We note that the explanations were presented using the plots in Figure~\ref{fig:user_study}b and that no further UX research was performed to optimize the display of the explanations. The study was conducted via Google forms, after obtaining participant consent. All participants received incentives for their time. Please see Supplement~\ref{app:user_study_design} for full details on the user study design.

\begin{figure}[!t]
\centering
\begin{subfigure}[t]{0.01\textwidth}
\textbf{a}
\end{subfigure}
\begin{subfigure}[t]{0.42\textwidth}
\includegraphics[width=0.9\linewidth,valign=t]{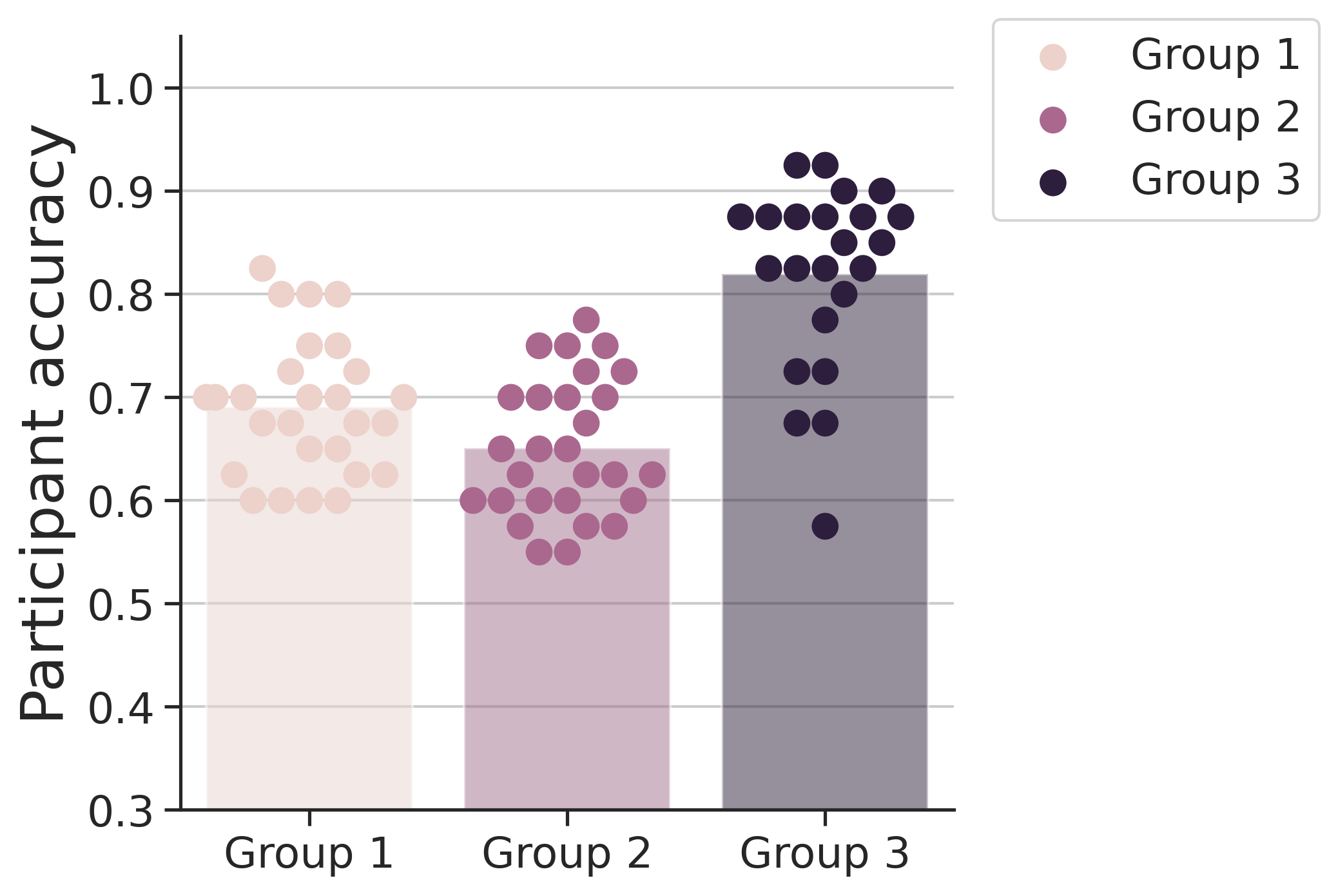} 
\end{subfigure}
\hspace{0.9cm}
\begin{subfigure}[t]{0.01\textwidth}
\textbf{c}
\end{subfigure}
\begin{subfigure}[t]{0.42\textwidth}
\includegraphics[width=0.7\linewidth,valign=t]{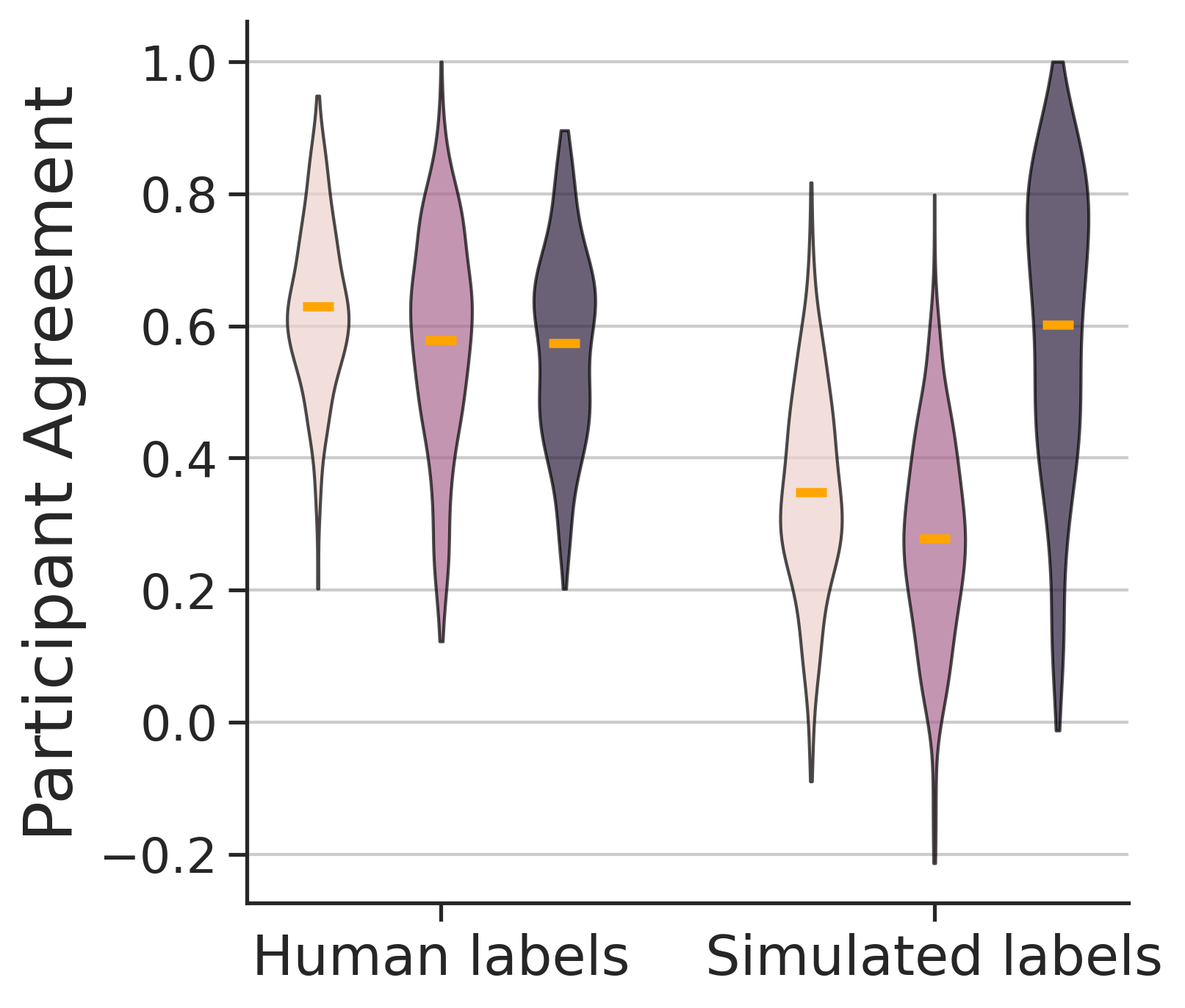}
\end{subfigure} \\
\vspace{0.1cm}
\begin{flushleft}
\textbf{b}\hspace{0.1\columnwidth}Correct\hspace{0.26\columnwidth}Difficult\hspace{0.26\columnwidth}Incorrect \\
\end{flushleft}
\begin{subfigure}[t]{0.01\textwidth}
\end{subfigure}
\begin{subfigure}[t]{0.32\textwidth}
\includegraphics[width=0.8\linewidth,valign=t]{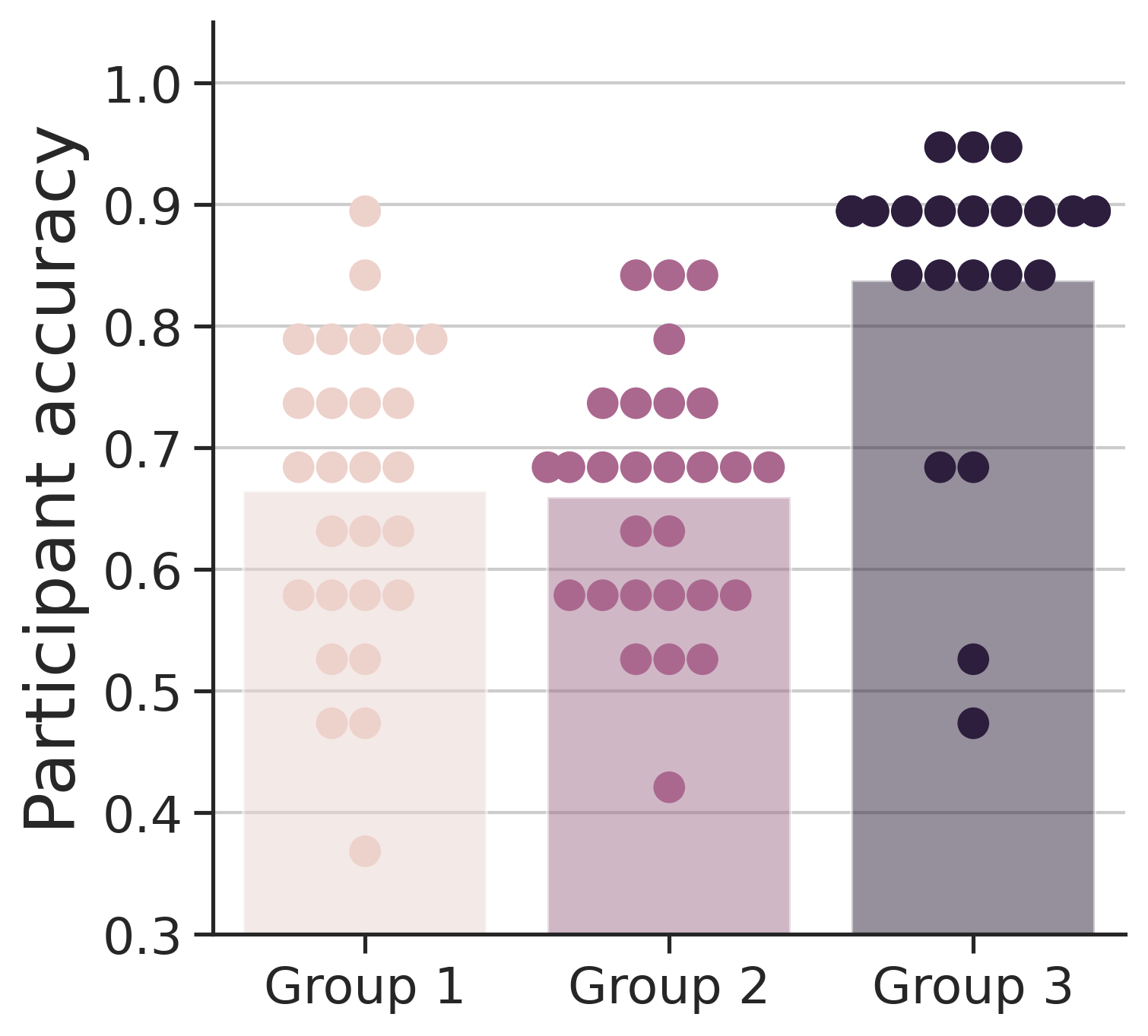} 
\end{subfigure}
\begin{subfigure}[t]{0.32\textwidth}
\includegraphics[width=0.8\linewidth,valign=t]{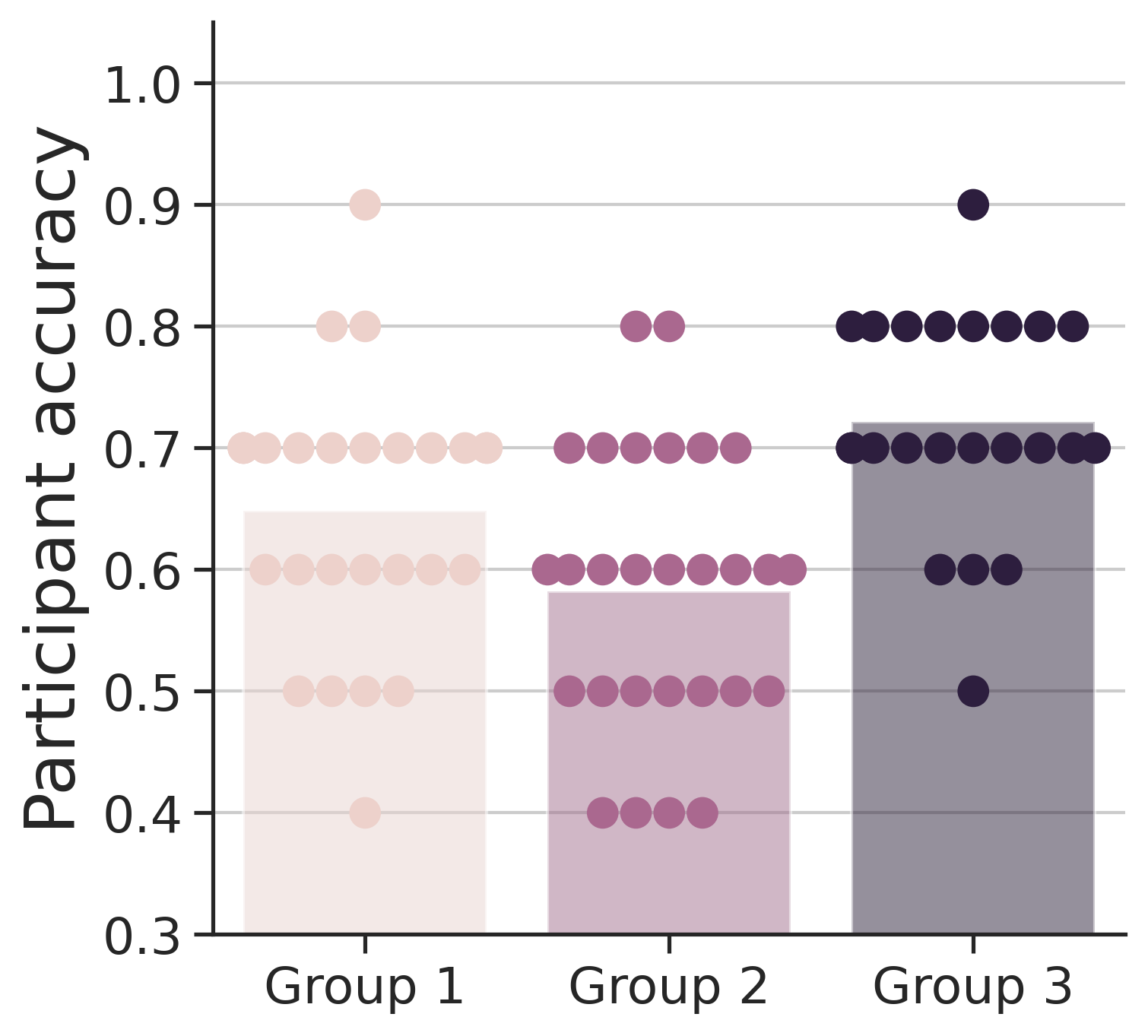} 
\end{subfigure}
\begin{subfigure}[t]{0.32\textwidth}
\includegraphics[width=0.8\linewidth,valign=t]{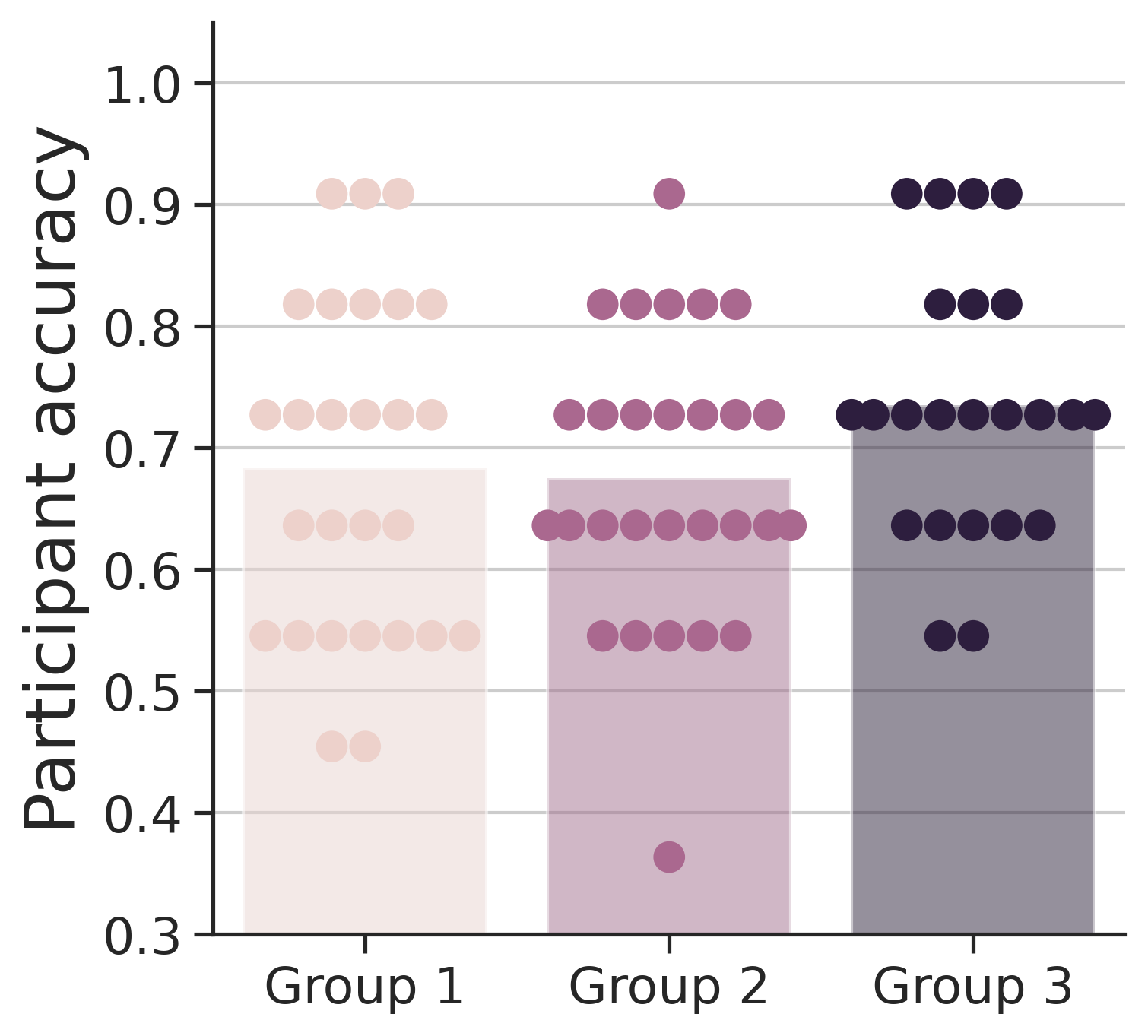}
\end{subfigure}
\caption{\textbf{User study results. a} Participant performance in each group, as measured by their accuracy to predict the model's top-1 prediction for the zebra class. Each point represents the performance of a participant across the 40 questions and each bar represents the average within group. \textbf{b} Same plot as \textbf{a} for `correct', `difficult' and `incorrect' predictions, as defined by the majority human label. \textbf{c} Cross-participant agreement (Cohen's kappa) in each group on their labeling of the image (left) and their simulated labels of the model's top-1 output (right). Orange lines depict the mean across pair-wise comparisons.}
\label{fig:user_study_results}
\end{figure}

78 participants completed the study (group 1: $n=27$, group 2: $n=28$, group 3: $n=23$), leading to 3,120 human labels and 3,120 model simulability estimates. There were no significant differences across participants in terms of machine learning experience (in years, group 1: $5.9 \pm 3.8$, group 2: $ 7.0 \pm 5.2$, group 3: $6.2 \pm 4.9$), deep learning experience, or experience with interpretability techniques (Supplement~\ref{app:user_study_participants}). We note that 38 participants were researchers and 40 were ML practitioners, providing a balance between research and applied ML. All participants were instructed to refer to their ML experience, and those in groups 2 and 3 were instructed to use the explanations as an extra help to accomplish the task.

We first estimate each participant's accuracy at simulating the model's top-1 prediction across all questions in each group. Surprisingly, our results show that providing global explanations does not help participants compared to providing no explanations (group 1 accuracy: $0.69 \pm 0.06$, group 2 accuracy: $0.65 \pm 0.07$, Figure~\ref{fig:user_study_results}a), and might even hurt slightly (t-test: $p=0.03$, uncorrected). However, providing local explanations in addition to the global explanations led to a significant increase in participants' performance (group 3: $0.82 \pm 0.09$, $p<0.05$ Bonferroni corrected).

As the images are unlabelled, we refer to participants' answers to the first question (`Would you label this image as zebra?') to define human labels for each image. The label for an image was defined as the majority vote (`zebra' or `not zebra') if participants agreed at an $80\%$ rate for this image ($n=78$, arbitrarily selected threshold). Images for which this threshold was not met were labelled as `difficult' and not given a `zebra/not zebra' label ($n=10$ questions). Assuming that the derived human labels are correct, we note that providing local and global explanations helped most across the`correct' predictions (Figure~\ref{fig:user_study_results}b, group 1 accuracy: $0.66 \pm 0.12$, group 2 accuracy: $0.66 \pm 0.10$, group 3 accuracy: $ 0.84 \pm 0.12$), as well as for the difficult questions (group 1 accuracy: $0.65 \pm 0.11$, group 2 accuracy: $0.58 \pm 0.11$, group 3 accuracy: $ 0.72 \pm 0.09$). The improvement for incorrect predictions was marginal (group 1 accuracy: $0.68 \pm 0.14$, group 2 accuracy: $0.68 \pm 0.12$, group 3 accuracy: $ 0.74 \pm 0.11$, $p>0.05$). We caveat this analysis by the small sample size ($n=40$ questions, $n=11$ incorrect predictions and $n=10$ `difficult' questions). 

Finally, we assess the agreement between participants on the human labels (i.e. answer to the first question) and on the human simulated labels of the model's top-1 output (i.e. answer to the second question), in each group (Figure~\ref{fig:user_study_results}c). We first note that the human label agreement across all participants (pair-wise Cohen's Kappa \cite{cohen1960,Artstein2008}: $\kappa=0.60 \pm 0.14, n=3,003$), was similar to the agreement within the groups (group 1: $\kappa=0.63 \pm 0.13$, group 2: $\kappa=0.58 \pm 0.17$, group 3: $\kappa=0.57 \pm 0.15$). On the other hand, the participants' agreement on the simulated labels was starkly higher in group 3 ($\kappa=0.60 \pm 0.22$) than in the other groups (group 1: $\kappa=0.35 \pm 0.16$, group 2: $\kappa=0.28 \pm 0.16$). We therefore conclude that participants in group 3 were better able to predict the model's behavior, and that their understanding of the model's predictions aligned more between each other than in the other groups.

\section{Discussion}

TCAV \cite{Kim2017} sheds light into deep neural networks inner workings by allowing users to probe their learnt internal representations, while IG made headway to provide principled individual data point explanations. Each techniques is well validated in the literature and applications, and has mostly mutually exclusive pros and cons (e.g., local vs global).
By combining TCAV with integrated gradients \cite{Sundararajan2017}, we show that TCAV can provide \textit{local explanations}. This combination hence opens a new avenue for both techniques.

Our work also allowed to uncover a limitation of TCAV and improve on it. We demonstrated that $\mathrm{TCAV}^{sign(\mathrm{CS})}$ might not be faithful, with scores for irrelevant concepts flipping from 0 to 1 across bootstraps. $\mathrm{TCAV}^\mathrm{ICS}$ consistently provided more sensitive global explanations on two synthetic datasets and multiple networks.
  
Interpretability techniques can be evaluated in terms of functionally-grounded (proxy metrics, proxy tasks), human-grounded (humans, proxy tasks) and application-grounded (humans, real-world tasks) metrics \cite{Doshi-Velez2018}. Using multiple models, concepts and baselines, we provide a thorough functionally-grounded evaluation of ICS. More importantly, we demonstrate the usefulness of the method to machine learning researchers and practitioners when they are tasked with reproducing the behavior of a deep learning model, i.e. a human-grounded evaluation. Our results show consistent improvement in participant performance across multiple types of predictions when they are provided with both the global and the local explanations. This is especially promising given that no UX research was performed to ease their understanding of the ICS scores or plots. Combined with the human-understandability of concept-based explanations, this result is encouraging for end-user explanations and could be further explored to bridge the understandability gap at the local level \cite{Ghassemi2021-ic}, potentially leading to application-grounded evaluations.

Interestingly, the user study displayed that global explanations did not lead to an increase in participant performance. We however note that the design of the study is inherently local, and that many images selected represent edge cases (even difficult for humans to label). Therefore, it is possible that this study was not best suited to assess the usefulness of global explanations. Nevertheless, participants in Group 3 have reported using both types of explanations as a `weighted combination', in which they would weight local ICS scores by the global scores (such that a positive ICS score for `animal' had more weight than a negative ICS score for `color') before predicting the model's output. Therefore, the combination of local and global explanations seemed useful to (some) participants.

\textbf{Limitations and future work: }While we obtain the `best of both worlds', our technique still inherits the limitations of both methods. 
More specifically, building reliable CAVs is crucial for TCAV methods to be trusted, which can be challenging in high-dimensionality or due to confounding. Confounding can be mitigated by careful inspection of the training set, or by using causal approaches~\cite{Goyal2019, debiasingConcept2020}. Regularizing the CAV, considering smaller layers, or increasing the CAV training set size can help in higher dimensions. Selecting concepts can also be challenging in complex, real-world applications. Automatic concept-definitions have been proposed to tackle this issue in computer vision \cite{Ghorbani2019,Yeh2020-xq}. In real-world applications, practitioners have defined concepts based on domain knowledge \cite{Mincu2021,Cai2019,Reyes2020-ym,Clough2019,Graziani2018}. We note that these limitations are independent of the choice of CS or ICS.

For IG, a baseline needs to be defined, which affects the obtained explanation. In our work, we have observed variability in the support of ICS in BARS across baselines, and different levels of improvements in terms of MCS when using augmentation to alleviate the curse of dimensionality on CAVs. As discussed in multiple previous works (e.g. \cite{Sundararajan2017, Ancona2019, Sturmfels2020}), results are expected to vary across baselines as each baseline represents a specific definition of `missingness'. It would however be interesting to consider recent approaches to baselines such as aggregations over multiple baselines, a combination of white and black baselines \cite{Kapishnikov_2019_ICCV}, or reformulations of integrated gradients \cite{Erion2020}. Results should also be evaluated in terms of robustness to adversarial attacks \cite{Alvarez2018, Dombrowski2019, Ghorbani2017, Chih-Kuan2019}, as \cite{Dombrowski2019} have shown that integrated gradients are highly sensitive to this kind of attack. We leave this evaluation for future work.  We also note that alternative formulations of (integrated) gradients could be investigated for a potential combination with TCAV, e.g. Smoothgrad \cite{Smilkov2017-vc}, XRAI \cite{Kapishnikov_2019_ICCV} or guided IG \cite{Kapishnikov2021-aa}.

At the local level on ImageNet, we observe that ICS can provide intuitive results. However, the user study revealed that users were questioning the `coverage' of the set of concepts selected, and wondered whether null scores could be trusted as the reflection of a negative prediction for the class of interest. This is likely reflected in the marginal improvement in participant performance observed on questions where the model performed `incorrectly' (according to the human labels). Therefore, future research could investigate how concept-based explanations could potentially be combined with feature-based attributions to provide a full picture to the user.

In conclusion, we hope that our work demonstrates the potential and pitfalls of combining local and global explanation techniques for model interpretability, and will encourage further work in this direction.

\section*{Acknowledgements}
We would like to thank Mengjiao Yang for sharing BAM models, and Yash Goyal for sharing the synthetic data generation. We would also like to thank Mahima Pushkarna for helping with figures and advising on the user study. Finally, we thank collaborators in Google Research, Search and in Google Health and all participants in the user study.

\bibliographystyle{alpha}
\bibliography{refs}   

\newpage
\section*{Appendix}
\appendix
\input{FAccT_2022_supp}

\end{document}

%% file: FAccT_2022_supp.tex
\section{Methods}
\subsection{Choice of the baseline}
\label{app:baseline}
It has been suggested~\cite{Kindermans2017,Ancona2018} that the attributions may not be invariant to the choice of the baseline used in the computation of integrated gradients (IG, \cite{Sundararajan2017}). In this section, we display the baselines considered for investigation throughout this work, while their influence on the proposed method is extensively studied in sections \ref{app:BARS_baseline} and \ref{app:BAM_baseline}. We divide the baselines based on their `informativeness'.

\subsubsection{Uninformative baselines}
\label{section:uninformative_baselines}

As in \cite{Sundararajan2017}, the baseline is typically chosen to be uninformative. This characteristic is fuzzy and can be defined in various ways. \cite{Sturmfels2020} provides interactive visualization of the impact of this choice on the resulting saliency maps. 

Below, you can find a list of various ways of defining `uninformative'. 
\begin{itemize}
    \item \textit{Zero image baseline}, i.e. black image: $\mathbf{a'} = f_l(\mathbf{0})$
    \item \textit{One image baseline}, i.e. white image: $\mathbf{a'} = f_l(\mathbf{1})$
    \item \textit{Noisy image baseline}: $\mathbf{a'} = f_l(\mathcal{N})$ where $\mathcal{N}$ is a noise distribution.
    \item \textit{Pixel-wise average baseline}: $\mathbf{a'} = f_l(\frac{1}{N}\sum_{n=1}^N\mathbf{x_n})$
    \item \textit{Pixel-wise median baseline}: $\mathbf{a'} = f_l(\mathrm{median}_n\mathbf{x_n})$
    \item \textit{Entropy-maximizing baseline}: $\mathbf{a'} = \mathrm{argmin}_{\mathbf{x}} ||\mathbf{a}-\mathbf{x}|| + \lambda \mathcal{H}(h(\mathbf{x}))$ where $\lambda \in \mathbb{R}+^*$ is some large number, $\mathcal{H}(h(\mathbf{x}))$ is the entropy of the output of the neural network when activations at hidden layer $l$ are equal to $\mathbf{x}$.
\end{itemize}
The main text mostly reports results using the `One image baseline' (i.e. a white image input) \cite{Kindermans2017,Kapishnikov_2019_ICCV}.

\subsubsection{Informative baselines}
\label{app:informative_baselines}

\begin{itemize}
    \item \textit{Concept-occluding baseline}: $\mathbf{a'} = \mathbf{a} - (\mathbf{a}^T\mathbf{v_C}+b)\frac{\mathbf{v_C}}{||\mathbf{v_C}||_2^2}$: the baseline with the concept removed, where $b$ is the bias of the linear model. In that case, ICS is equal to the occluded concept prediction difference.
    \item \textit{Concept-forgetting baseline}: Similarly to the previous baseline, we can define $\mathbf{a'} = \mathbf{a} - \lambda\mathbf{v_C}$ for some $\lambda \in \mathbb{R}^*+$, which is then a generalization.
\end{itemize}

Please note that $\mathbf{v_C}$ is not assumed to be unit-normed for informative baselines.

In this work, we present results for the `Concept-forgetting baseline' as this baseline leads to a closed form solution for ICS (see Sec.\ref{app:closed_form}). We however note the presence of a hyper-parameter $\lambda$ that influences the results: small values for $\lambda$ mean that the prediction won't change much and hence lead to small ICS scores for all values, while large values for $\lambda$ mean that predictions can be changed so much that all directions are considered as `relevant'. We leave the exploration of the tuning of this hyperparameter for future work. 

Finally, while some baselines `remove' the concept, this is not equivalent to estimating the causal effect of the concept on the model's output, as suggested by \cite{Goyal2019}. They propose to estimate the causal effect of a concept by generating counterfactuals with a conditional VAE to emulate the \textit{do} operator. While the concept-forgetting baseline is the closest activation displaying as much opposing concept as there is concept in the original activation (for $\lambda = 2(a^Tv_C+b)/||v_C||_2^2$), there is no guarantee that there exists a corresponding counterfactual input mapping to this baseline, nor that this input is a good approximation for the \textit{do} operation. In our experiments, we observed that it is possible to find a visually identical input mapping to the symmetric activations (the concept prediction is flipped), with the model's prediction being unchanged.

\subsection{Analytical forms of ICS}
\label{app:closed_form}

In some cases, it is possible to derive an analytical formula for ICS. We present the derivations of the formulations proposed in the main text below, using the notation defined in the main text.

\subsubsection{Last layer of a binary classification model with entropy-maximizing baseline}

In the binary case, we have $$h : \mathbf{a} \rightarrow \sigma(\mathbf{w}^T\mathbf{a}+b)$$ where $\sigma$ is the sigmoid function and $\mathbf{w}^T\mathbf{a}+b$ represents the logits. Therefore, $$\nabla_{\mathbf{v_C}} h(\mathbf{a}) = \sigma'(\mathbf{w}^T\mathbf{a}+b) \mathbf{w}^T\mathbf{v_C}$$ where  $\sigma'$ is the derivative of $\sigma$. Note that the $k$ index is dropped since the model has a single output.

We select a baseline $\mathbf{a'}$ that maximizes the entropy of the prediction, as suggested in \cite{Sundararajan2017}. This means that $$h(\mathbf{a'}) = 0.5$$ and $\mathbf{a'}$ is the orthogonal projection of $\mathbf{a}$ on the decision boundary $\mathbf{w}^T\mathbf{x}+b = 0$, i.e. $$\mathbf{a'}:= \mathbf{a} - (\mathbf{w}^T\mathbf{a}+b)\frac{\mathbf{w}}{||\mathbf{w}||_2^2}$$

With the change of variable $u := \mathbf{w}^T\mathbf{a'} + b + \alpha \mathbf{w}^T(\mathbf{a-a'}) = \alpha \mathbf{w}^T(\mathbf{a-a'})$ in the ICS formulation, we obtain:


\begin{align}
    \mathrm{ICS}_{C}(\mathbf{a},\mathbf{a'})  &= \mathbf{v_C}^T(\mathbf{a}-\mathbf{a'})\int_{0}^{\mathbf{w}^T\mathbf{a}+b}\sigma'(u)\frac{\mathbf{w}^T\mathbf{v_C}}{\mathbf{w}^T(\mathbf{a-a'})}du \nonumber \\ 
    &= \frac{\mathbf{v_C}^T(\mathbf{a-a'})\mathbf{w}^T\mathbf{v_C}}{\mathbf{w}^T(\mathbf{a-a'})}\bigg(\sigma(\mathbf{w}^T\mathbf{a}+b) - 0.5\bigg)
    \label{eq:ics_last_layer}
\end{align}

Noting that $\mathbf{a-a'} = (\mathbf{w}^T\mathbf{a}+b)\frac{\mathbf{w}}{||\mathbf{w}||_2^2}$, Eq. \ref{eq:ics_last_layer} can be rewritten as:

\begin{align*}
    \mathrm{ICS}_{C}(\mathbf{a},\mathbf{a'})  &= \frac{\mathbf{v_C}^T\mathbf{w}(\mathbf{w}^T\mathbf{a}+b)\mathbf{w}^T\mathbf{v_C}}{||\mathbf{w}||_2^2(\mathbf{w}^T\mathbf{a}+b)}\bigg(\sigma(\mathbf{w}^T\mathbf{a}+b) - 0.5\bigg) \\
    &= \bigg(\frac{\mathbf{v_C}^T\mathbf{w}}{||\mathbf{w}||_2}\bigg)^2\bigg(\sigma(\mathbf{w}^T\mathbf{a}+b) - 0.5\bigg)
\end{align*}



The conceptual sensitivity in that situation is given by $\mathrm{CS}_{C} = \mathbf{w}^T\mathbf{v_C}$. In contrast, ICS depends on the predicted probability for the considered input while not depending on the norm of $\mathbf{w}$, hence alleviating the aforementioned concerns. Note that the dependency on the square of the cosine similarity between the CAV and the model's last layer weights means that a poor estimate of $\mathbf{v_C}$ could result in a dramatically small value.

\subsubsection{Multi-class model, with concept-forgetting baseline}

Let's consider a multi-class model (i.e. $h_k$ is the $k$-th output of the softmax), with a concept-occluding baseline.  In this section, $\mathbf{v_C}$ is not assumed unit-normed. This baseline is the orthogonal projection on the hyperplane orthogonal to the CAV, i.e., assuming that the bias is 0 for compactness, it is defined by 

$$\mathbf{a'} := \mathbf{a} - \lambda\mathbf{v_C},\ \lambda \in \mathbb{R}$$

For $\lambda = 2$, it is akin to occlusion~\cite{Zeiler2014} for concepts.

Let
$$\mathcal{B}:=(\mathbf{e_1}, \mathbf{e_2}, ..., \mathbf{e_d})$$
be an orthonormal basis of $\mathbb{R}^d$ where $\mathbf{e_1}$ is chosen to be $\mathbf{v_C} \in \mathbb{R}^d$. We define $\mathrm{IG}_{i}^k(\mathbf{a},\mathbf{a'})$ as the integrated gradient for the $i$-th feature. Similarly, $\mathrm{ICS}_C^k(\mathbf{a}, \mathbf{a'})$ designates the ICS. 

As mentioned in Related works, integrated gradients verify completeness (Eq.\ref{eq:completeness}), which means that:
\begin{equation}
    F_k(\mathbf{x}) - F_k(\mathbf{x'}) = \sum_i \mathrm{IG}_i^k(\mathbf{x}, \mathbf{x'})
    \label{eq:completeness}
\end{equation}

By virtue of completeness, we have that:

\begin{equation}
    h_k(\mathbf{a}) - h_k(\mathbf{a'}) = \sum_{1\leq i \leq d} \mathrm{IG}_{i}^k(\mathbf{a}, \mathbf{a'})
    \label{eq:completeness_with_concept_occluding_baseline}
\end{equation}

Each $\mathrm{IG}_i^k$ is the product of two terms, one being the dot product between $\mathbf{a-a'}$ and $\mathbf{e_i}$. By definition, $\mathbf{a-a'}$ is colinear to $\mathbf{e_1}$ and $\mathcal{B}$ is orthonormal, therefore

$$\mathbf{(a-a')}^T\mathbf{e_i}=0,\ \forall i >1$$

Therefore, $\mathrm{IG}_i^k = 0 \ \forall i>1$ and Eq. \ref{eq:completeness_with_concept_occluding_baseline} becomes:
$$h_k(\mathbf{a}) - h_k(\mathbf{a'}) = \mathrm{IG}_{1}^k(\mathbf{a},\mathbf{a'})$$

Given $\mathbf{e_1} = \mathbf{v_C}$, ICS can be rewritten:
$$\mathrm{ICS}_C^k(\mathbf{a}, \mathbf{a'}) = \mathrm{IG_1^k}(\mathbf{a}, \mathbf{a'})$$

And by transitivity:

$$\mathrm{ICS}_C^k(\mathbf{a}, \mathbf{a'}) = h_k(\mathbf{a}) - h_k(\mathbf{a'})$$

i.e. ICS is equal to the difference in model probability between the model's prediction for this sample and the prediction should the concept be removed.

\subsection{Statistical procedure to compute confidence intervals on MCS}
\label{app:bootstrap_mcs}

\begin{enumerate}
    \item Sample $N$ iid images from the distribution.
    \item Compute the forward passes of $F_1$ and $F_2$ on these images to obtain two datasets of activations: $\mathcal{D}_i, \ \forall i \in \{1,2\}$
    \item For all models, layers, and concepts, compute $K$ unit CAVs by sampling with replacement $N$ samples from $\mathcal{D}_i$ to obtain $K$ bootstrapped datasets $\mathcal{D}_{i,k}^*, \ \forall k \in \{1, K\}$.
    \item Compute bootstrapped TCAV scores for all models, layers, and concepts using the trained CAVs. The values are computed on unseen data.
    \item Derive bootstrapped MCS scores as the differences between two bootstrapped TCAV scores
    \item Compute the 2.5th and 97.5th percentiles to obtain a 95\% nonparametric confidence interval.
\end{enumerate}

\section{Additional results on BARS}
\subsection{MLP model $F_o$ does not rely on the bar's color for its predictions}
\label{app:MLP_color}

In the BARS dataset, it is possible to evaluate how much the model relies on a given concept by evaluating the model on counterfactual examples \cite{Goyal2019}. We define $g_C$ as follows:

\begin{equation}
    g_C(\mathbf{x}) := \mathrm{max}_{(C_1, C_2) \in \mathcal{C}^2} |F(\mathbf{x}_{C_1}) - F(\mathbf{x}_{C_2})|
    \label{eq:local_gc}
\end{equation}

where $x_C$ is a counterfactual version of $x$ where its concept is set to $C$, and $\mathcal{C}$ is the set of possible values for the concept of interest. In the present case, $C_1$ and $C_2$ represent the color and orientation concepts.

This quantity can be aggregated over the entire test set to obtain the global influence of a concept on the model's output:

\begin{equation}
    G_C := \mathbb{E}_\mathbf{x}\big(g_C(\mathbf{x})\big)
    \label{eq:gc}
\end{equation}

We verify that $F_o$ is almost invariant to the color of the bar: on the test set, we obtain $G_\mathcal{C}(F_o) = 0.0011$. 

\begin{figure*}[!th]
    \centering
    \includegraphics[width=0.8\textwidth]{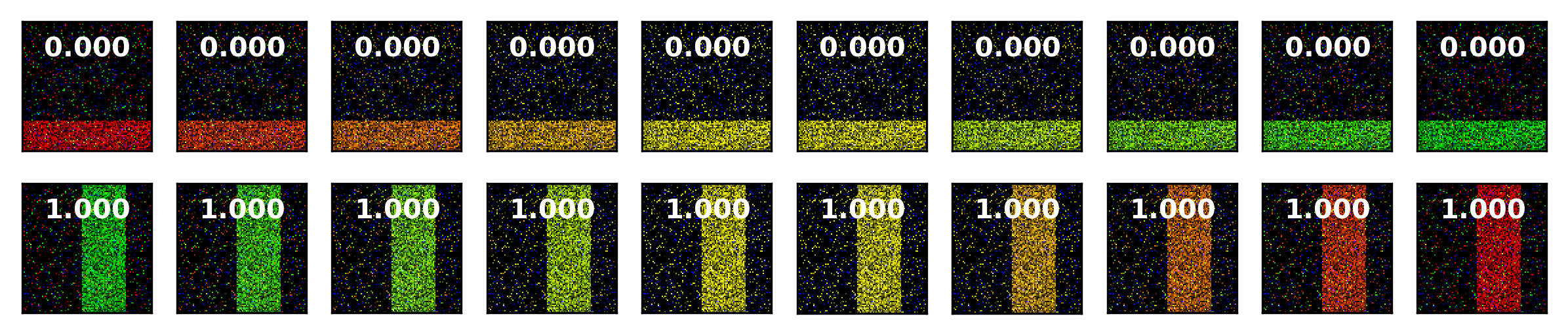}
    \caption{Vertical and horizontal bars with varying colors and the associated predicted probabilities ($F_o$) displayed.}
    \label{fig:invariance_to_colour}
\end{figure*}

Figure \ref{fig:invariance_to_colour} shows the output of the model on two random samples (vertical and horizontal) when slowly changing the color from green to red.

\subsection{Classification performance of $F_o$ and $F_c$}
The BARS dataset being very simple, our models (trained to detect either the color or the orientation) reach 100\% test accuracy after only a few epochs.

\subsection{Statistical significance and classification performance of trained CAVs}
\label{app:cavs_synth}

Table \ref{tab:cav_perf_bars} displays the ROCAUC on held-out data (mean and standard deviation across bootstraps) of the CAVs for both concepts, and both models.

\begin{table}[!ht]
\centering
\begin{tabular}{|l|l|l|l|l|}
\hline
      & \multicolumn{2}{l|}{$F_o$} & \multicolumn{2}{l|}{$F_c$} \\ \hline
layer & Color      & Orientation   & Color      & Orientation   \\ \hline
1     & 100 (0)    & 100 (0)       & 100 (0)    & 59.1 (5.6)    \\ \hline
2     & 100 (0)    & 100 (0)       & 100 (0)    & 54.0 (4.6)    \\ \hline
3     & 100 (0)    & 100 (0)       & 100 (0)    & 50.8 (3.7)    \\ \hline
\end{tabular}
\caption{ROCAUC of the bootstrapped CAVs. Average value rounded to closest decimal, and standard deviation in parentheses.}
\label{tab:cav_perf_bars}
\end{table}

Note that while the color concept is not being used by $F_o$, its CAVs have perfect discriminative performance for identifying that concept in the activation spaces. However, the orientation concept is impossible to identify with a linear classifier in the activation spaces of $F_c$.

\subsection{Influence of the baselines}
\label{app:BARS_baseline}

\begin{figure}[!t]
\centering
\begin{subfigure}[t]{0.01\textwidth}
\textbf{a}
\end{subfigure}
\begin{subfigure}[t]{0.475\textwidth}
\includegraphics[width=\linewidth,valign=t]{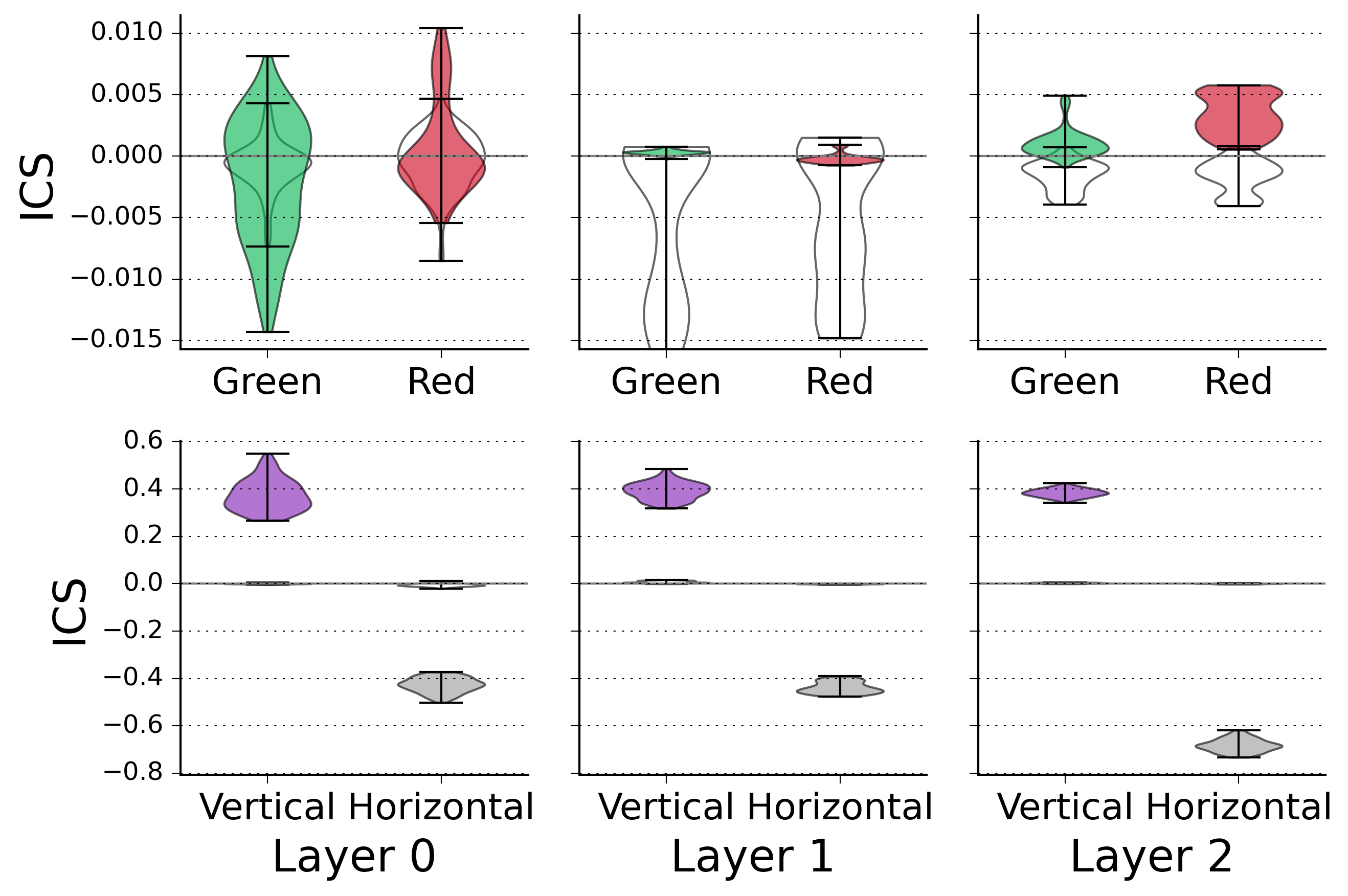} 
\end{subfigure}
\begin{subfigure}[t]{0.01\textwidth}
\textbf{b}
\end{subfigure}
\begin{subfigure}[t]{0.475\textwidth}
\includegraphics[width=\linewidth,valign=t]{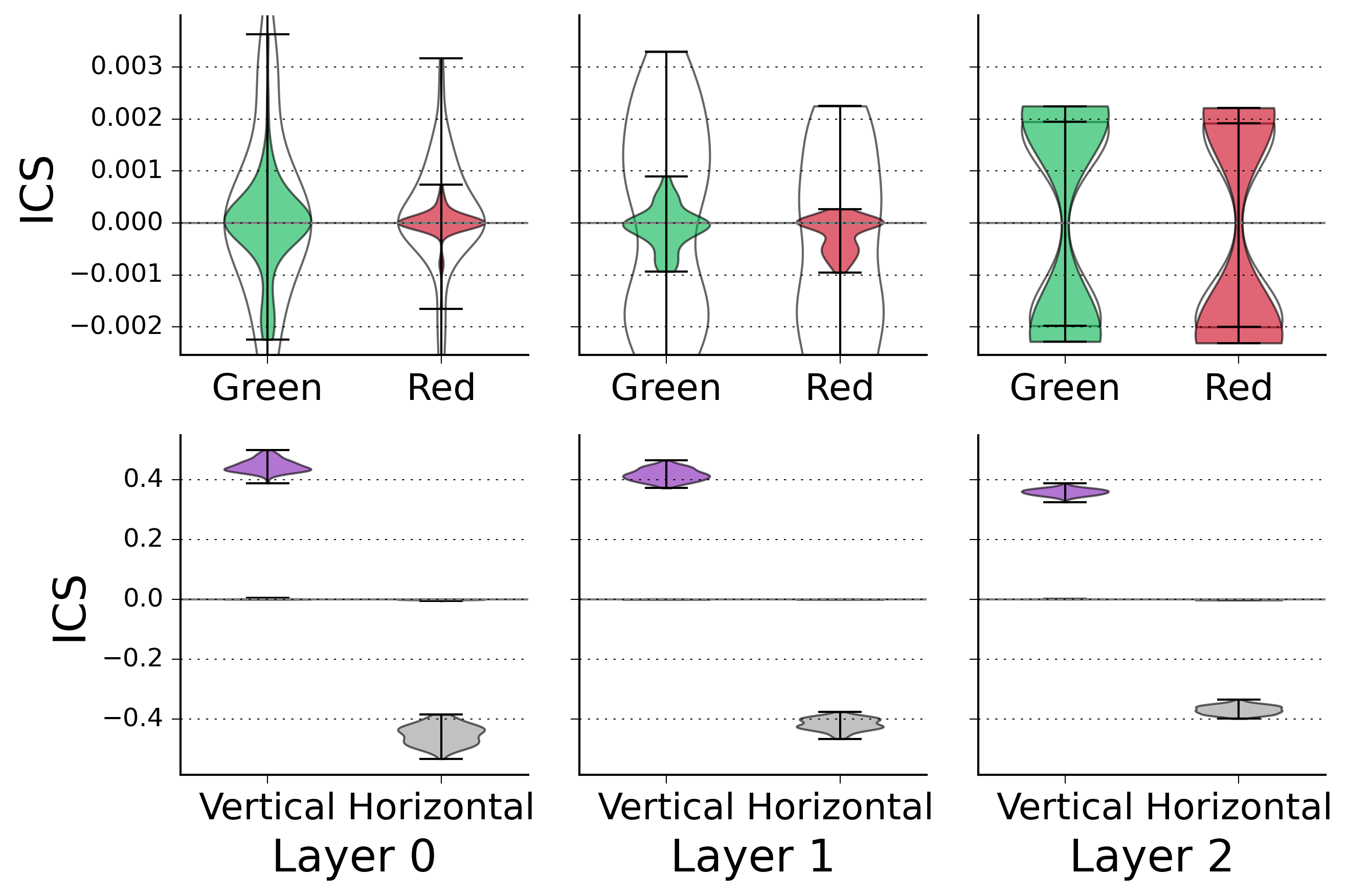} 
\end{subfigure}
\begin{subfigure}[t]{0.01\textwidth}
\textbf{c}
\end{subfigure}
\begin{subfigure}[t]{0.475\textwidth}
\includegraphics[width=\linewidth,valign=t]{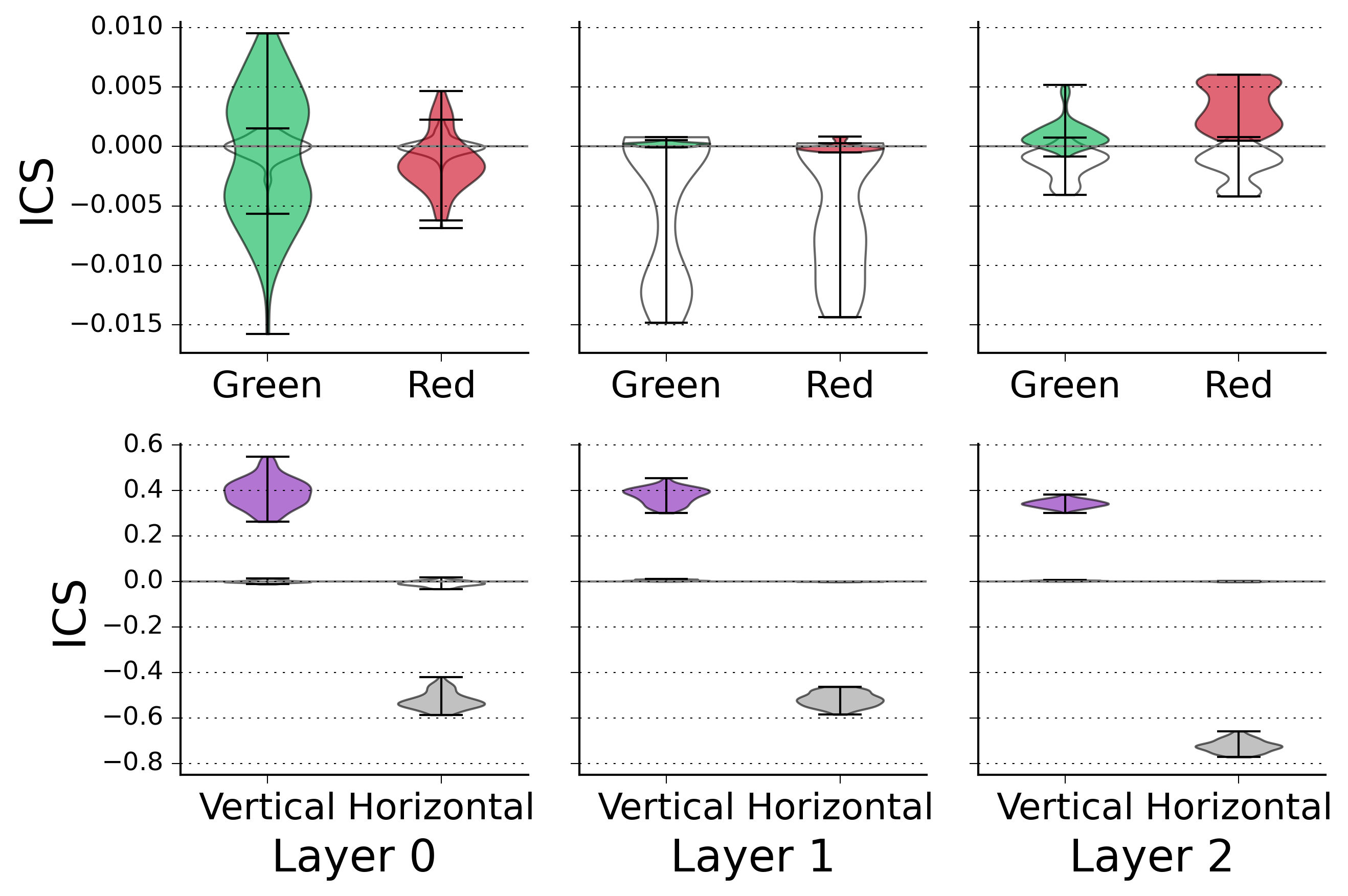} 
\end{subfigure}
\begin{subfigure}[t]{0.01\textwidth}
\textbf{d}
\end{subfigure}
\begin{subfigure}[t]{0.475\textwidth}
\includegraphics[width=\linewidth,valign=t]{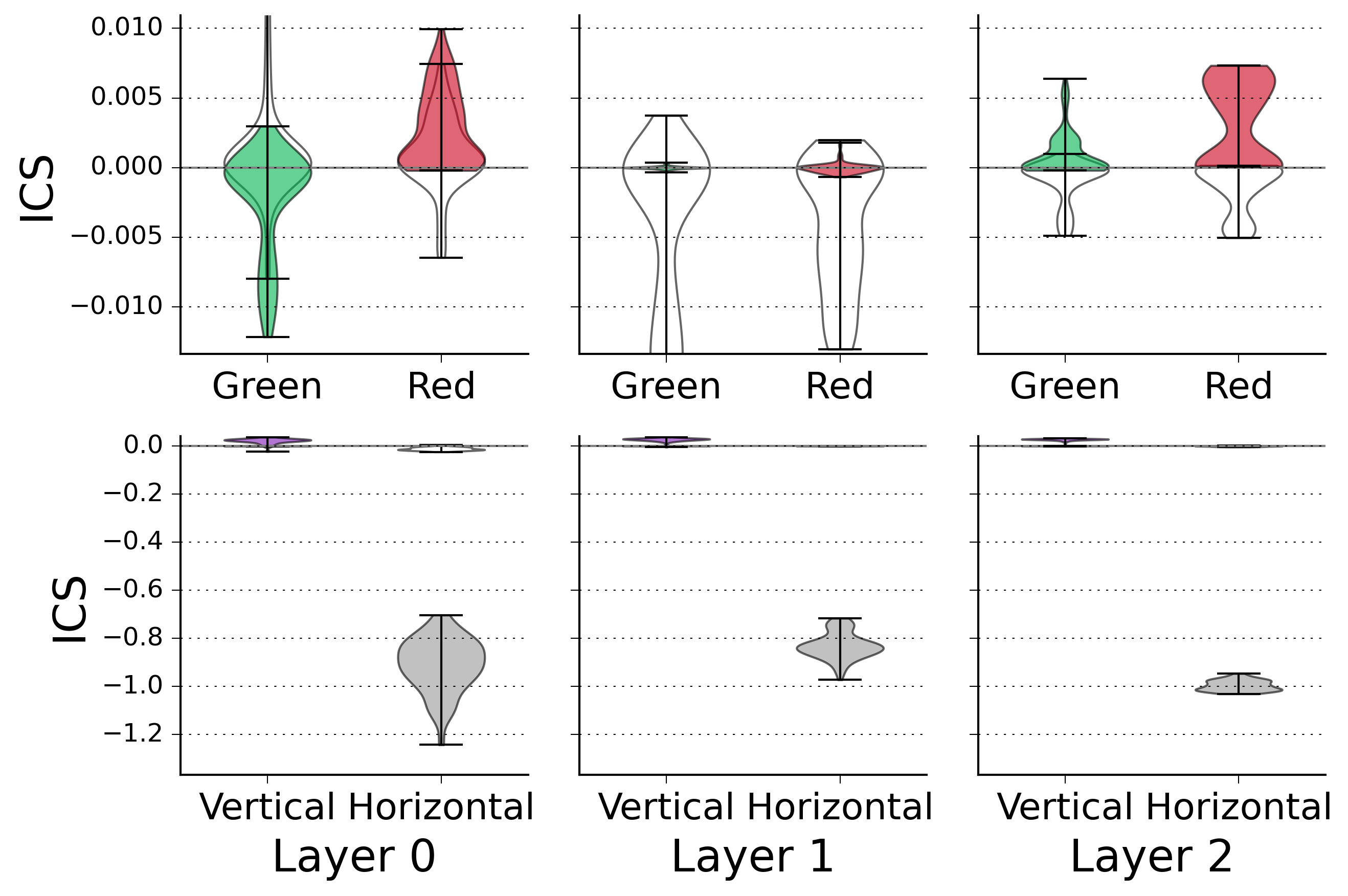} 
\end{subfigure} \\
\begin{subfigure}[t]{0.01\textwidth}
\textbf{e}
\end{subfigure}
\begin{subfigure}[t]{0.475\textwidth}
\includegraphics[width=\linewidth,valign=t]{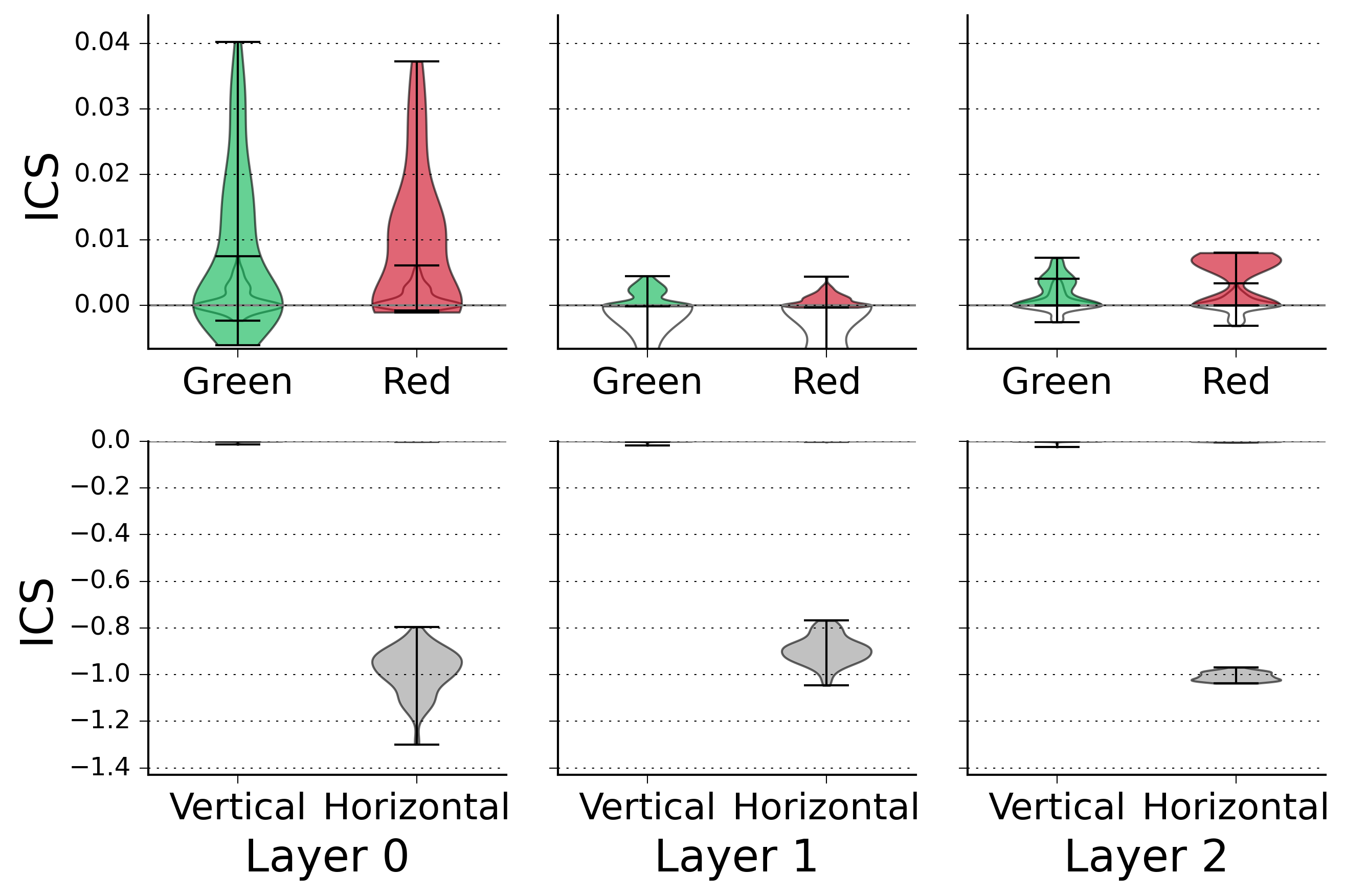} 
\end{subfigure}
\begin{subfigure}[t]{0.01\textwidth}
\textbf{f}
\end{subfigure}
\begin{subfigure}[t]{0.475\textwidth}
\includegraphics[width=\linewidth,valign=t]{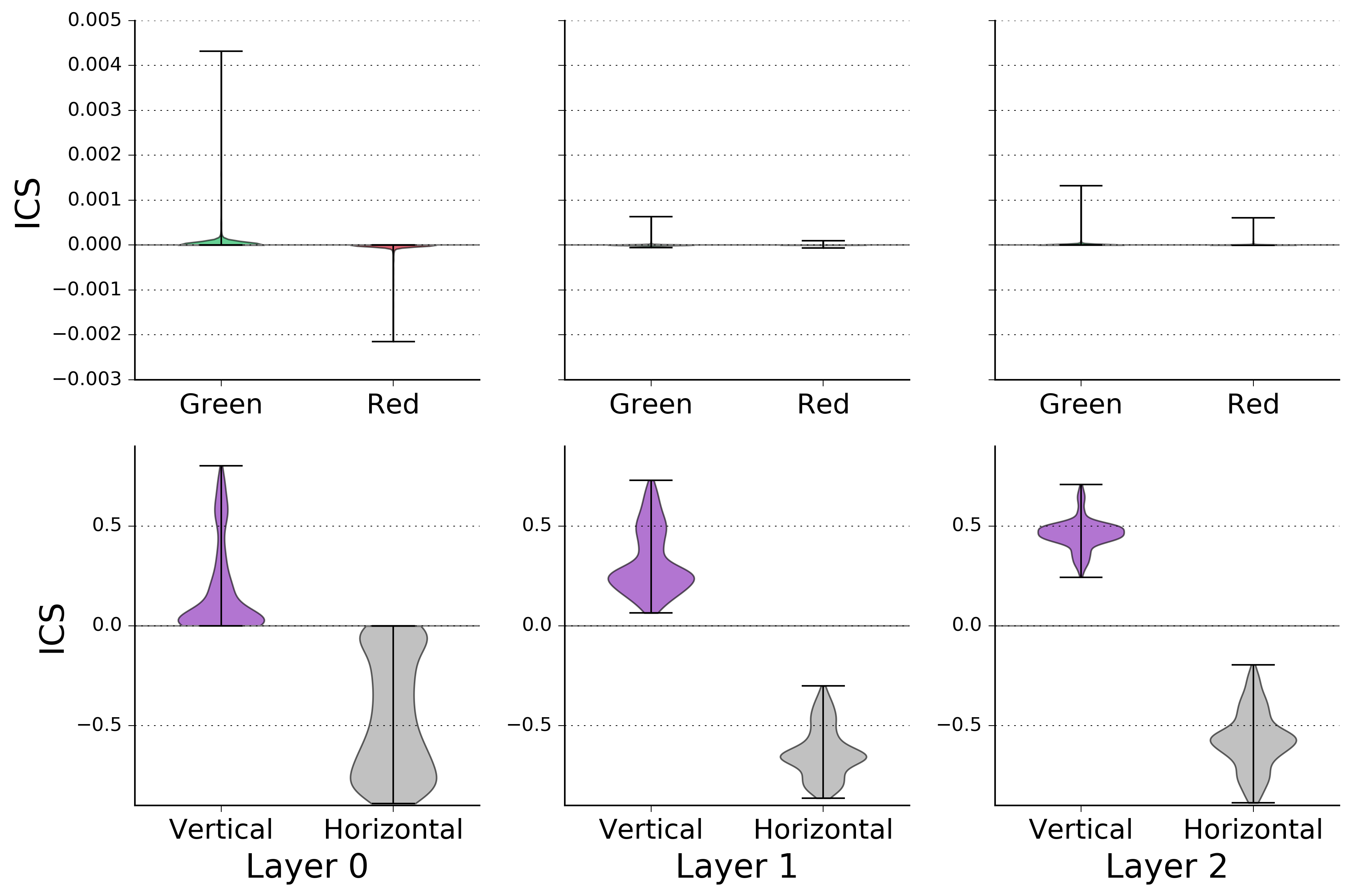} 
\end{subfigure} \\
\caption{\textbf{Distributions of ICS for $F_o^\mathrm{BARS}$. a} Black baseline. \textbf{b} Entropy-maximizing baseline. \textbf{c} Pixel-wise average baseline. \textbf{d} Noisy baseline, i.e. the baseline is a sample of $\mathcal{N}(0,1)$. \textbf{e} White image baseline. \textbf{f} Concept-occluding baseline.}
\label{fig:BARS_baseline}
\end{figure}

In the BARS dataset, we computed the predicted probability associated with all baselines (see Table~\ref{tab:predicted_risk_on_baselines} for a selected sample).  Pixel-wise average and entropy-maximizing baselines seem to be truly uninformative: the output of the model is a tie. This results in bimodal ICS distributions clustering around -50\% and 50\% (Figure~\ref{fig:BARS_baseline}a,b).

If uninformative means having a 50\% predicted risk, then it is clear that some of these baselines are not uninformative. The iid Gaussian noise $\mathcal{N}(0,1)$ in the pixel space leads to a bimodal distribution that has two modes in 0 and in 1, meaning that this seemingly meaningless noise is interpreted very confidently by the model. Similarly, the white baseline leads to a very confident prediction of `vertical' for $F_o$.

In terms of ICS, we observe that different `informative' baselines (as interpreted by the model) lead to different supports for ICS. The difference across concepts, hence MCS, is however relatively consistent across baselines (Table~\ref{tab:mcs_bars}).

\begin{table}[!ht]
\centering
\begin{tabular}{|l|l|l|l|}
\hline
                                                & layer 1      & layer 2      & layer 3      \\ \hline
i.i.d. $\mathcal{N}(0,1)$ pixels & 60\% (41\%)  & 60\% (41\%)  & 60\% (41\%)  \\ \hline
Black image (zero)                                           & 62\%         & 62\%         & 62\%         \\ \hline
Average activation                 & 1\%         & 1\%         & 6\%         \\ \hline
Average pixel-wise                      & 66\%         & 66\%         & 66\%         \\ \hline
Entropy-maximizing                              & 50\% (0.1\%) & 50\% (0.1\%) & 50\% (0.1\%) \\ \hline
 White image (zero) & 100\% & 100\%  & 100\% \\ \hline
\end{tabular}
\caption{Predicted probability of the bar being vertical for several baselines. Standard deviation is provided in parenthesis when it is not 0.}
\label{tab:predicted_risk_on_baselines}
\end{table}

\begin{table}[!h]
\centering
\begin{tabular}{|l|l|l|l|l|}
\hline
concept     & layer & $\mathrm{MCS}(\mathrm{TCAV}^{sign(\mathrm{CS})})$ & $\mathrm{MCS}(\mathrm{TCAV}^\mathrm{ICS})$ black &  $\mathrm{MCS}(\mathrm{TCAV}^\mathrm{ICS})$ max ent. \\ \hline
orientation & 0     & 0.00 [0.00, 1.00]     &    0.36 [0.31, 0.41] & 0.46 [0.43, 0.47] \\ \hline
orientation & 1     & 0.00  [0.00, 1.00]    &    0.33 [0.27, 0.39] & 0.47 [0.45, 0.48]   \\ \hline
orientation & 2     & 1.00 [0.00, 1.00]      &    0.42 [0.37, 0.48] & 0.35 [0.34, 0.36]  \\ \hline
color      & 0     & 0.93 [0.00, 1.00]      &    0.41 [0.33, 0.47] & 0.47 [0.45, 0.47]   \\ \hline
color      & 1     & 0.00  [0.00, 0.00]     &    0.45 [0.38, 0.50]  & 0.36 [0.33, 0.37] \\ \hline
color      & 2     & 1.00 [0.00, 1.00]       &    0.56 [0.46, 0.64] & 0.34 [0.33, 0.34]  \\ \hline
\end{tabular}
\caption{Median MCS scores (\%) computed on the entire BARS test set, with relevant 95\% bootstrapped non-parametric confidence intervals.}
\label{tab:mcs_bars}
\end{table}

\section{Additional results on BAM}
\label{app:BAM_section}

\subsection{$F_o^\mathrm{BAM}$ does not rely on the scene background, and $F_s^\mathrm{BAM}$ does not rely on the pasted object}
Using a similar approach \cite{Yang2019} showed that the accuracy of the models is no better than a random guess when the relevant concept are removed from the images.

\subsection{Classification performance of $F_o$ and $F_s$}
For BAM, we used pre-trained ResNet50 \cite{He2015} (74.9\% top-1 accuracy) and EfficientNet-B3 \cite{Tan2020} (81.6\% top-1 accuracy) available at \url{www.tensorflow.org}. They were fine-tuned on the BAM dataset to predict either scenes (reaching ~91\% test top-1 accuracy) or objects (reaching ~80\% test top-1 accuracy) by iteratively updating the weights of the final dense layers, followed by convolutional stacks, one after the other with Adam optimizer (learning rate $10^{-4}$) \cite{Kingma2014} until convergence of the validation accuracy (about 1-5 epochs). Training is performed with classical data augmentations (random flips, contrast, brightness, hue, saturation changes). Standard pre-processing techniques were applied to the images (see \url{https://www.tensorflow.org/api_docs/python/tf/keras/applications/imagenet_utils/preprocess_input}).

\subsection{Statistical significance and classification performance of trained CAVs}
\label{app:bam_cavs}

\begin{figure}[!ht]
\centering
\begin{subfigure}[t]{0.01\textwidth}
\textbf{a}
\end{subfigure}
\begin{subfigure}[t]{0.475\textwidth}
\includegraphics[width=\linewidth,valign=t]{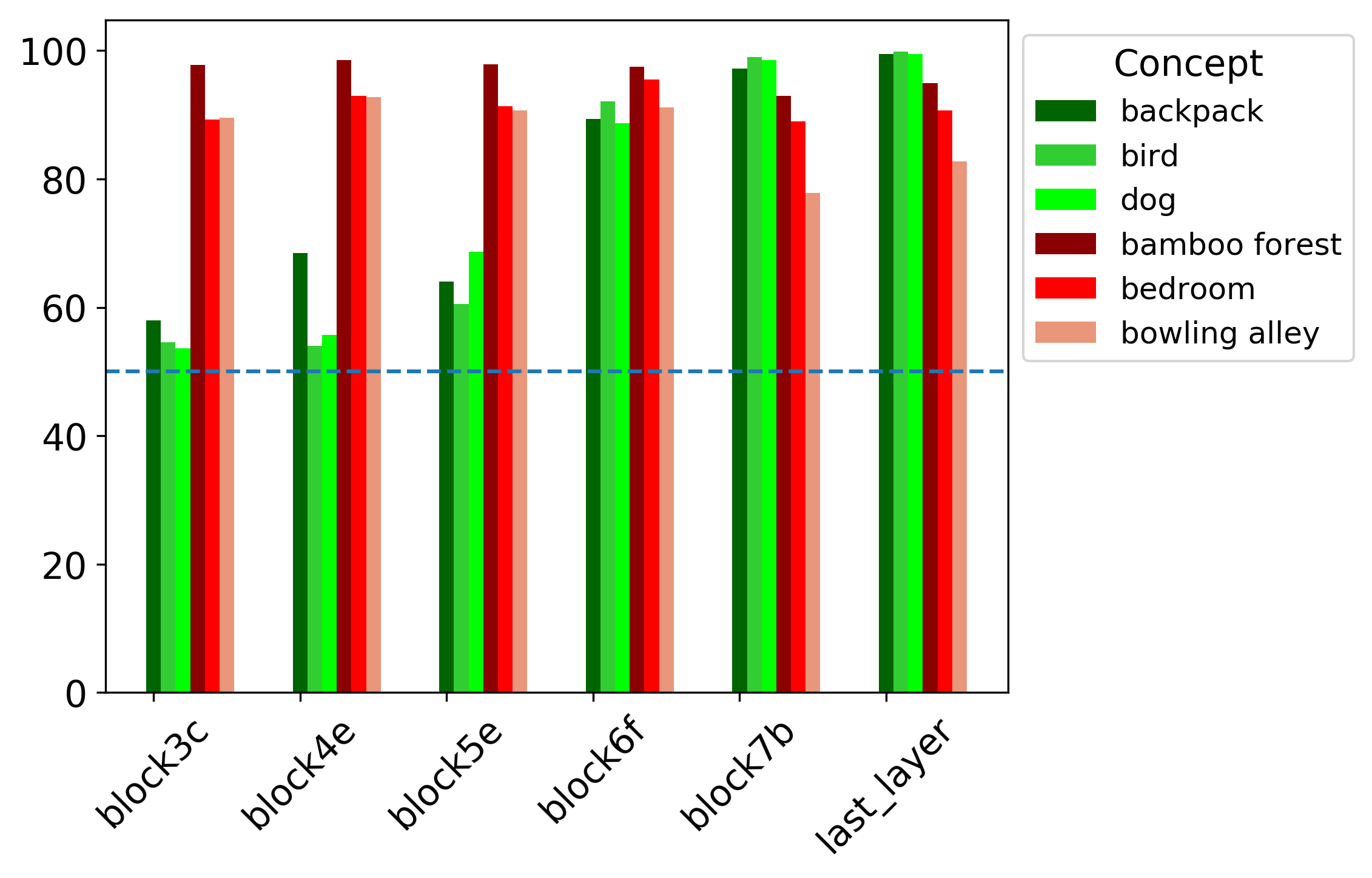} 
\end{subfigure}
\begin{subfigure}[t]{0.01\textwidth}
\textbf{b}
\end{subfigure}
\begin{subfigure}[t]{0.475\textwidth}
\includegraphics[width=\linewidth,valign=t]{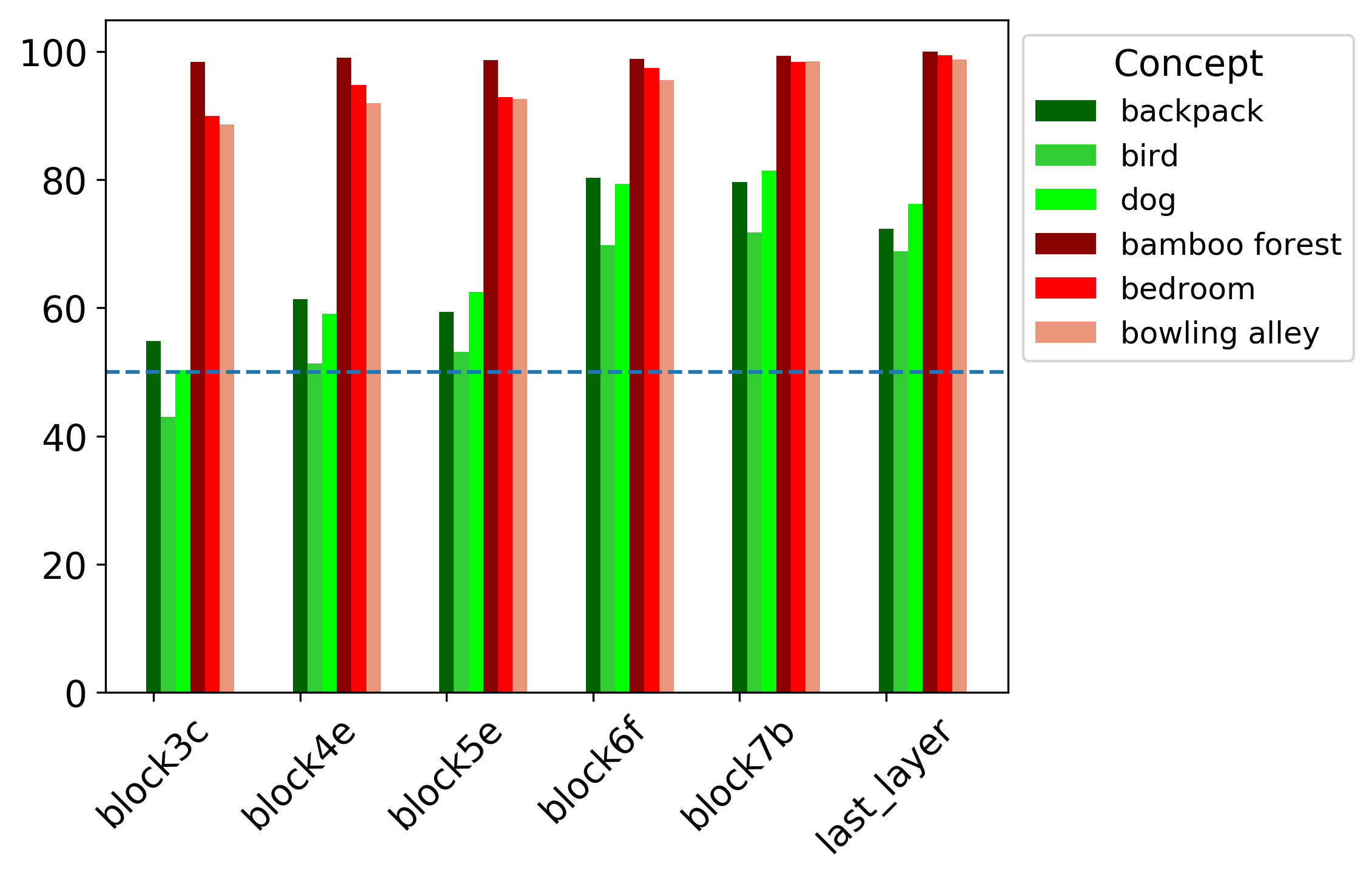}
\end{subfigure}
\caption{ROCAUC of EfficientNet's CAVs for 6 concepts, estimated on heldout data. \textbf{a.} For model $F_o$. \textbf{b.} For model $F_s$.}
\label{fig:perf_cavs_efficientnet}
\end{figure}

We trained CAVs using ElasticNet regularization \cite{Zou05}, which resulted in sparse weights and best classification performance on held-out samples (compared to Ridge and Lasso regression). Regularization coefficient is estimated via cross-validation. CAVs are not all statistically significant, as assessed by a permutation test. Figure \ref{fig:perf_cavs_efficientnet} displays the predictive performance of CAVs for both EfficientNet-B3 models, 6 layers, and 6 concepts. Shallow CAVs, which weren't fine-tuned and kept Imagenet-learned weights, tend to be very good at identifying scenes for both models. They however cannot identify objects. Deeper CAVs keep a high discriminative performance for scenes for both $F_o$ and $F_s$. Only deep CAVs of $F_o$ reach high discriminative performance for objects. We observed similar results for ResNet50.

\subsection{Influence of the baseline}
\label{app:BAM_baseline}

While the choice of the baseline mostly affected the support of ICS for BARS, we observe that the scale of the results can be affected for BAM, whereby small values of ICS are observed for the relevant concepts. This phenomenon is more pronounced for shallow layers of the models (Table \ref{tab:mcs_bam}), and is variable across baselines (Table \ref{tab:mcs_bam_other_baselines}). We hypothesize that these results are due to (the combination of) two factors: the higher dimensionality of the CAVs, and a deterioration in the quality of the projection of IG onto the CAV (variable across baselines). We explore the former in section \ref{app:curse_of_dimensionality}.


\begin{table*}[!ht]
\centering
\begin{tabular}{|l|l|l|l|}
\hline
concept       & layer & $\mathrm{MCS}(\mathrm{TCAV}^{sign(\mathrm{CS})})$ (\%) & $\mathrm{MCS}(\mathrm{TCAV}^\mathrm{ICS})$ (\%) \\ \hline
backpack      & conv3 & 0.21 (-0.11, 0.61)                                      & -0.03 (-0.24, 0.17)                                 \\ \hline
bird          & conv3 & 0.15 (-0.26, 0.47)                                    & 0.09 (0.01, 0.21)                                 \\ \hline
dog           & conv3 & 0.04 (-0.36, 0.51)                                     & -0.00 (-0.19, 0.23)                                \\ \hline
bamboo forest & conv3 & 0.36 (0.08, 0.62)                                     & 0.25 (-0.01, 0.54)                                \\ \hline
bedroom       & conv3 & 0.55 (0.32, 0.69)                                      & 0.10 (-0.08, 0.29)                                 \\ \hline
bowling alley & conv3 & 0.29 (-0.08, 0.52)                                      & 0.11 (-0.14, 0.33)                                \\ \hline
backpack      & conv4 & 0.42 (-0.19, 0.86)                                       & 0.09 (-0.03, 0.24)                                 \\ \hline
bird          & conv4 & 0.70 (-0.02, 0.96)                                      & 0.06 (-0.07, 0.19)                                 \\ \hline
dog           & conv4 & 0.37 (-0.02, 0.91)                                               & 0.07 (-0.07, 0.21)                               \\ \hline
bamboo forest & conv4 & 0.52 (0.17, 0.76)                                      & 0.53 (0.35, 0.68)                               \\ \hline
bedroom       & conv4 & 0.42 (0.17, 0.75)                                      & 0.44 (0.17, 0.68)                                 \\ \hline
bowling alley & conv4 & 0.56 (0.26, 0.86)                                      & 0.31 (0.08, 0.57)                                \\ \hline
backpack      & conv5 & 1.00 (-0.00, 1.00)                                         & 0.61 (0.45, 0.71)                               \\ \hline
bird          & conv5 & 1.00 (-0.00, 1.00)                                           & 0.49 (0.30, 0.62)                               \\ \hline
dog           & conv5 & 1.00 (-0.00, 1.00)                                               & 0.48 (0.36, 0.65)                               \\ \hline
bamboo forest & conv5 & 0.00 (0.00, 1.00)                                        & 0.75 (0.61, 0.86)                               \\ \hline
bedroom       & conv5 & 0.85 (0.00, 1.00)                                          & 0.72 (0.57, 0.84)                               \\ \hline
bowling alley & conv5 & 1.00 (0.00, 1.00)                                         & 0.60 (0.45, 0.74)                               \\ \hline
backpack      & last  & -0.00 (-0.00, 1.00)                                         & 0.75 (0.70, 0.79)                               \\ \hline
bird          & last  & 1.00 (-0.00, 1.00)                                         & 0.73 (0.66, 0.77)                               \\ \hline
dog           & last  & 1.00 (-0.00, 1.00)                                         & 0.75 (0.70, 0.79)                               \\ \hline
bamboo forest & last  & 0.00 (0.00, 0.89)                                       & 0.92 (0.90, 0.94)                               \\ \hline
bedroom       & last  & 0.00 (0.00, 0.07)                                         & 0.90 (0.87, 0.92)                              \\ \hline
bowling alley & last  & 0.06 (0.00, 1.00)                                         & 0.83 (0.79, 0.87)                              \\ \hline
\end{tabular}
\caption{Median bootstrapped MCS scores for two concept-attribution methods: $\mathrm{TCAV}^{sign(\mathrm{CS})}$ and $\mathrm{TCAV}^\mathrm{ICS}$ with concept-forgetting baseline ($\lambda=100$). They were computed for 3 scene and 3 object concepts. 95\% nonparametric confidence intervals in parenthesis. conv3, conv4, conv5 designate the outputs of the stacks of 2d convolutions having respectively 512, 1024, and 2048 filters in the Resnet50 architecture (see Figure 3 in \cite{He2015}).}
\label{tab:mcs_bam}
\end{table*}

$\mathrm{TCAV}^\mathrm{ICS}$ leads to higher scores for deeper layers of the network, with narrower confidence intervals. On the other hand, MCS scores for $\mathrm{TCAV}^{sign(\mathrm{CS})}$ have wider confidence intervals, especially for deeper layers.

\begin{table*}[!ht]
\centering
\begin{tabular}{|l|l|l|l|}
\hline
concept       & layer & zero                   & entropy-maximizing        \\ \hline
backpack      & conv4 & 0.00 (-0.00, 0.00) & 0.00 (0.00, 0.00)   \\ \hline
bird          & conv4 & -0.00 (-0.00, 0.00) & 0.00 (0.00, 0.00) \\ \hline
dog           & conv4 & 0.00 (0.00, 0.00)     & 0.00 (0.00, 0.00)  \\ \hline
bamboo forest & conv4 & 0.02 (0.01, 0.03)      & 0.00 (0.00, 0.00)      \\ \hline
bedroom       & conv4 & 0.01 (0.01, 0.02)     & 0.00 (0.00, 0.00)      \\ \hline
bowling alley & conv4 & 0.01 (0.00, 0.02)     & 0.00 (0.00, 0.00)      \\ \hline
backpack      & conv5 & 0.08 (0.04, 0.11)        & 0.00 (0.00, 0.01)        \\ \hline
bird          & conv5 & 0.06 (0.03, 0.09)        & 0.00 (0.00, 0.01)         \\ \hline
dog           & conv5 & 0.06 (0.03, 0.08)        & 0.00 (0.00, 0.01)        \\ \hline
bamboo forest & conv5 & 0.04 (0.02, 0.06)      & 0.00 (0.00, 0.00)       \\ \hline
bedroom       & conv5 & 0.06 (0.03, 0.08)       & 0.00 (0.00, 0.00)       \\ \hline
bowling alley & conv5 & 0.06 (0.04, 0.08)       & 0.00 (0.00, 0.00)      \\ \hline
backpack      & last  & 0.18 (0.16, 0.21)       & 0.05 (0.04, 0.06)            \\ \hline
bird          & last  & 0.15 (0.12, 0.18)        & 0.06 (0.04, 0.07)            \\ \hline
dog           & last  & 0.17 (0.15, 0.20)       & 0.05 (0.03, 0.05)            \\ \hline
bamboo forest & last  & 0.21 (0.18, 0.23)       & 0.03 (0.02, 0.03)           \\ \hline
bedroom       & last  & 0.19 (0.16, 0.22)      & 0.04 (0.03, 0.05)            \\ \hline
bowling alley & last  & 0.19 (0.15, 0.22)       & 0.05 (0.04, 0.05)            \\ \hline
\end{tabular}
\caption{Median bootstrapped MCS for $\mathrm{TCAV}^\mathrm{ICS}$ with two baselines: zeros (black image) and entropy-maximizing. They were computed for 3 scene and 3 object concepts. 95\% nonparametric confidence intervals in parenthesis. conv3, conv4, conv5 designate the outputs of the stacks of 2d convolutions having respectively 512, 1024, and 2048 filters in the Resnet50 architecture, and last is the output of the final layer (before the softmax) (see Figure 3 in \cite{He2015}).}
\label{tab:mcs_bam_other_baselines}
\end{table*}


\subsection{Curse of dimensionality}
\label{app:curse_of_dimensionality}

We hypothesize that the divergence between the directionality of bootstrapped CAVs increases with the dimensionality $d$ of the space, or more accurately, with the decrease in the ratio $\frac{n}{d}$ where $n$ is the number of samples used to train the CAV. This increase in variance of the CAV directionality leads to smaller dot products (the average cosine similarity between two vectors scales with $d^{-\frac{1}{2}}$) and therefore ICS goes to 0. We verified on BARS that decreasing the $\frac{n}{d}$ ratio from 30 to 0.1 resulted in MCS scores 3 times smaller (~45\% vs ~15\%).

\begin{figure}[!ht]
\centering
\includegraphics[width=0.6\linewidth,valign=t]{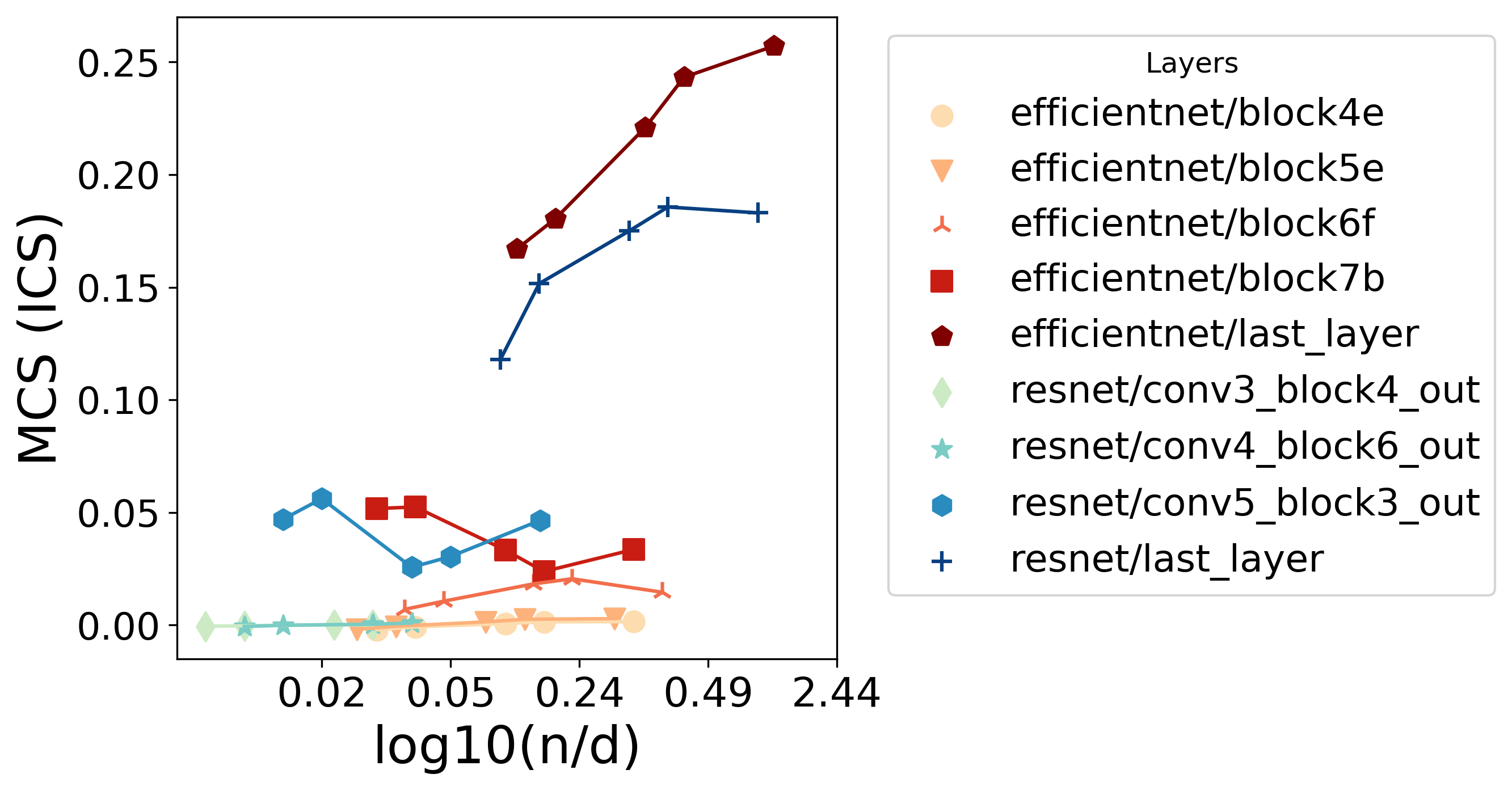}
\caption{\textbf{MCS with varying $n/d$ ratio.} Influence of $\frac{n}{d}$ ratio over MCS computed with concept ``bird'' for several layers of a EfficientNet-B3 model (in shades of orange to red) \cite{Tan2020} and a ResNet50 (in shades of blue to green) \cite{He2015} fine-tuned on BAM. Darker colors represent deeper layers. MCS(ICS) with black image baseline.}
\label{fig:nd_mcs}
\end{figure}

In this section, we evaluate different techniques that would lead to more consistent directions of the CAV. We first increase $n$ by augmenting the CAV training set with image transformations such as random flips and changes of contrast, brightness, hue, and saturation. The figure in the main text shows how increasing $\frac{n}{d}$ improved MCS for $\mathrm{TCAV}^{\mathrm{ICS}}$ using a concept-forgetting baseline. For the zero baseline, the MCS scores benefit only marginally from increased $\frac{n}{d}$ from data augmentation (Figure~\ref{fig:nd_mcs}). We also experimented with various regularization schemes for training CAVs, which did not lead to any improvement in terms of ICS score magnitude.

\section{ImageNet}
\label{app:imagenet}

\subsection{Image selection and CAV building}
\label{app:imagenet_cavs}
The first three concepts (striped, dotted and zigzagged) rely on the same images as used in \cite{Kim2017}. For the other concepts, we used Google search images, automatically downloaded using \url{https://github.com/hardikvasa/google-images-download} (MIT license). A manual review of those images was performed to discard images that included multiple concepts (e.g. ``hoop'' also displaying the wooden floor), or obvious confounders such as watermarks, drawings, captions, etc. We caveat the manual review of the Google search images, especially in the last 2 concepts as ``gender'' was assessed by the captions or web links of the images (e.g. ``Women League'') as well as physical appearance. For skin tone (referred to as the ``race'' concept), physical appearance was used. We also note that these images were directly scraped, and that no consent was obtained from e.g. basketball players. In a real-world use case of our method, these two aspects would need to be addressed with care, e.g. by obtaining consented images, with self-reported demographics information.

Our selection resulted in $\sim100$ images per concept. Some concepts were then built by distinguishing between the selected images and random images as defined in \cite{Kim2017}. For the ``race'' and ``gender'' concepts, we compute \emph{relative} concepts by contrasting pictures of men and women, and pictures of lighter skin tone and darker skin tone basketball players (in an attempt to control for clothing, background or occupation biases in our image search). 

\subsection{Global results for zebras}

\begin{figure}[!t]
    \centering
    \includegraphics[width=0.7\columnwidth]{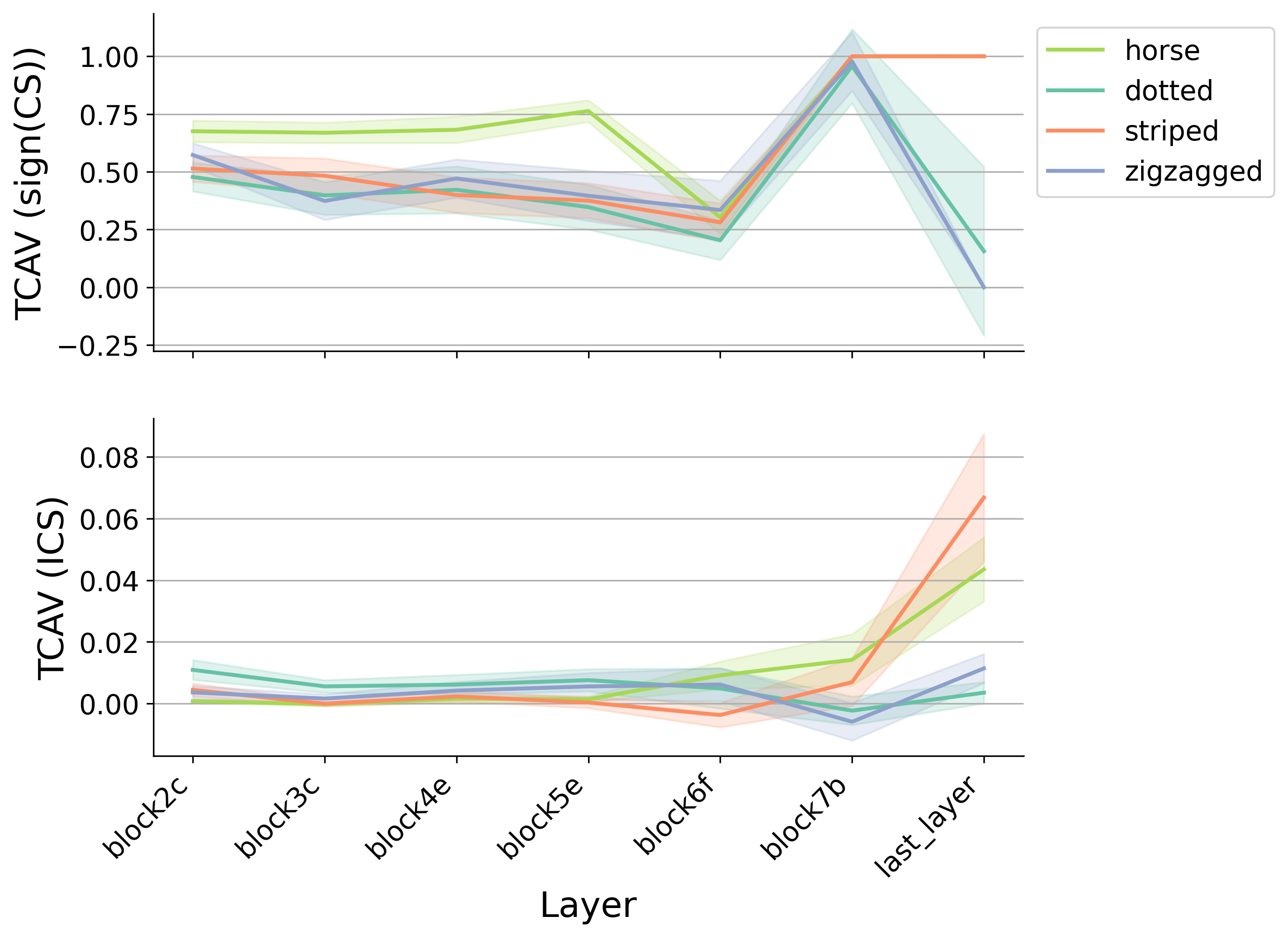}
    \caption{TCAV, CS and ICS scores of several concepts on the `zebra' output of an EfficientNet-B3 model, for 100 pictures of zebras, computed on all layers, with a white image baseline.}
    \label{fig:global_zebra_efficientnet}
\end{figure}

\begin{figure}[!ht]
    \centering
    \includegraphics[width=0.7\columnwidth]{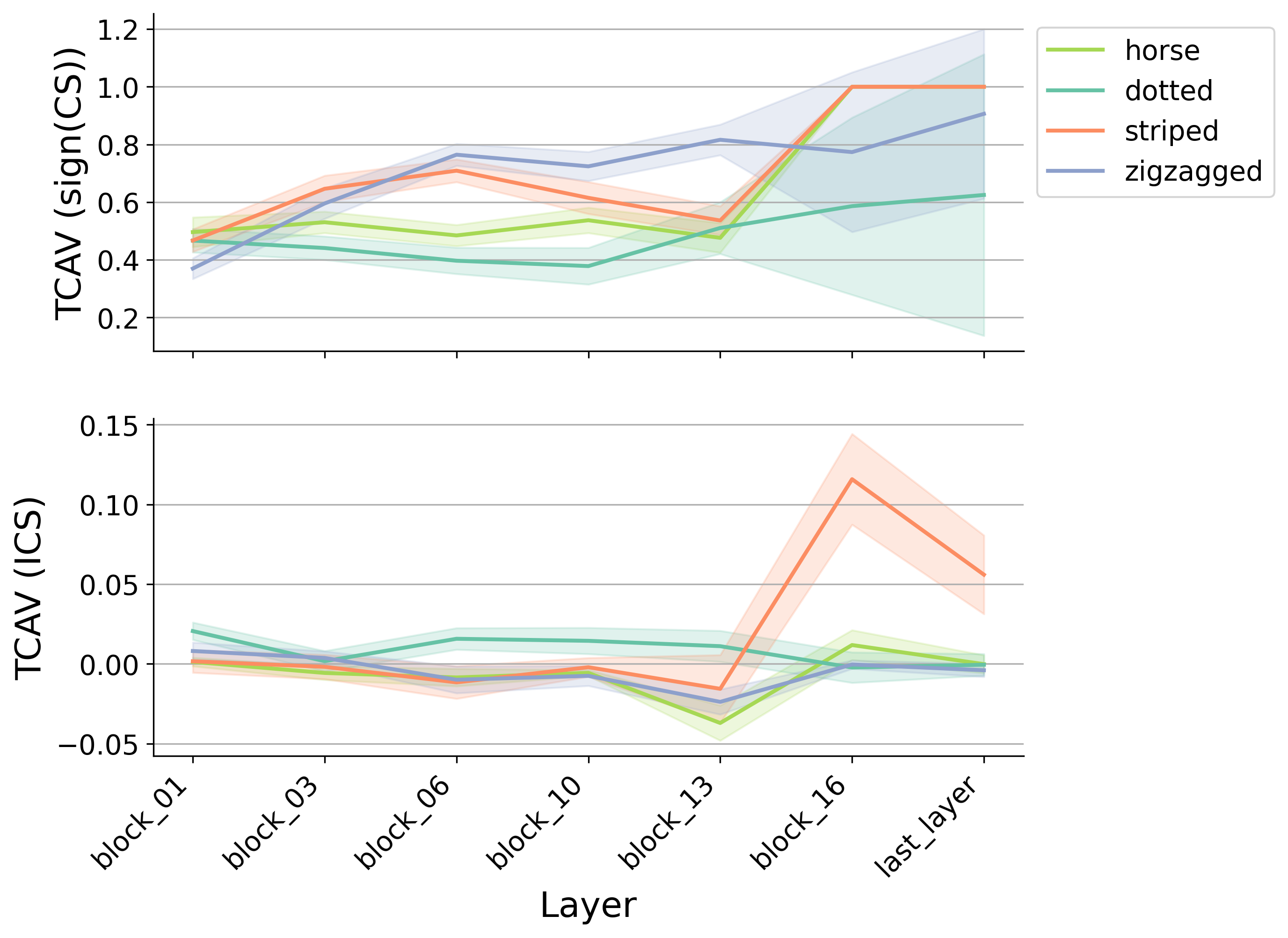}
    \caption{MobileNet V2.}
    \label{fig:global_zebra_mobilenet}
\end{figure}

\begin{figure}[!ht]
    \centering
    \includegraphics[width=0.7\columnwidth]{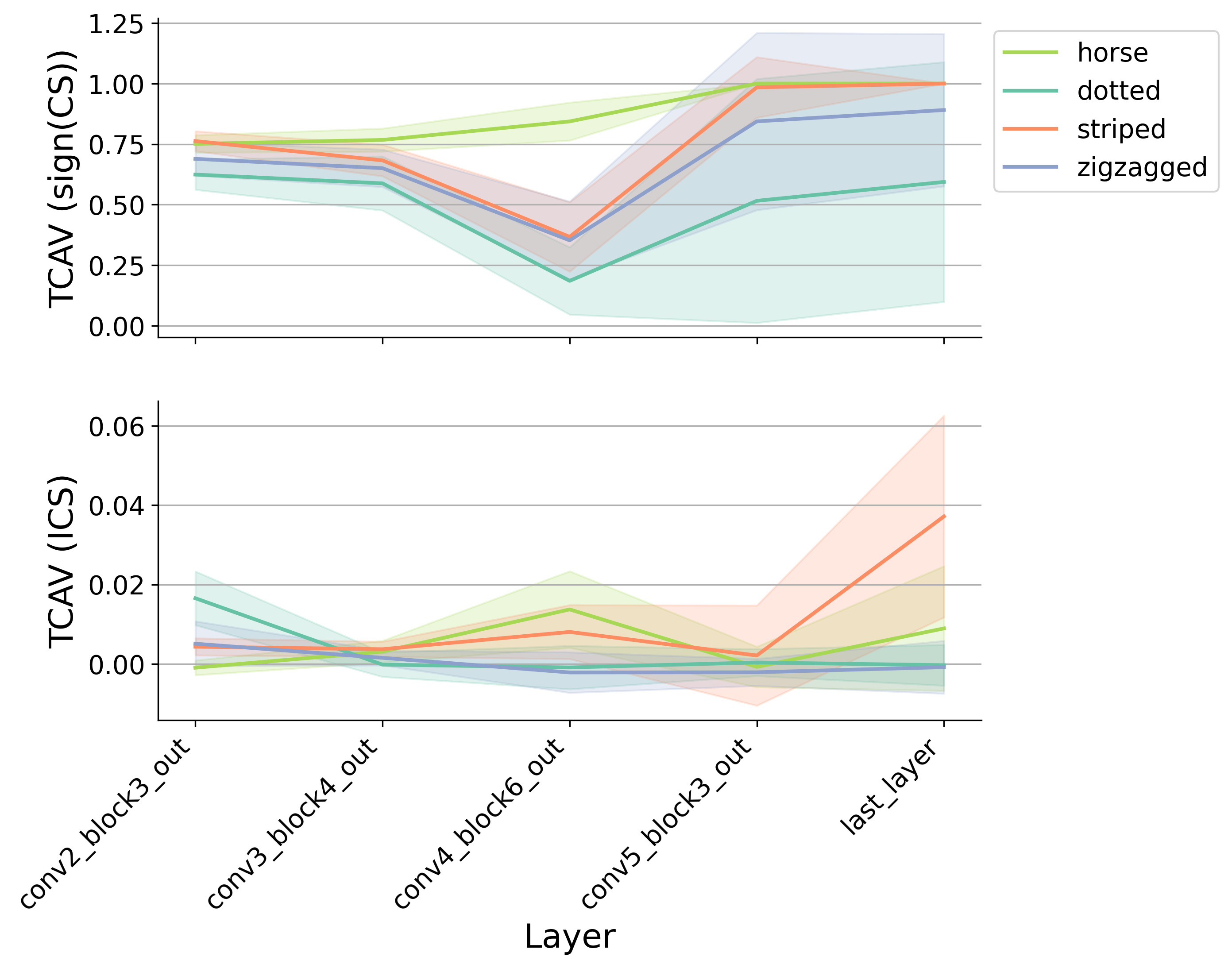}
    \caption{ResNet50.}
    \label{fig:global_zebra_resnet}
\end{figure}

\begin{figure}[!ht]
    \centering
    \includegraphics[width=0.7\columnwidth]{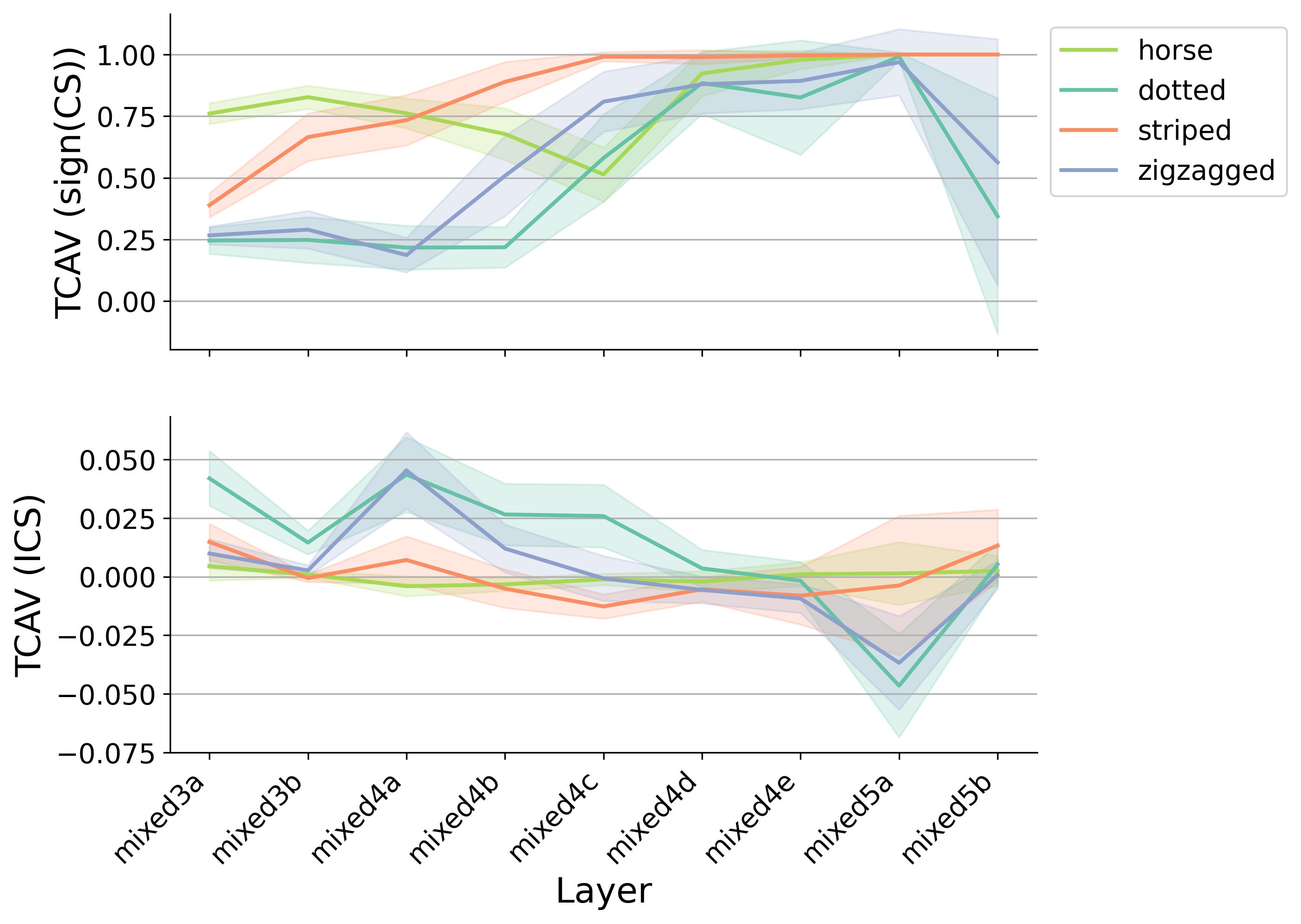}
    \caption{Inception.}
    \label{fig:global_zebra_inception}
\end{figure}

We perform the global analysis for zebras on different models and layers, for the white image baseline. Our results are consistent across both versions of TCAV: the ``zebras'' concept has highest scores (not displayed due to scale), followed by ``stripes'' and then ``horse''. Different architectures (EfficientNet, MobileNet, ResNet, Inception) display different distributions of the results across their layers (Figures \ref{fig:global_zebra_efficientnet}, \ref{fig:global_zebra_mobilenet}, \ref{fig:global_zebra_resnet} and \ref{fig:global_zebra_inception}). It however seems that the deepest layers are the most suited for TCAV analyses across architectures. We also observe that the standard deviation of $\mathrm{TCAV}^{sign(\mathrm{CS})}$ can be large for irrelevant concepts like dotted and zigzagged.

\section{Comparison of computational efficiency}
While ICS seems to provide more reliable concept attribution scores, it is more computationally intensive, since it requires the evaluation of several gradients to estimate the integral. Modern hardware and software capabilities make this constraint not so restrictive, since the gradients can be computed in parallel. Computing ICS score for a single image with a given deep learning model only requires to be able to run a forward pass with batch size 100 (100 values in the sum that approximates the integral). It should work on most PC (e.g. any CPU and 8 - 16Go RAM). However, for high resolution inputs containing many features and large models (e.g. ResNet50 and Inception-v1 have hundreds-of-thousands dimensional activation spaces), all the interpolated samples may not fit in memory.

Efficient quadratures may help mitigate this problem. For example, Bayesian \cite{Briol2015} and Gaussian \cite{Ubaru2017} quadratures may provide up to exponential convergence rates when the integrand is smooth enough. This means that far fewer samples (we used 50) are needed to achieve the same level of estimation error.

We also note that some specific use cases can lead to closed form solutions (e.g. Sec~\ref{app:closed_form}), which are inexpensive to compute.

\section{User study}
\label{app:user_study}

\subsection{Participant statistics}
\label{app:user_study_participants}
Participants provided consent to participate in the study and were informed that they could retract at all times without penalty. Among the 110 ML practitioners and researchers who indicated an interest in the study, 78 completed the study (attrition rate: $29\%$) and were thanked for their time.

ML experience varied extensively within each subgroup, with researchers or practitioners having only followed ML classes (minimum ML experience [in years]: group 1: 0, group 2: 1, group 3: 0), to senior ML researchers/practitioners (maximum ML experience [in years]: group 1: 15, group 2: 25, group 3: 20). Based on paired t-tests, there were no differences between participants in terms of ML experience (main text) or deep learning experience (in years, group 1: $3.3 \pm 1.9$, group 2: $4.1 \pm 2.0$, group 3: $3.6 \pm 2.4$).

During sign up, participants reported their expertise with interpretability methods as `Explai - what?' (given a score of 0), `Somewhat - I have read about explainability' (score of 1), `Yes  - I have implemented some methods' (score of 2) or `Very - I have researched this field and implemented methods' (score of 3). According to this scoring system, there are no significant differences between participants in terms of model interpretability expertise (group 1: $1.8 \pm 0.9$, group 2: $1.5 \pm 1.0$, group 3: $1.1 \pm 0.5$).

Finally, participants were asked whether machine learning was `The topic of research' or `Applied'. 30 out of 78 participants were researchers (group 1: 11, group 2: 12, group 3: 7).

\subsection{Questionnaire and instructions}
\label{app:user_study_design}
Some participants ($\sim 2/3$ in each group) were invited to participate in a 1 hour session where live instructions were provided for 10-15 minutes, participants completed the study without further guidance and the group debriefed the study. Other participants filled in the form on their own time and were sent a recording of the most recent live instructions. In all cases, the form contained all instructions written as preamble sections:

\subsubsection{For all groups}

\noindent \textbf{Task:}
In this experiment, your task is to understand how a deep learning model trained to recognize objects, animals, sports and scenes predicts the class ``zebra''.
 
\begin{figure}[!t]
    \centering
    \includegraphics[width=0.5\textwidth]{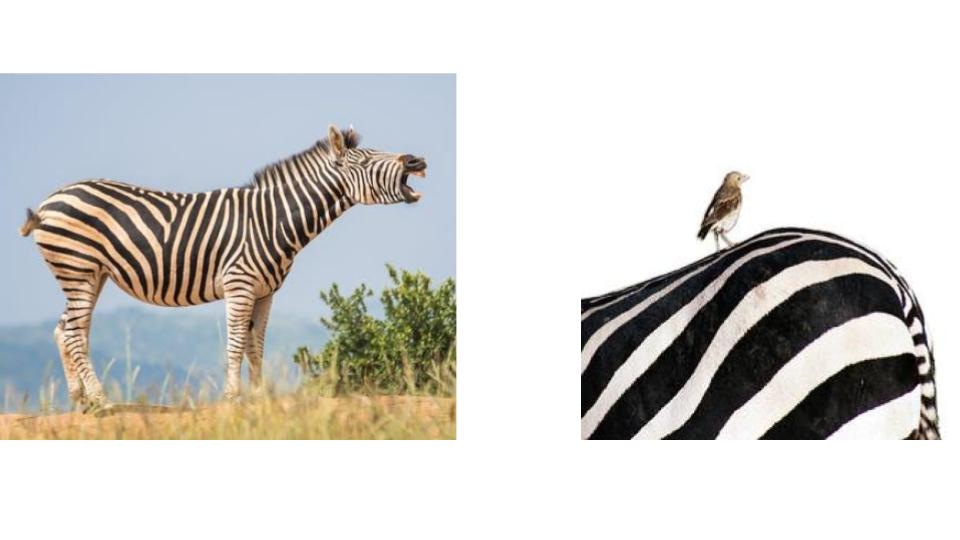}
    \caption{Examples of images you will encounter, some clearly depicting a zebra (left) while others might be more ambiguous (right: is it a bird or a zebra?)}
    \label{fig:user_study_labels}
\end{figure}

\begin{figure}[!t]
    \centering
    \includegraphics[width=0.5\textwidth]{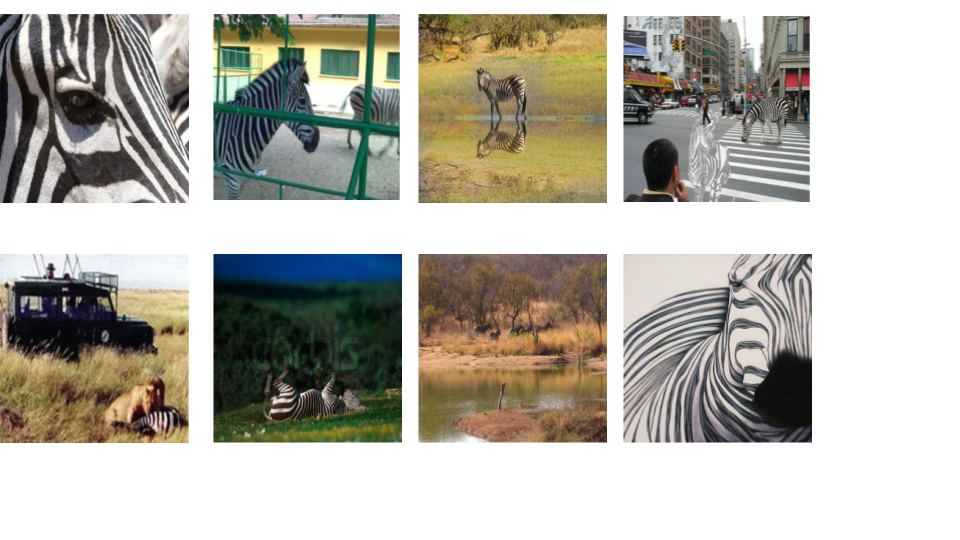}
    \caption{Examples of images labelled as `zebra' in the ImageNet dataset.}
    \label{fig:user_study_training}
\end{figure}

You will be presented with images. For each image, you will be asked 2 questions:

\begin{enumerate}
    \item \textbf{Would you label this image as `zebra'?}  For the image presented, you will first be asked to provide a label. This label should be your top-1 ``ground truth'', i.e. if you could give only one label to the image, would this be ``zebra''? For some images, the answer will be straightforward, while for others it might feel like guessing. This is expected (see Figure \ref{fig:user_study_labels}). 
    \item \textbf{Do you expect the model to predict `zebra' for this image?} By referring to the model's performance and the examples of training images provided below (see Figure \ref{fig:user_study_training}), you then need to guess whether the model identified a zebra in this picture or not as its main class (i.e. top-1 prediction). For instance, you might think that an image that clearly depicts a zebra in the savannah will be predicted as ``zebra''. However,  you might believe that an image of zebras seen from very far away or with multiple species might not be classified as ``zebra''.
\end{enumerate}

\noindent \textbf{Model details}
The model is an Inception-v3 (details here: \url{https://cloud.google.com/tpu/docs/inception-v3-advanced}) model trained on ImageNet. The model predicts 1000 classes for ImageNet, including animals, sports, objects, scene types (although not natural scenes), \dots A full list of the classes can be found here: 
\url{https://storage.googleapis.com/download.tensorflow.org/data/ImageNetLabels.txt}. The overall model performance is $77.9\%$ (top - 1 accuracy).

We focus on one class in this task: zebra

The model is trained with images labelled as ``zebra'' that are quite diverse: zebras are seen from close up, from far away, with other species, in their natural surroundings (mostly) or in human scenes. A few examples are presented below (see Figure \ref{fig:user_study_training}).

On a subset of images labelled as ``zebra'' from ImageNet, the model correctly predicts zebra as its top-1 class for $92\%$ of images.

\subsubsection{Group 1: no explanations}

Participants then viewed three duples of zebra image (from ImageNet as these are labelled) and model prediction, including two positives and a false negative. They were then presented with the following text before starting the study:

``You have been assigned to group 1. This means that you will be presented with the images only. No model explainability will be presented. In this case, we are trying to understand whether humans can predict which images would be easy for the model to correctly identify as a zebra, compared to more difficult. Ready to start?''

\begin{figure}
\centering
\includegraphics[width=0.15\textwidth]{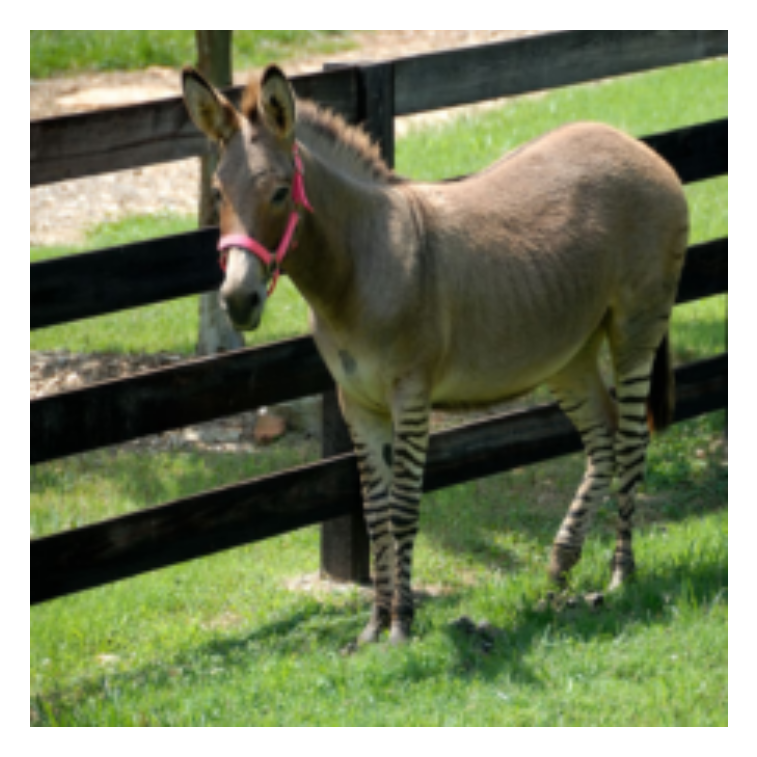}
\includegraphics[width=0.15\textwidth]{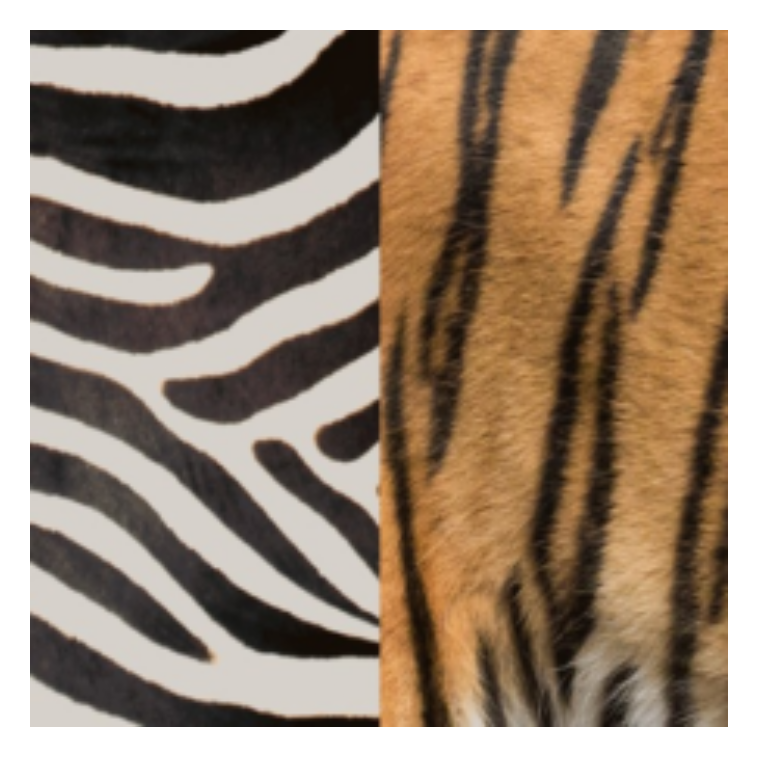}
\includegraphics[width=0.15\textwidth]{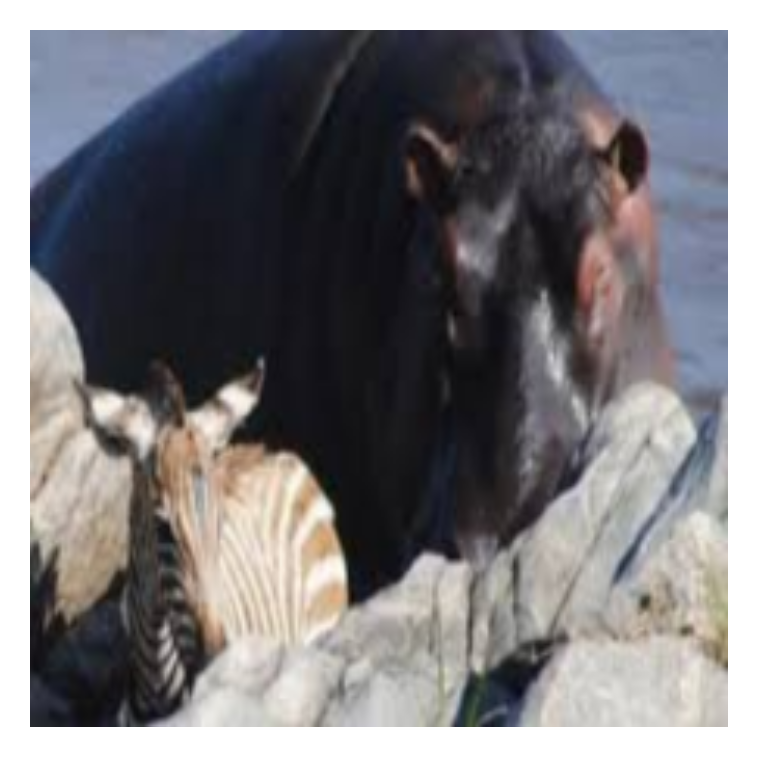}
\includegraphics[width=0.15\textwidth]{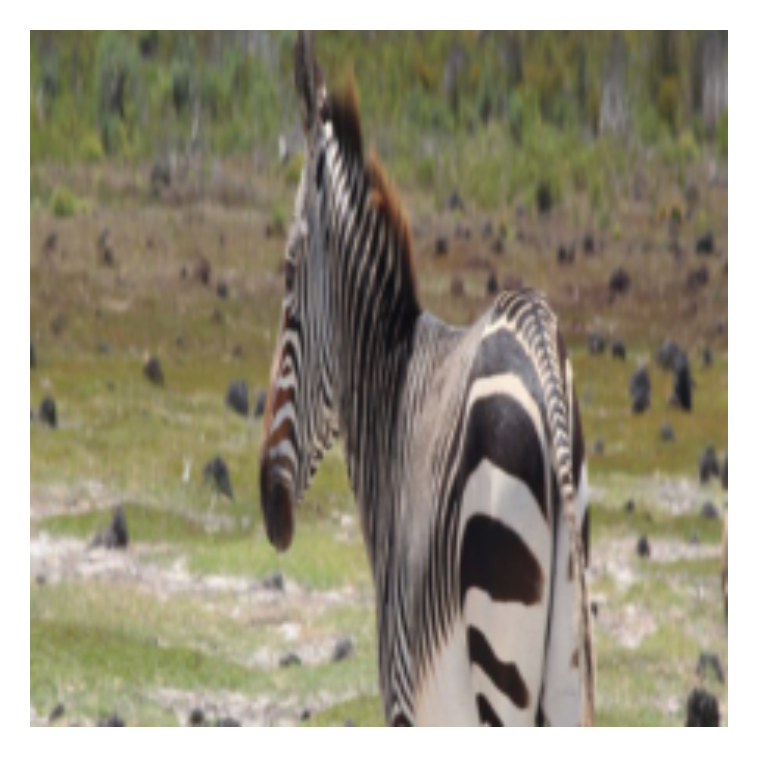}
\includegraphics[width=0.15\textwidth]{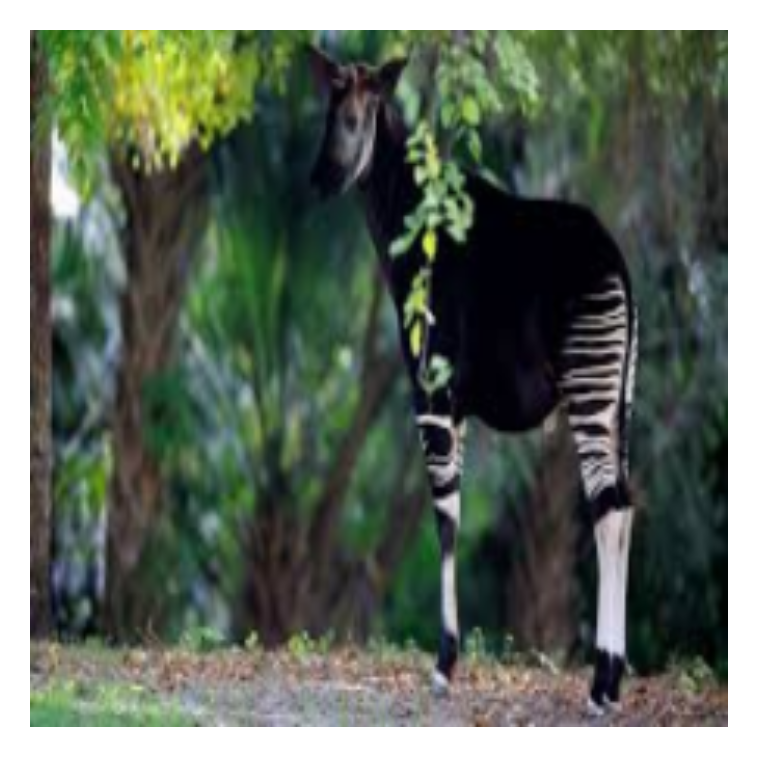} \\
\includegraphics[width=0.15\textwidth]{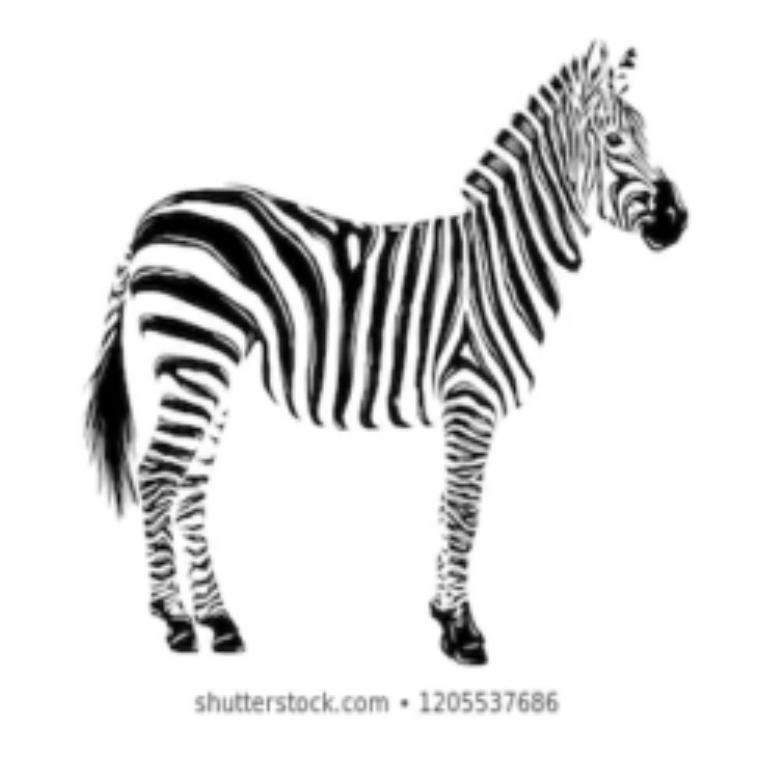}
\includegraphics[width=0.15\textwidth]{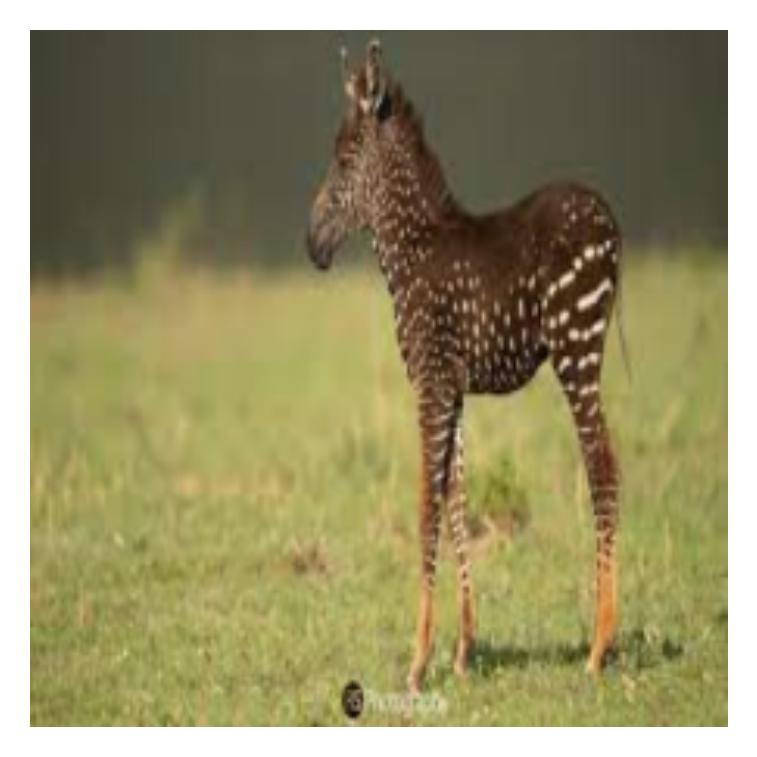}
\includegraphics[width=0.15\textwidth]{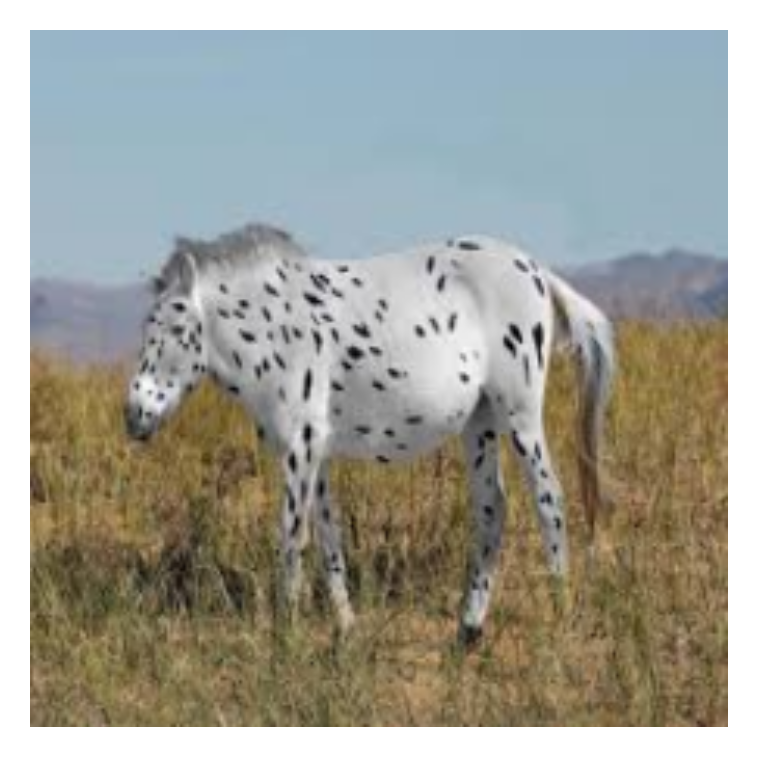}
\includegraphics[width=0.15\textwidth]{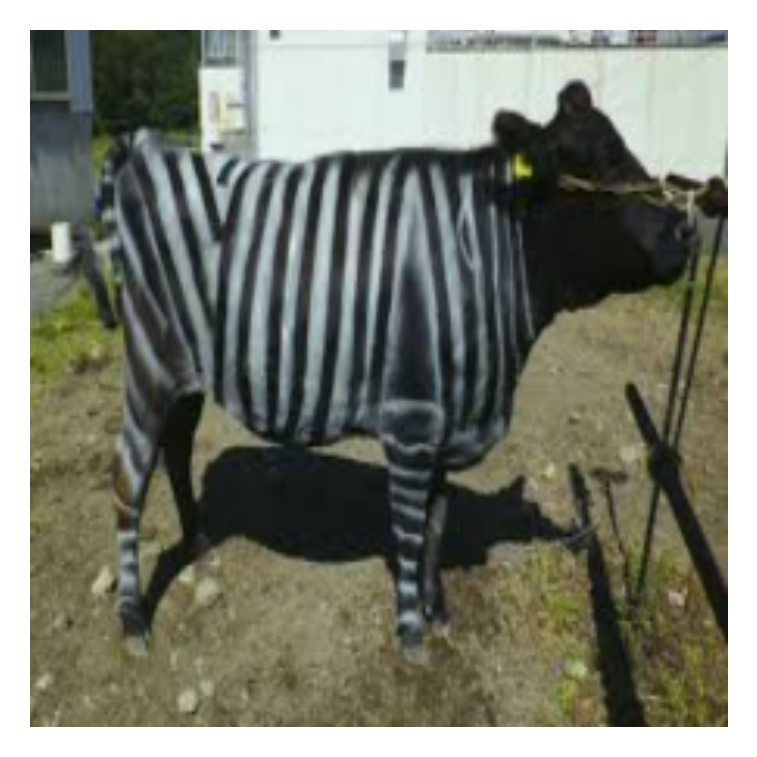}
\includegraphics[width=0.15\textwidth]{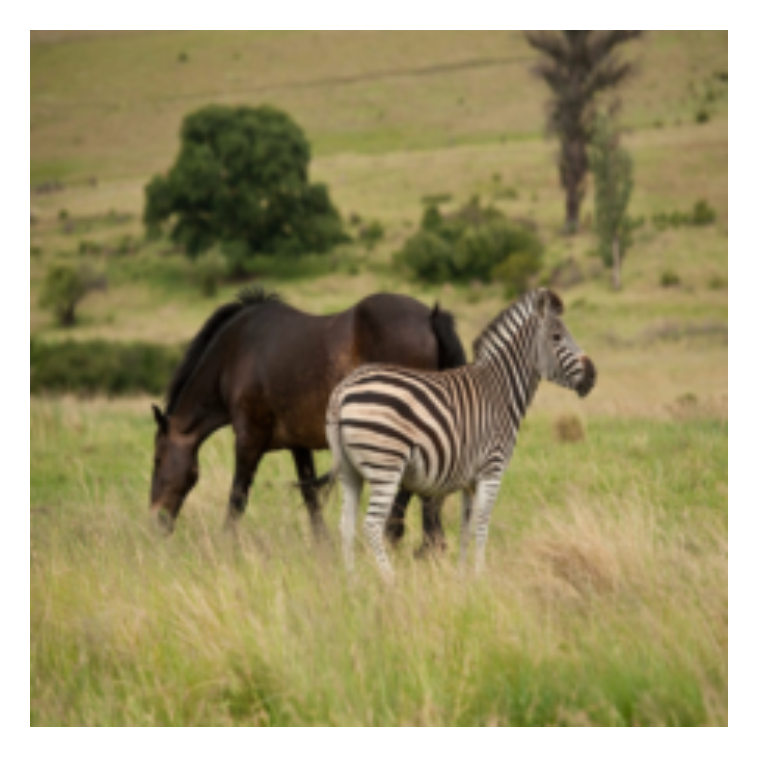} \\
\includegraphics[width=0.15\textwidth]{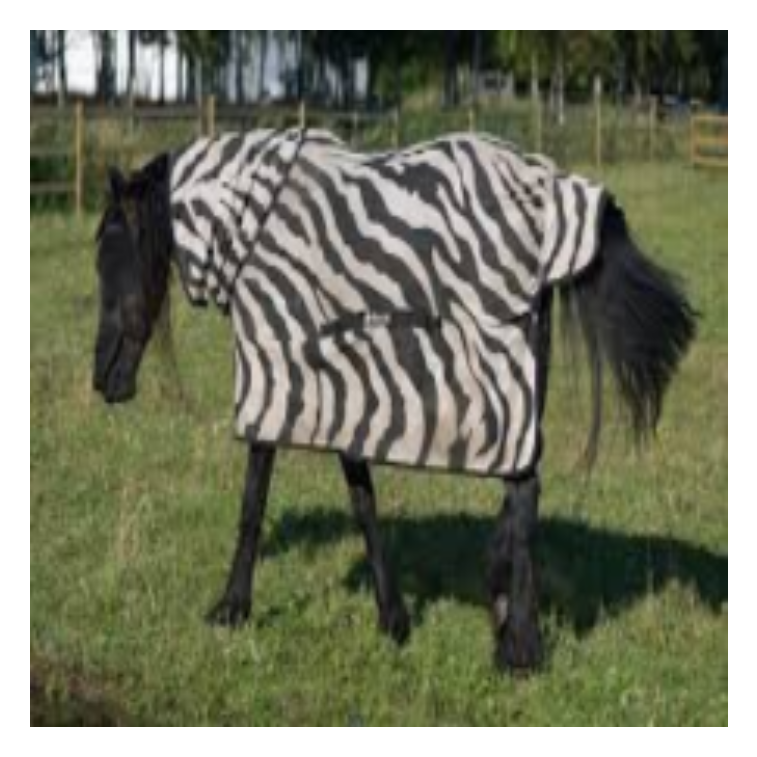}
\includegraphics[width=0.15\textwidth]{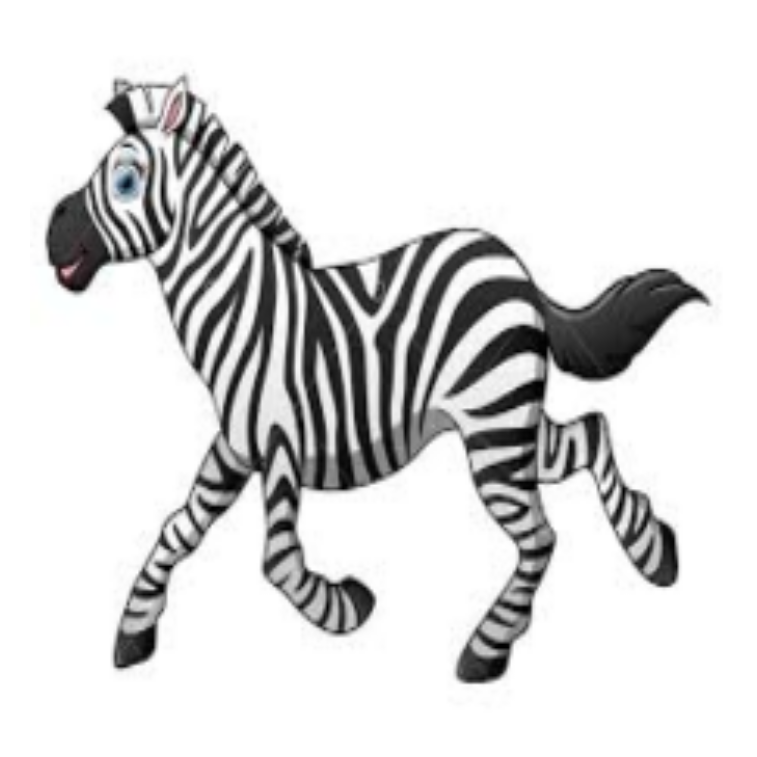}
\includegraphics[width=0.15\textwidth]{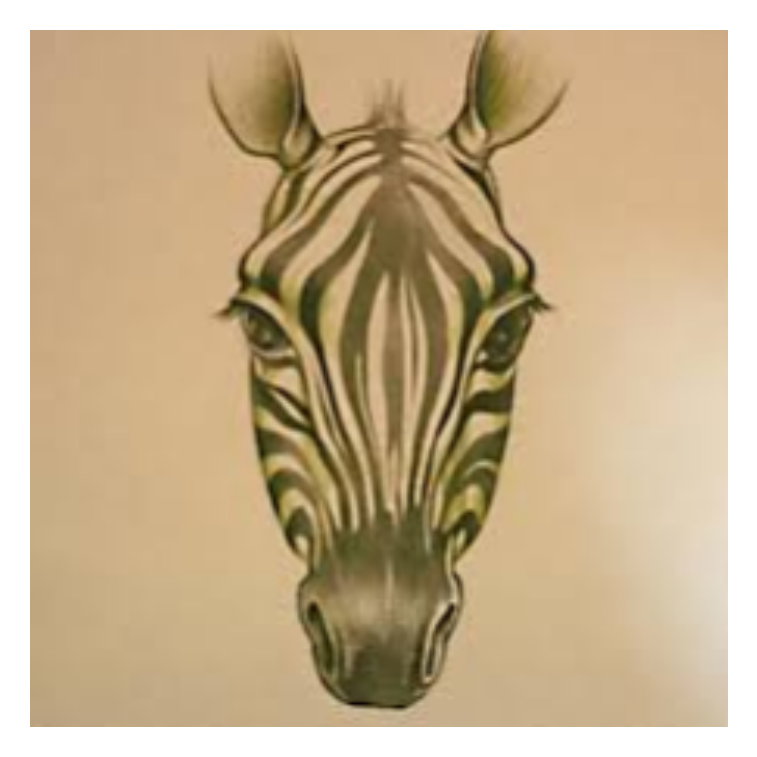}
\includegraphics[width=0.15\textwidth]{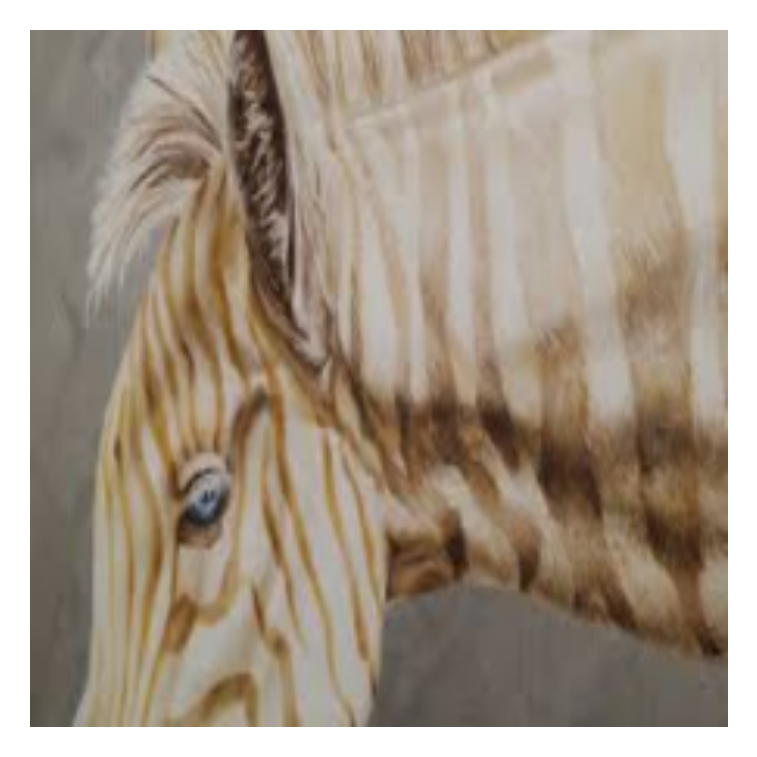}
\includegraphics[width=0.15\textwidth]{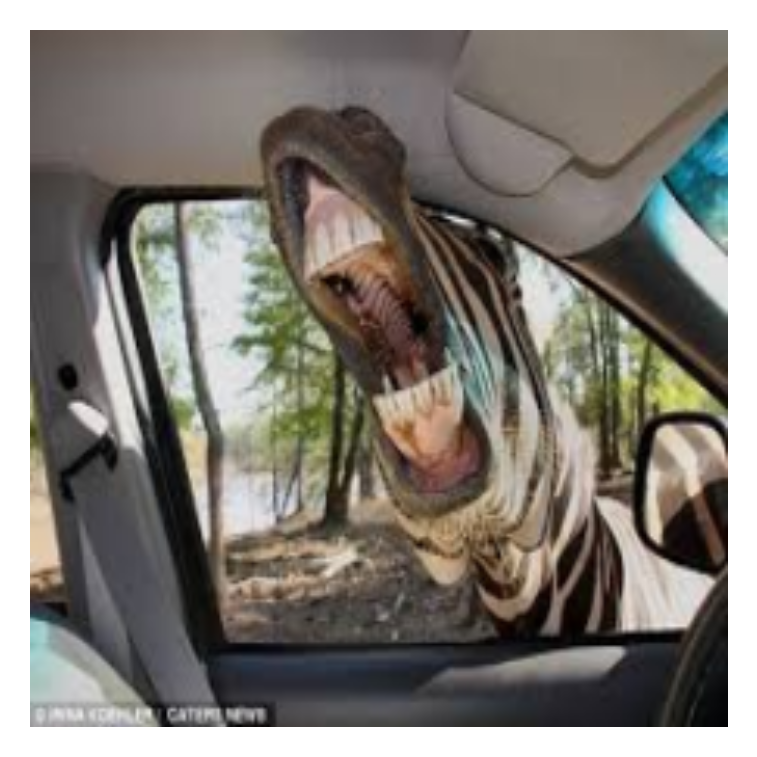} \\
\includegraphics[width=0.15\textwidth]{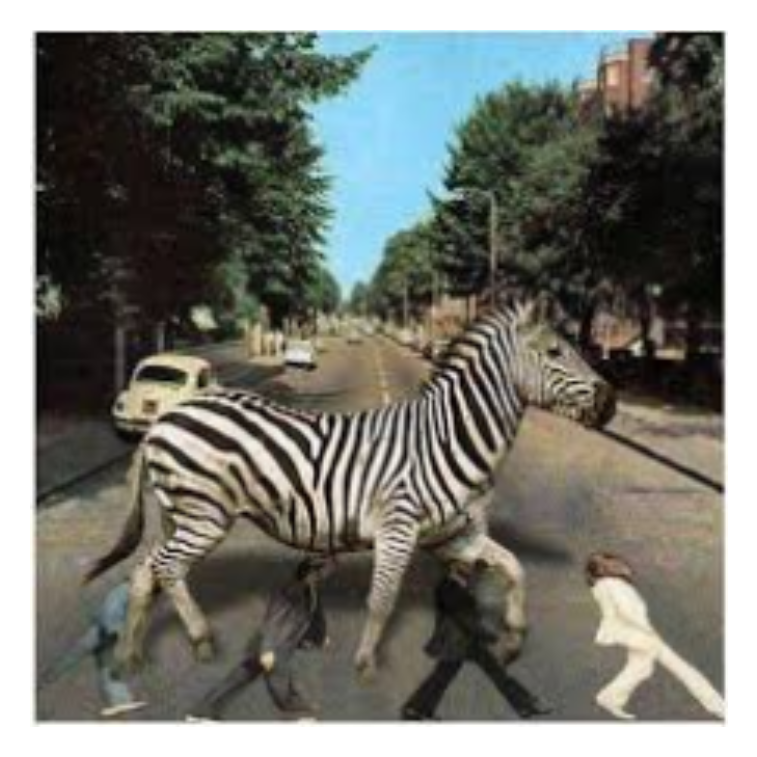}
\includegraphics[width=0.15\textwidth]{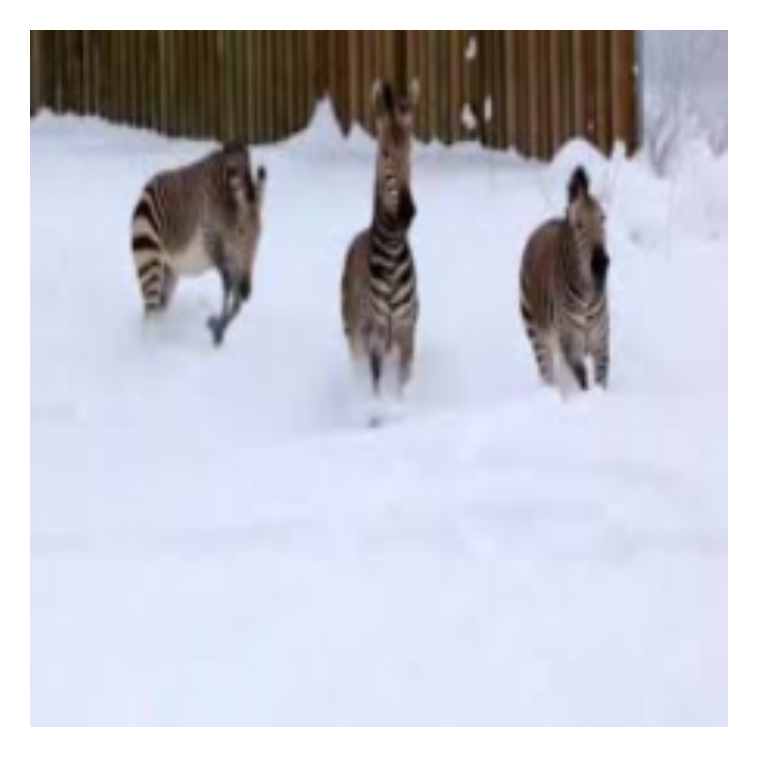}
\includegraphics[width=0.15\textwidth]{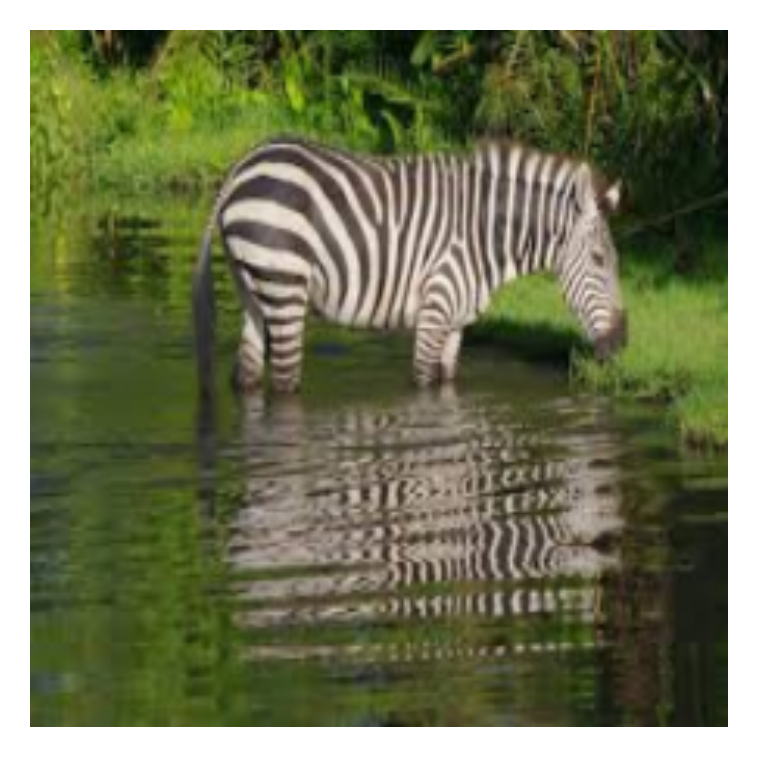}
\includegraphics[width=0.15\textwidth]{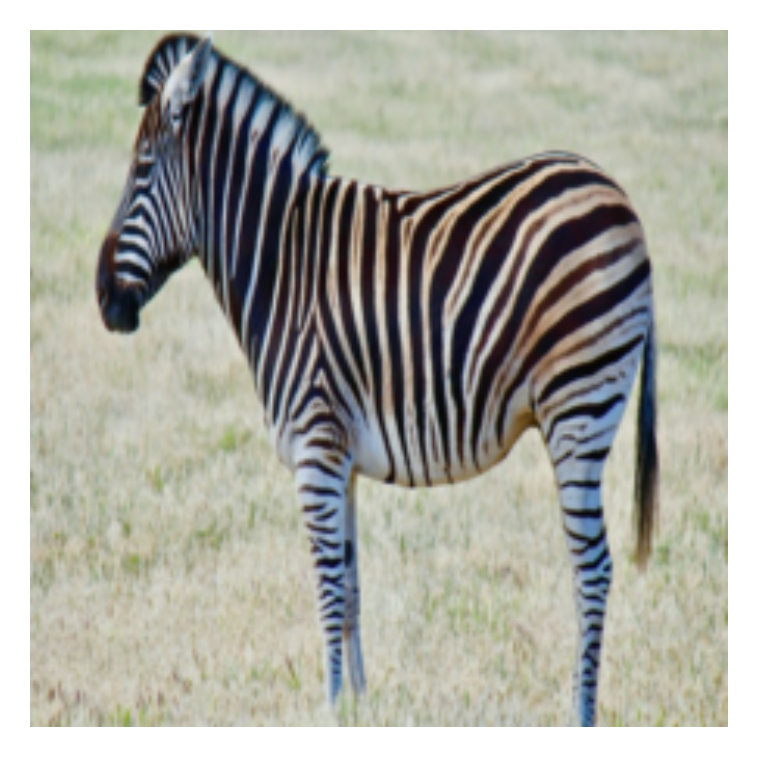}
\includegraphics[width=0.15\textwidth]{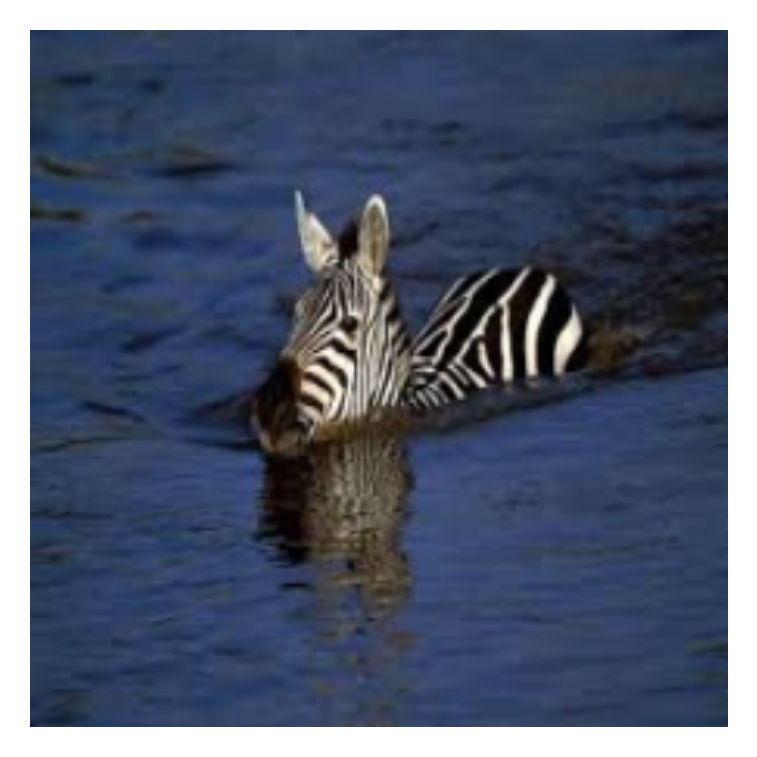} \\
\includegraphics[width=0.15\textwidth]{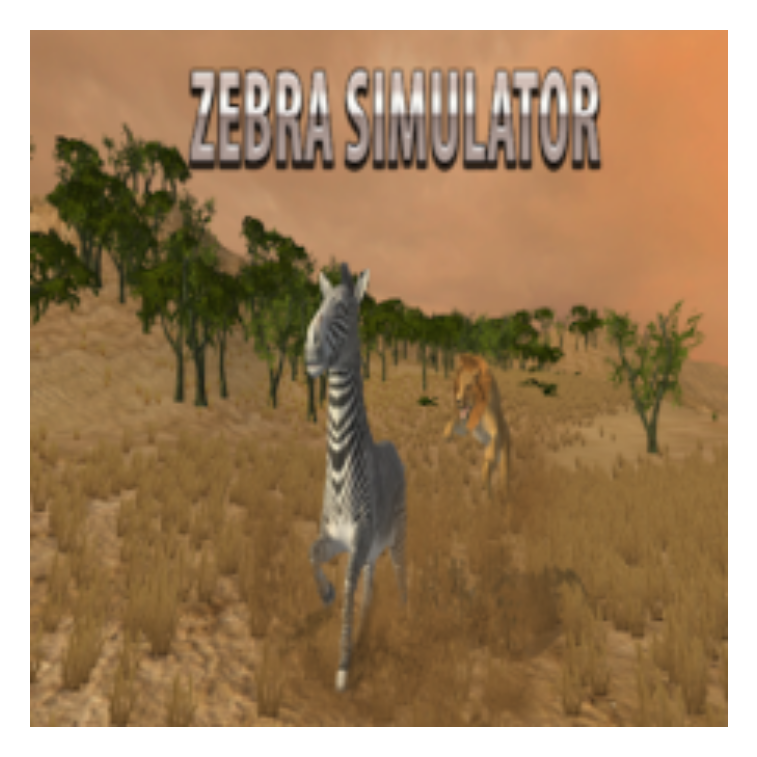}
\includegraphics[width=0.15\textwidth]{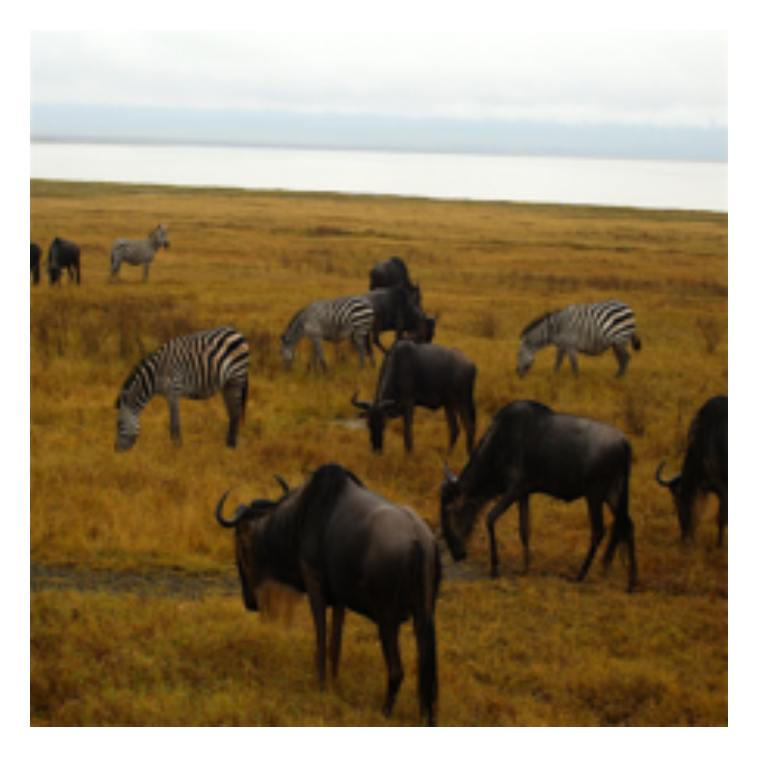}
\includegraphics[width=0.15\textwidth]{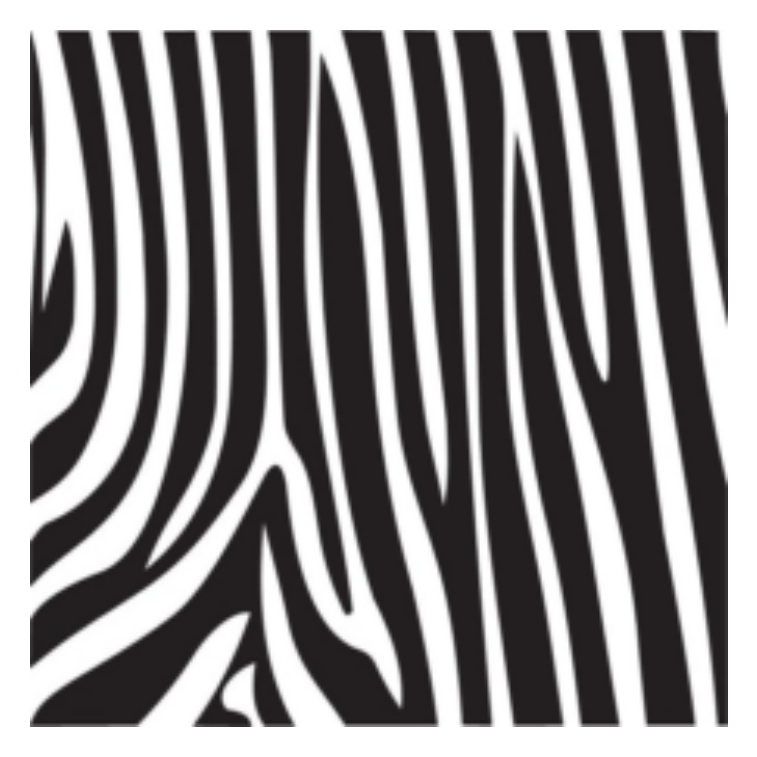}
\includegraphics[width=0.15\textwidth]{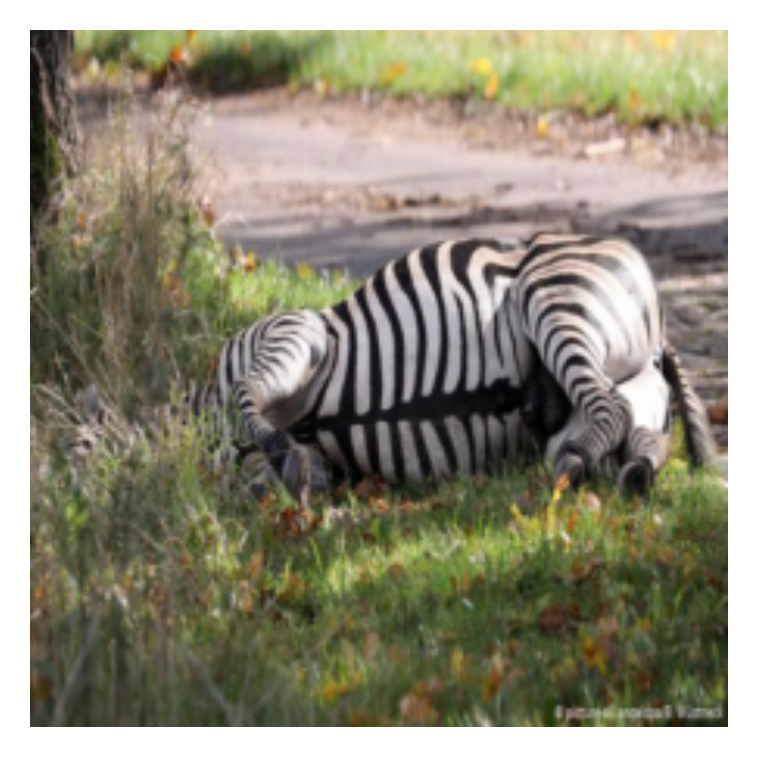}
\includegraphics[width=0.15\textwidth]{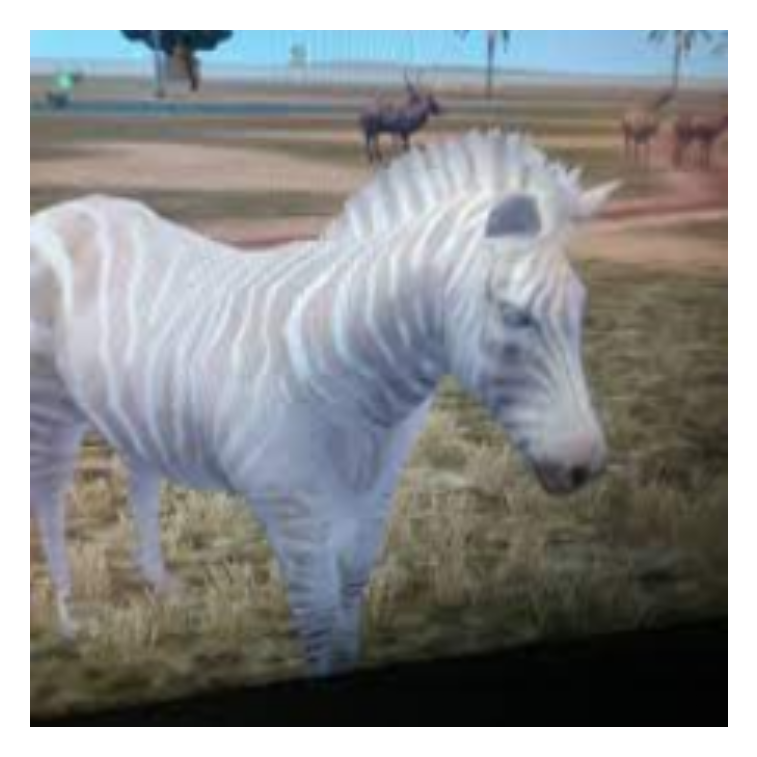} \\
\includegraphics[width=0.15\textwidth]{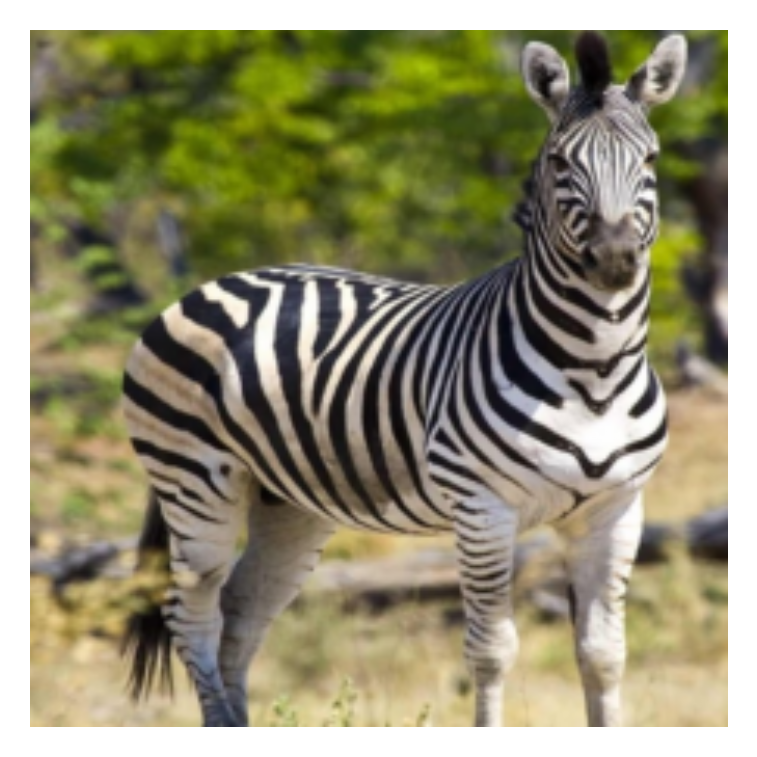}
\includegraphics[width=0.15\textwidth]{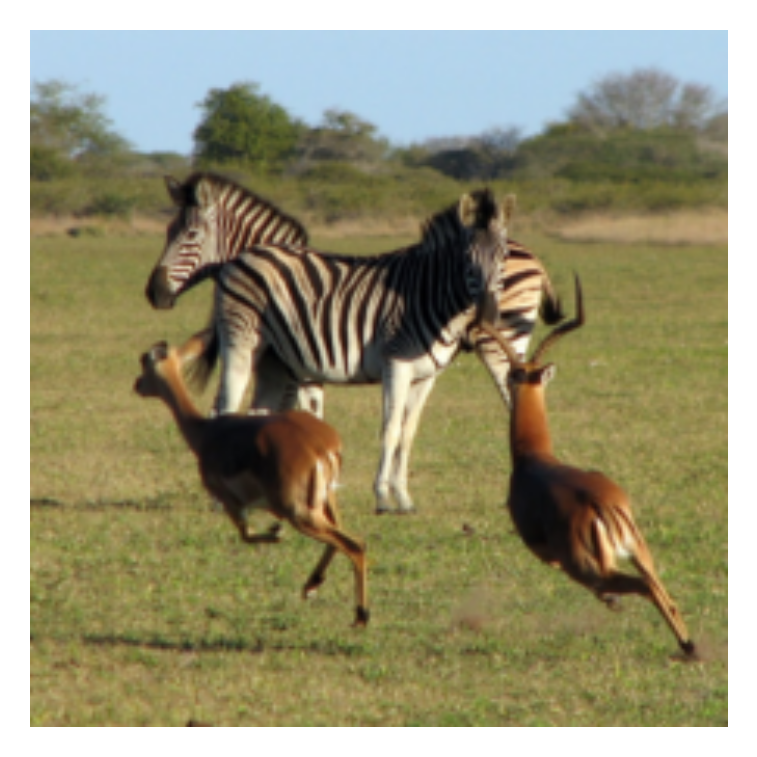}
\includegraphics[width=0.15\textwidth]{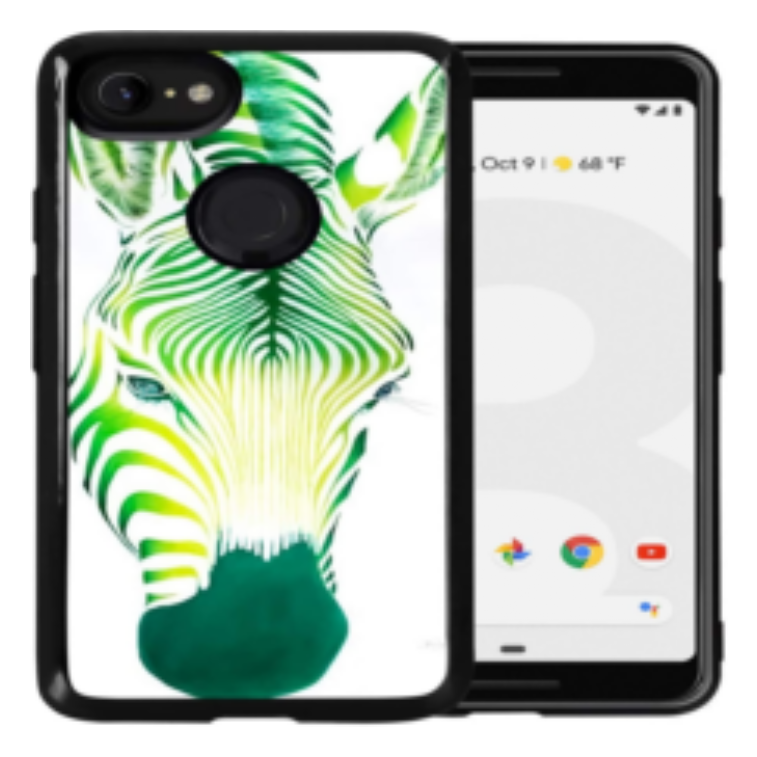}
\includegraphics[width=0.15\textwidth]{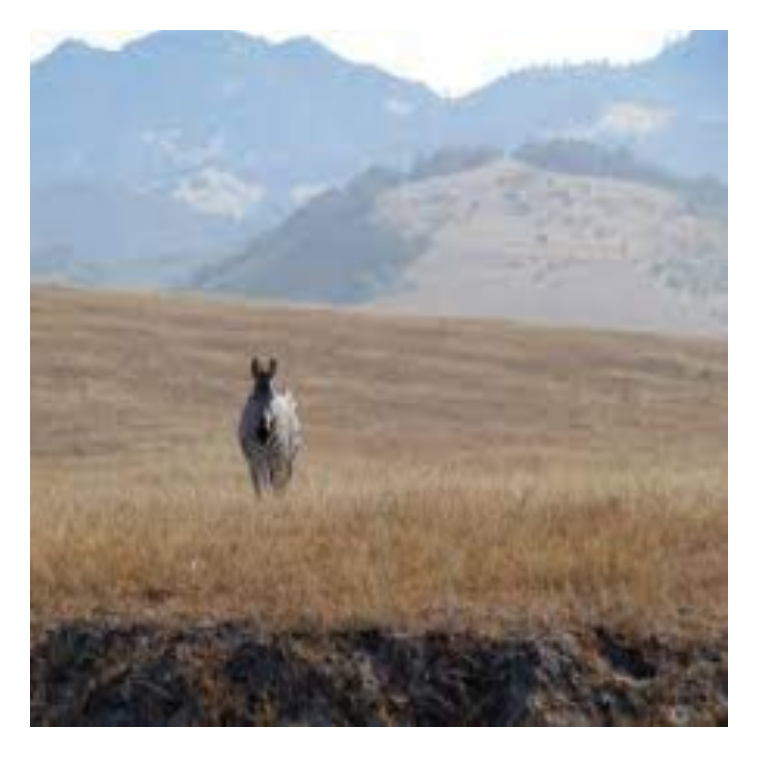}
\includegraphics[width=0.15\textwidth]{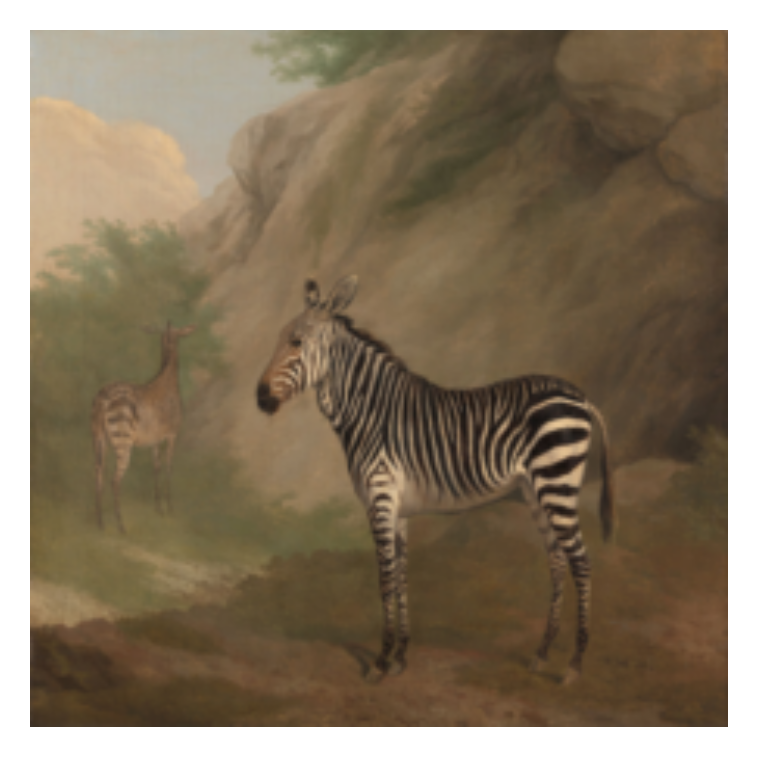} \\
\includegraphics[width=0.15\textwidth]{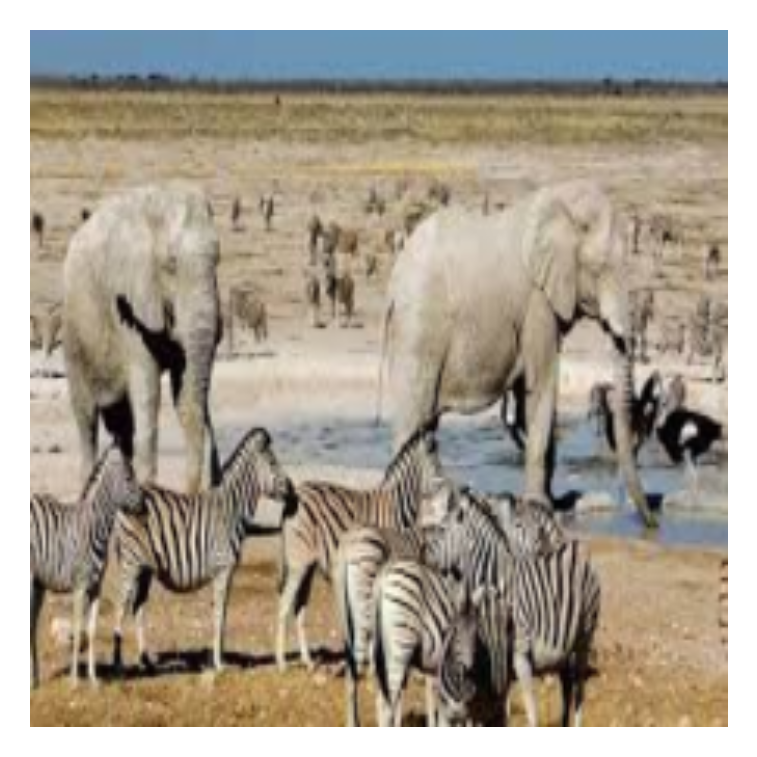}
\includegraphics[width=0.15\textwidth]{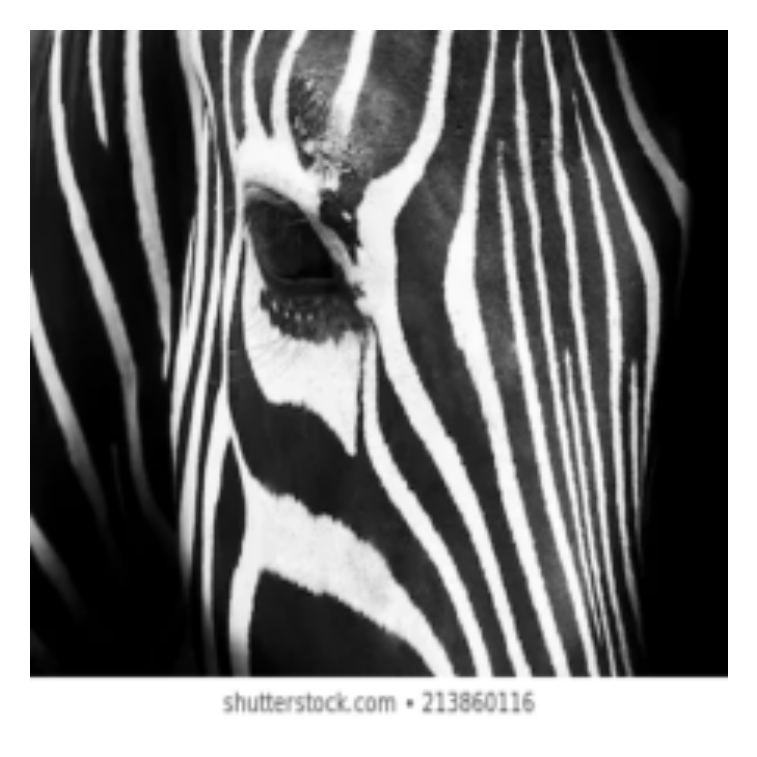}
\includegraphics[width=0.15\textwidth]{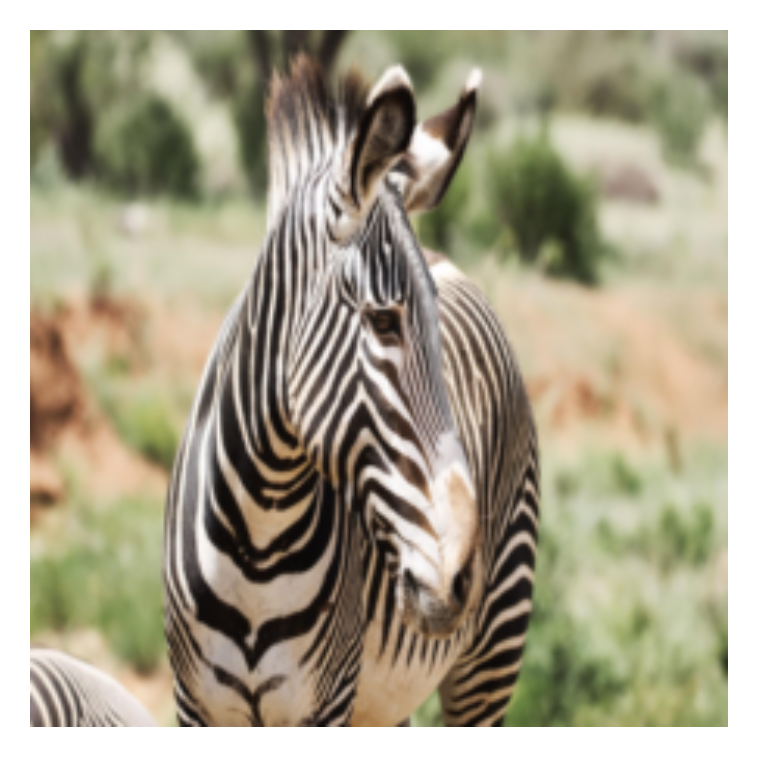}
\includegraphics[width=0.15\textwidth]{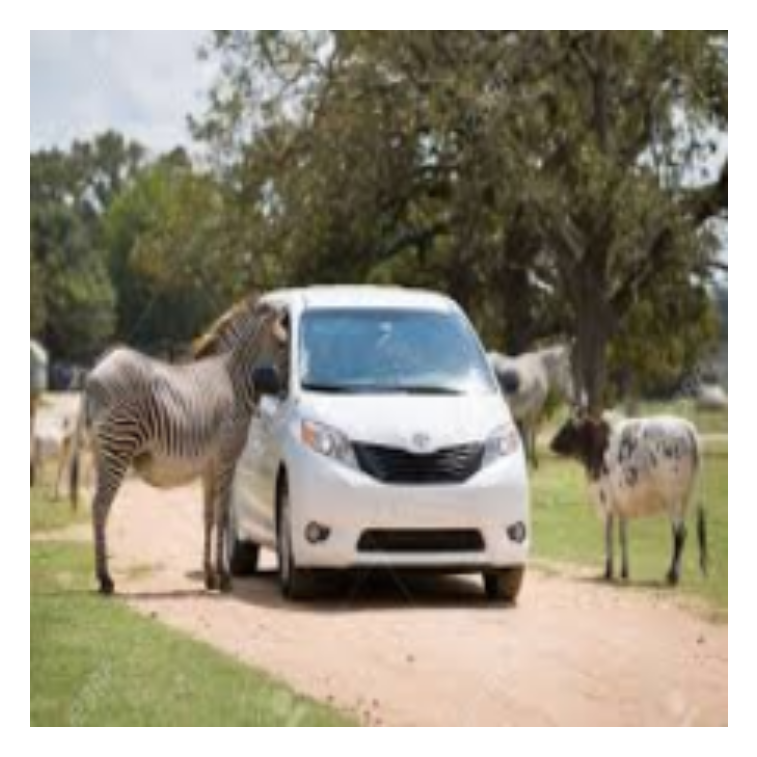}
\includegraphics[width=0.15\textwidth]{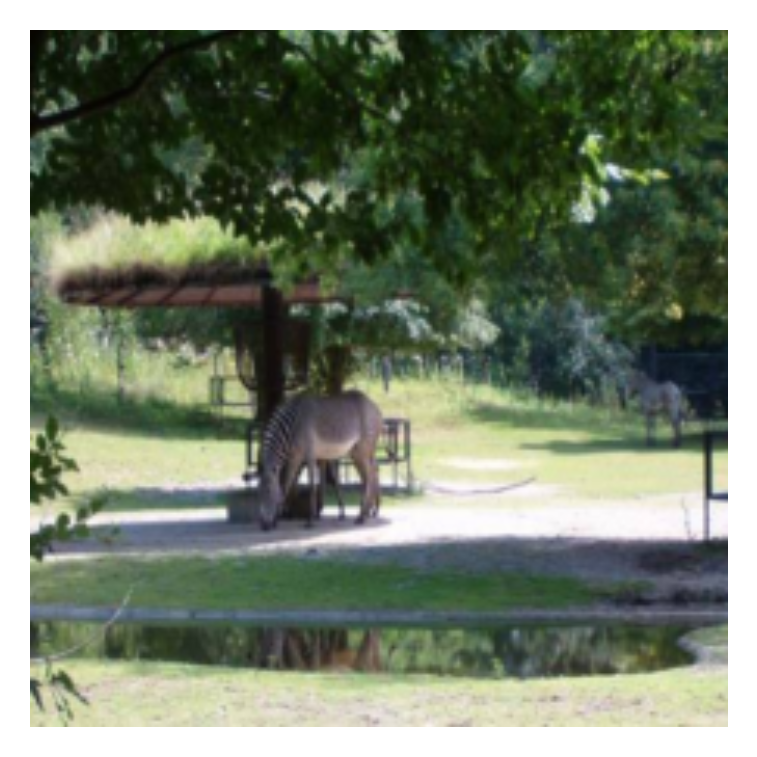} \\
\includegraphics[width=0.15\textwidth]{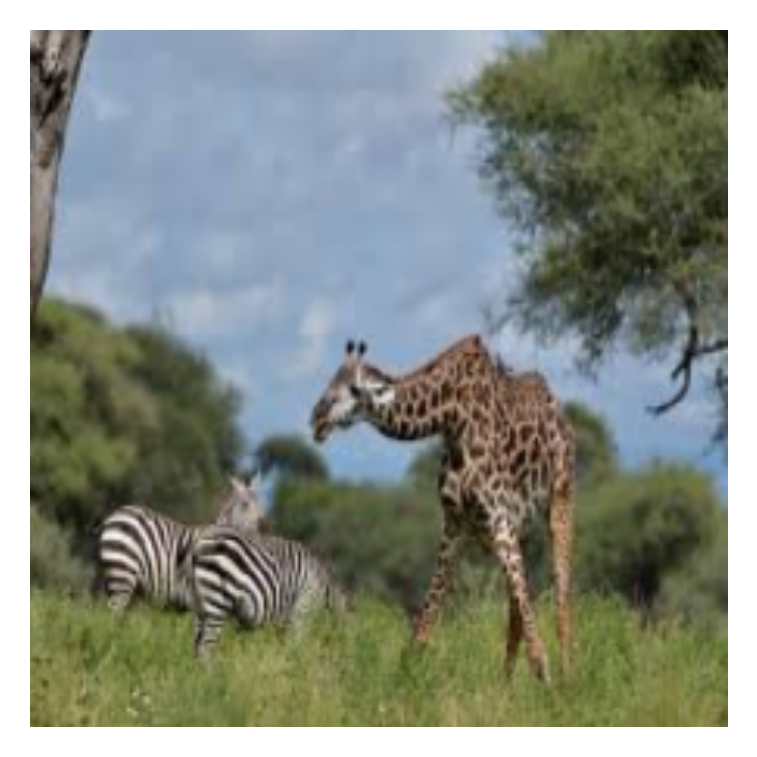}
\includegraphics[width=0.15\textwidth]{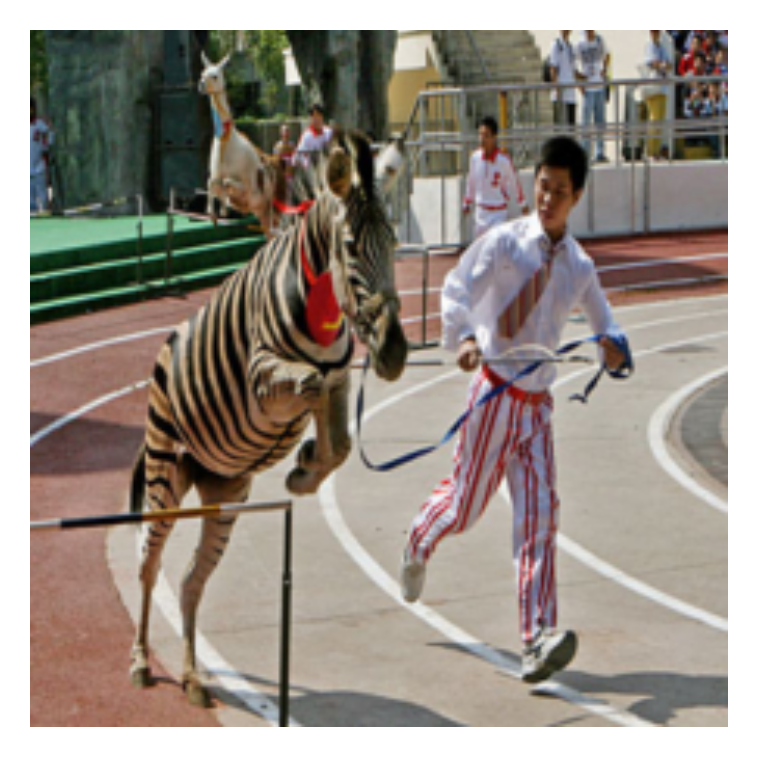}
\includegraphics[width=0.15\textwidth]{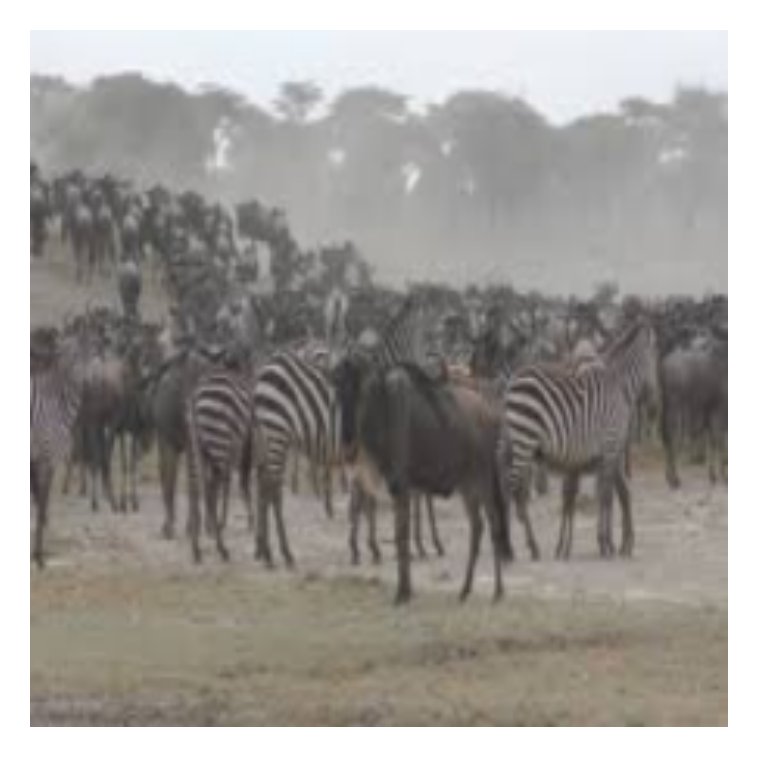}
\includegraphics[width=0.15\textwidth]{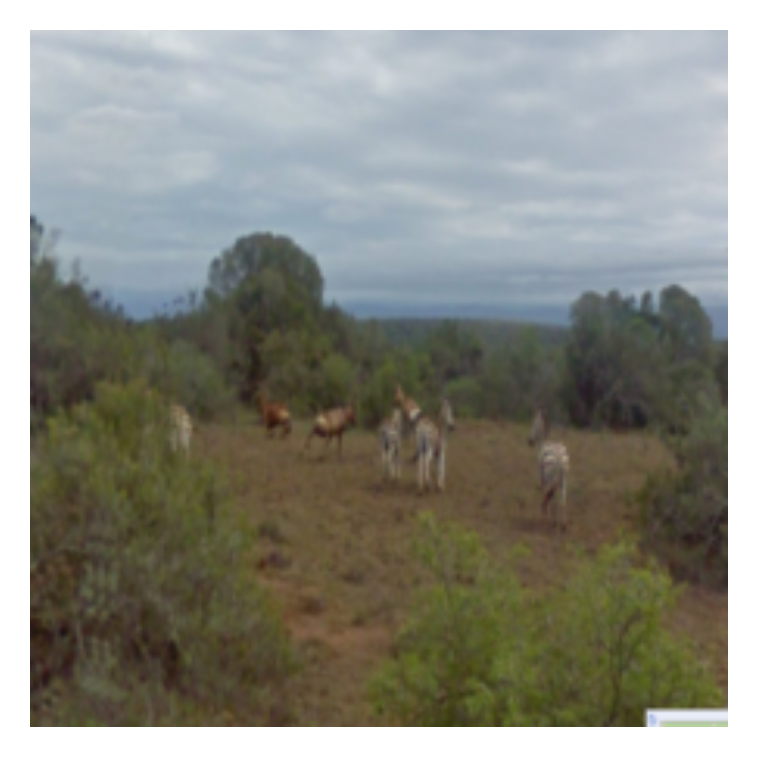}
\includegraphics[width=0.15\textwidth]{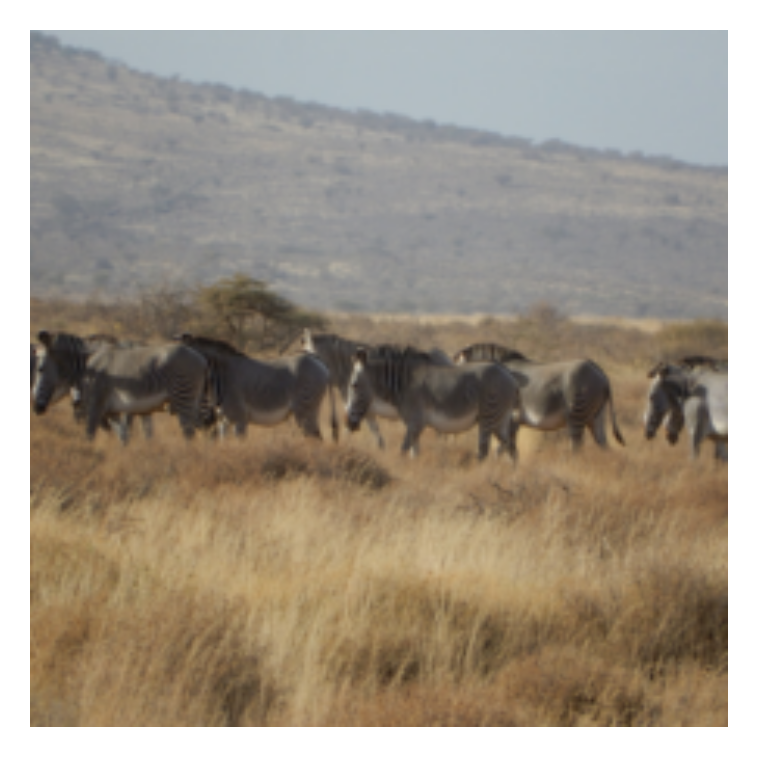} \\
\caption{Images used in the user study. We note that some images were consistently mislabeled by humans (e.g. a `polka dot' zebra not identified as a zebra).}
\label{app:fig_user_study_images}
\end{figure}

They were then presented with an image (see Figure~\ref{app:fig_user_study_images} for all images used) and the 2 questions as a section of the form (as illustrated in the main text).

\subsubsection{Group 2: global explanations}
Global explanations were described in terms of the concepts defined and a brief analysis of the results followed (see below). The global explanations were displayed with each question. Detailed instructions:

\begin{figure}[!t]
\centering
\begin{subfigure}[t]{0.42\textwidth}
\textbf{Animal}
\end{subfigure}
\hspace{0.09\textwidth}
\begin{subfigure}[t]{0.42\textwidth}
\textbf{Striped}
\end{subfigure} \\
\begin{subfigure}[t]{0.42\textwidth}
\includegraphics[width=\linewidth,valign=t]{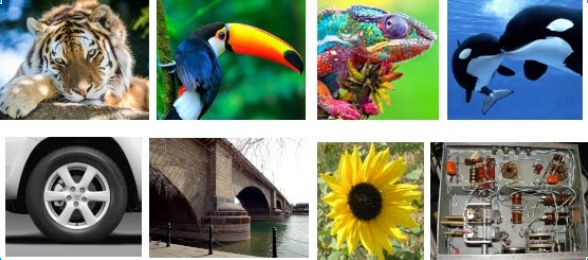} 
\end{subfigure}
\hspace{0.09\textwidth}
\begin{subfigure}[t]{0.42\textwidth}
\includegraphics[width=\linewidth,valign=t]{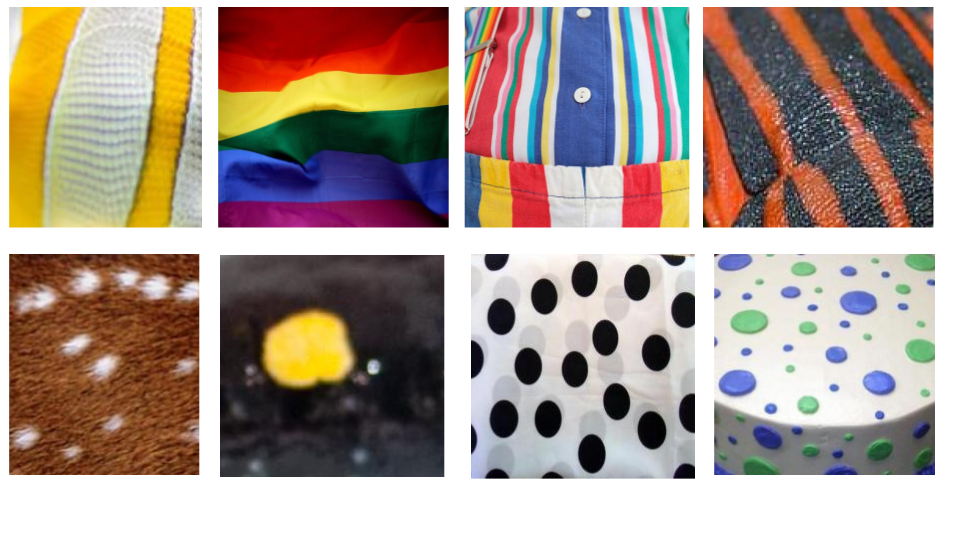} 
\end{subfigure} \\
\begin{subfigure}[t]{0.42\textwidth}
\textbf{Savannah}
\end{subfigure}
\hspace{0.09\textwidth}
\begin{subfigure}[t]{0.42\textwidth}
\textbf{Four-legged}
\end{subfigure} \\
\begin{subfigure}[t]{0.42\textwidth}
\includegraphics[width=\linewidth,valign=t]{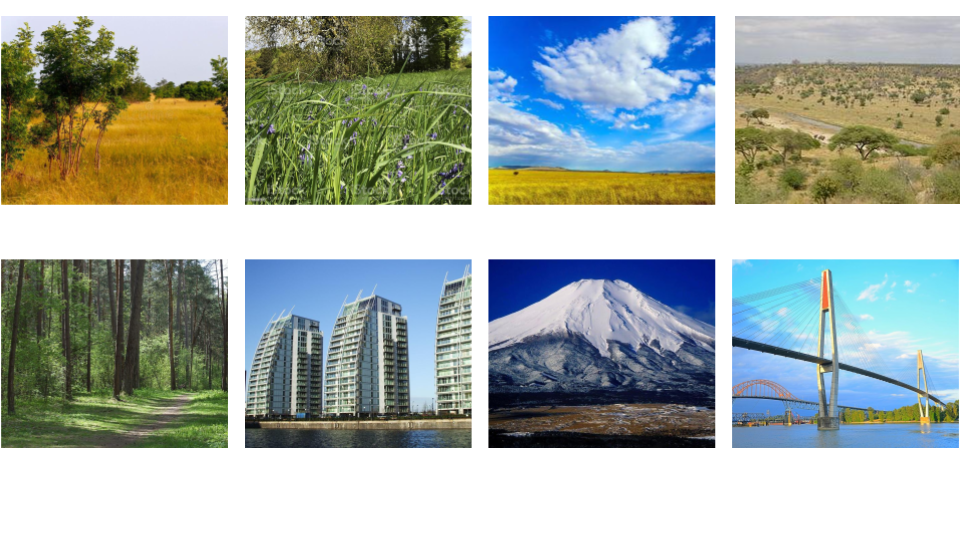} 
\end{subfigure}
\hspace{0.09\textwidth}
\begin{subfigure}[t]{0.42\textwidth}
\includegraphics[width=\linewidth,valign=t]{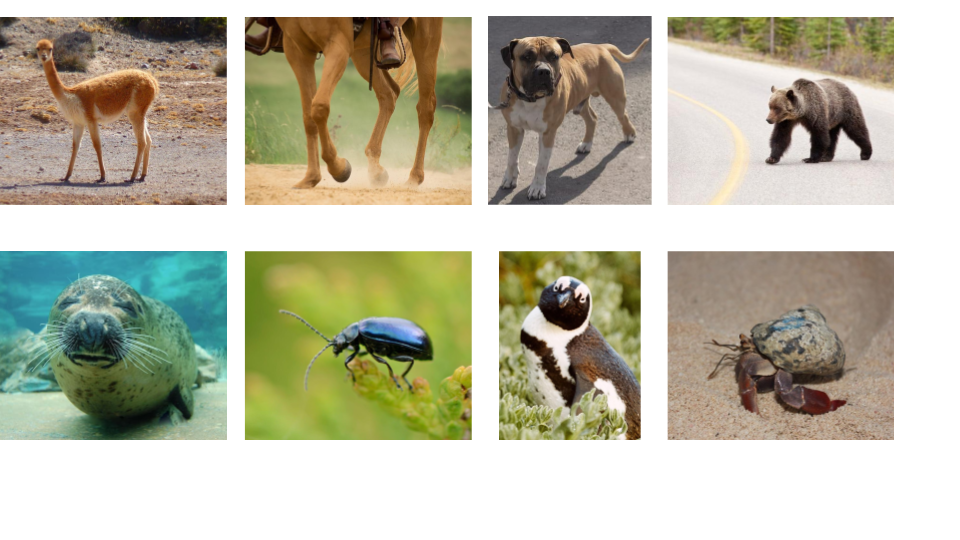} 
\end{subfigure} \\
\begin{subfigure}[t]{0.42\textwidth}
\textbf{Horse shape}
\end{subfigure}
\hspace{0.09\textwidth}
\begin{subfigure}[t]{0.42\textwidth}
\textbf{Color}
\end{subfigure}\\
\begin{subfigure}[t]{0.42\textwidth}
\includegraphics[width=\linewidth,valign=t]{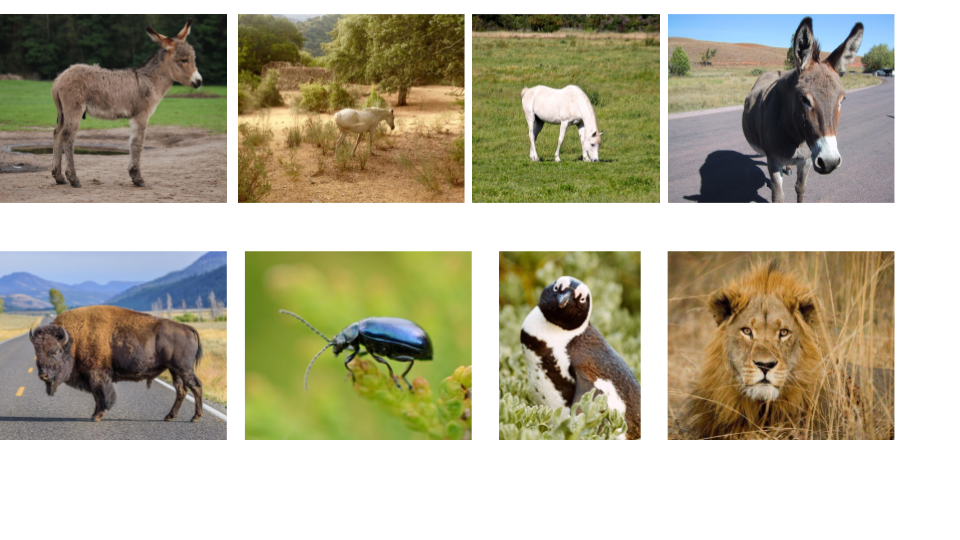} 
\end{subfigure}
\hspace{0.09\textwidth}
\begin{subfigure}[t]{0.42\textwidth}
\includegraphics[width=\linewidth,valign=t]{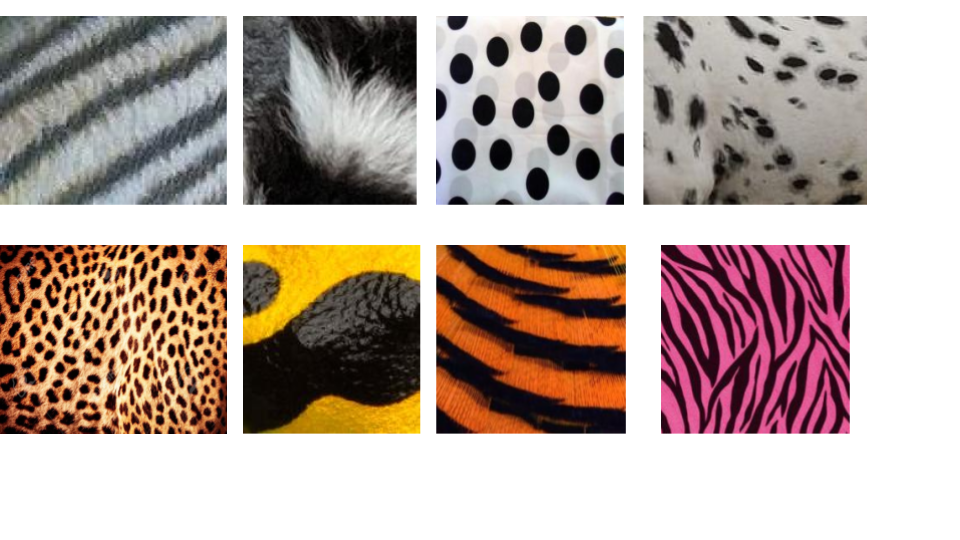} 
\end{subfigure}
\caption{For each concept, 5 images represented the concept (top row) and 5 images represented the `control' (bottom row) when building the relative concept vector between the 2 sets. On average, around 100 images from each set were used to build the CAVs.}
\label{fig:user_study_concepts}
\end{figure}

``You have been assigned to group 2. This means that you will be presented with the images to make your assessment, and that we provide global model explanations here to frame your reasoning. Global explanations highlight, on average, what features the model relies on to predict a certain class.

We built high-level concepts, and estimated how much the model relied on these concepts to provide a ``zebra'' prediction.

We investigated the following concepts (see Figure \ref{fig:user_study_concepts}):
\begin{itemize}
    \item \textbf{Animal}: This concept investigates the importance of animals compared to objects, plants, buildings and bridges. Below we display images that were used as examples of the concept (top row), compared to examples that were not considered as examples of the concept (bottom row).
    \item \textbf{Striped}: This concept investigates the importance of stripes from different textures to dotted patterns from different textures.
    \item \textbf{Savannah}:  This concept investigates the importance of a grassy backgrounds (fields or savannah) compared to other natural and man-made scenes.
    \item \textbf{Four-legged}: This concept investigates the importance of an animal with 4 legs compared to a random set of animals.
    \item \textbf{Horse shape}: This concept investigates the importance of the shape of horses and donkeys compared to a random set of animals.
    \item \textbf{Color}: This concept investigates the importance of the black and white combination compared to more colorful patterns.
\end{itemize}

\noindent\textbf{Results and how to use this information:}

\begin{figure}[!ht]
    \centering
    \includegraphics[width=\textwidth]{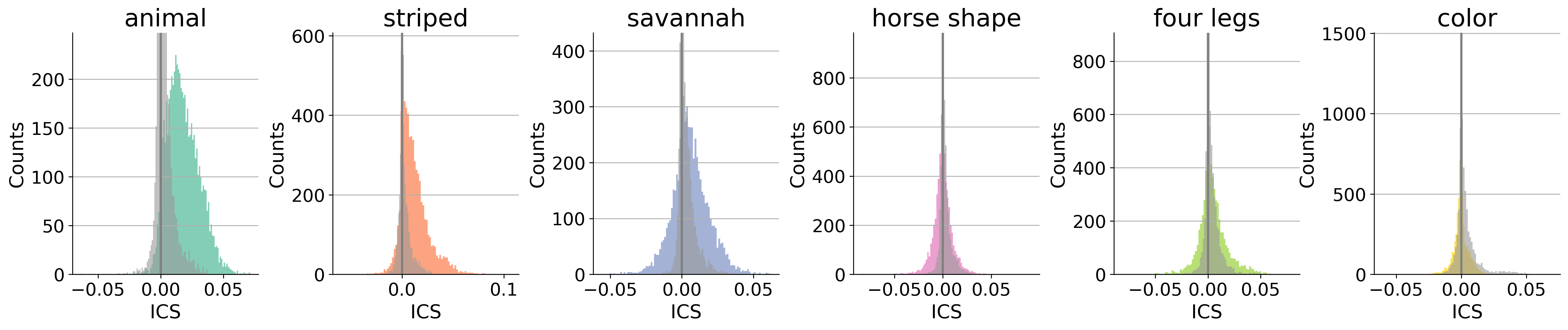}
    \caption{ICS distributions for the zebra class, for each concept considered in the user study. This plot is used as `global explanations.'}
    \label{fig:user_study_global}
\end{figure}

After analysis, we obtain a score called ``ICS'' for each image and each concept (Figure \ref{fig:user_study_global}). Across all zebra images, the plot displays how much the model relies on each concept (colored distribution), compared to a random concept (grey distribution).

We observe that the model seems to rely mostly on the presence of an animal, and on stripes. The background plays a role too, with grassy fields and grassy savannah being indicative of a zebra. The horse shape can be important for some images, but is not overall indicative of a zebra (displaying the diversity of images provided for training). Similarly, the detection of 4 legs points a little bit towards a zebra, but to a lesser extent. Finally, the color concept does not seem to play a large role, but more colorful images seem to decrease the likelihood of a zebra.

Based on this information, you can aim to detect whether the model is identifying these different concepts that (on average) push it to predict ``zebra''. For instance, an image that depicts stripes clearly has a higher chance of being predicted as ``zebra'' compared to an image where stripes are not visible.

We note that there might be other concepts that we are not considering here. This is a limitation of the technique.''

\begin{figure}[!t]
\centering
\begin{subfigure}[t]{0.05\textwidth}
\textbf{a}
\end{subfigure}
\begin{subfigure}[t]{0.35\textwidth}
\includegraphics[width=\textwidth,valign=t]{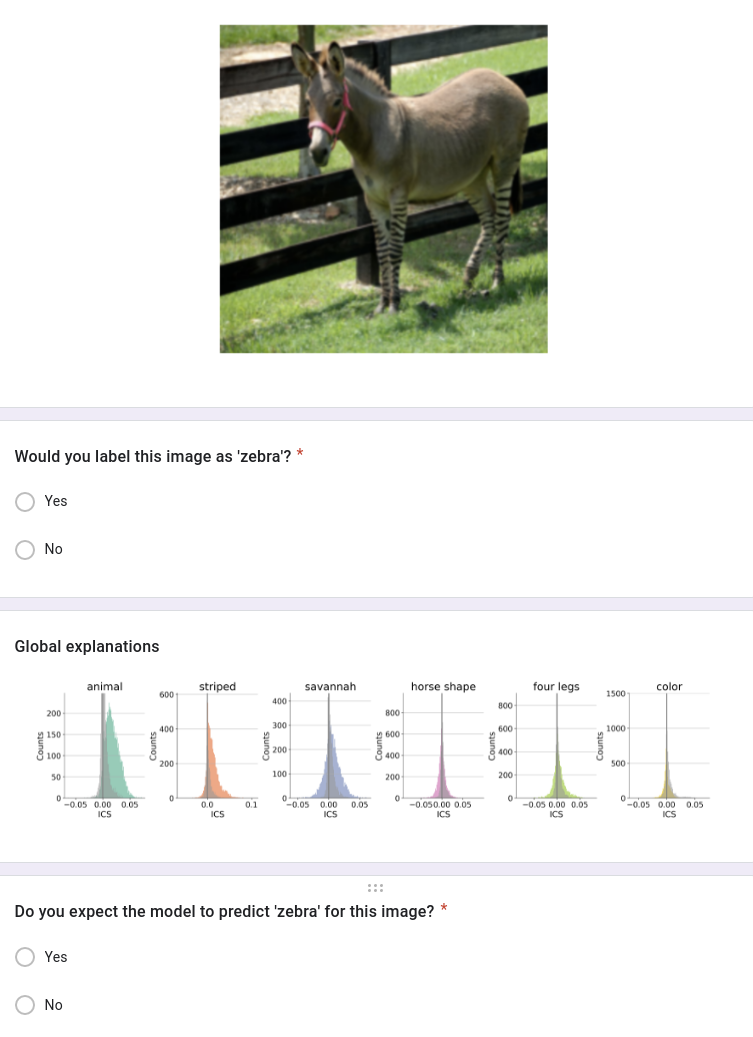}
\end{subfigure}
\begin{subfigure}[t]{0.05\textwidth}
\textbf{b}
\end{subfigure}
\begin{subfigure}[t]{0.35\textwidth}
\includegraphics[width=\textwidth,valign=t]{figures/User_study_supp/all_images/Question_Group3.png}
\end{subfigure}
\caption{Example of question for Group 2 (a) and Group 3 (b).}
\label{app:fig_user_study_group2}
\end{figure}

Each image was then displayed with the global explanations along the 2 questions, as illustrated in Figure~\ref{app:fig_user_study_group2}a.

\subsubsection{Group 3: local and global explanations}
Participants received the explanations described in group 2, and also received a description of local explanations using ICS:

``You have been assigned to group 3. This means that you will be presented with the images to make your assessment, and that we provide both global and local model explanations to help you in your task. Global explanations highlight, on average, what features the model relies on to predict a certain class. Local explanations give, for a specific image, which features were relevant.
[\dots]

\noindent\textbf{Local explanations:} Each image will have a similar graph (as Figure \ref{fig:user_study_global}), that will depict the score for each concept (colored line), compared to random concepts (grey bars). Typically, if a score is 0 and similar to scores for random concepts (i.e. the bar overlays the grey distribution), the model does not identify or rely on the concept in the image.

Importantly, not all concepts need to be significant for the image to be classified as ``zebra'', as this depends on the global importance of the concept, and on whether there is a ``more suited'' class for the image.

For cases that are difficult, it can be interesting to look at all concepts, knowing that a significant negative score suggests a decreased likelihood of zebra, and a positive score means a higher likelihood.

Please note that all scores are quite small, usually $<0.1$.''

\begin{figure}[!t]
\centering
\begin{subfigure}[t]{0.14\textwidth}
\textbf{Image}
\end{subfigure}
\begin{subfigure}[t]{0.04\textwidth}
\textbf{Pred.}
\end{subfigure}
\hspace{0.01\textwidth}
\begin{subfigure}[t]{0.18\textwidth}
\textbf{Global exp.}
\end{subfigure}
\begin{subfigure}[t]{0.6\textwidth}
\textbf{Local exp.}
\end{subfigure} \\
\vspace{0.01\textwidth}
\begin{subfigure}[t]{0.14\textwidth}
\includegraphics[width=\linewidth,valign=t]{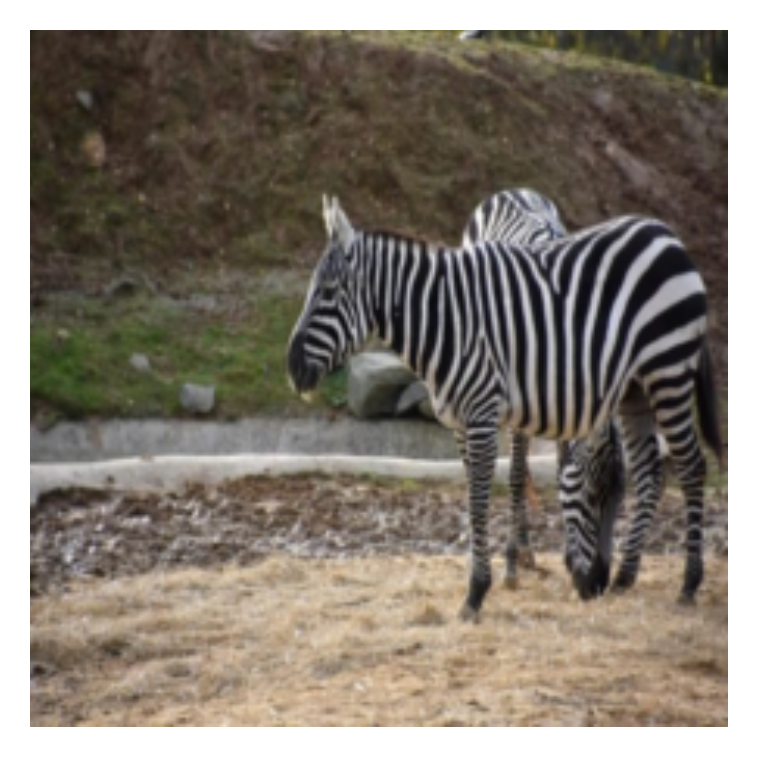} 
\end{subfigure}
\begin{subfigure}[t]{0.04\textwidth}
Zebra
\end{subfigure}
\hspace{0.01\textwidth}
\begin{subfigure}[t]{0.18\textwidth}
We clearly see stripes and guess that the shape of the horse and four legs are identified by the model.
\end{subfigure}
\begin{subfigure}[t]{0.58\textwidth}
\includegraphics[width=\linewidth,valign=t]{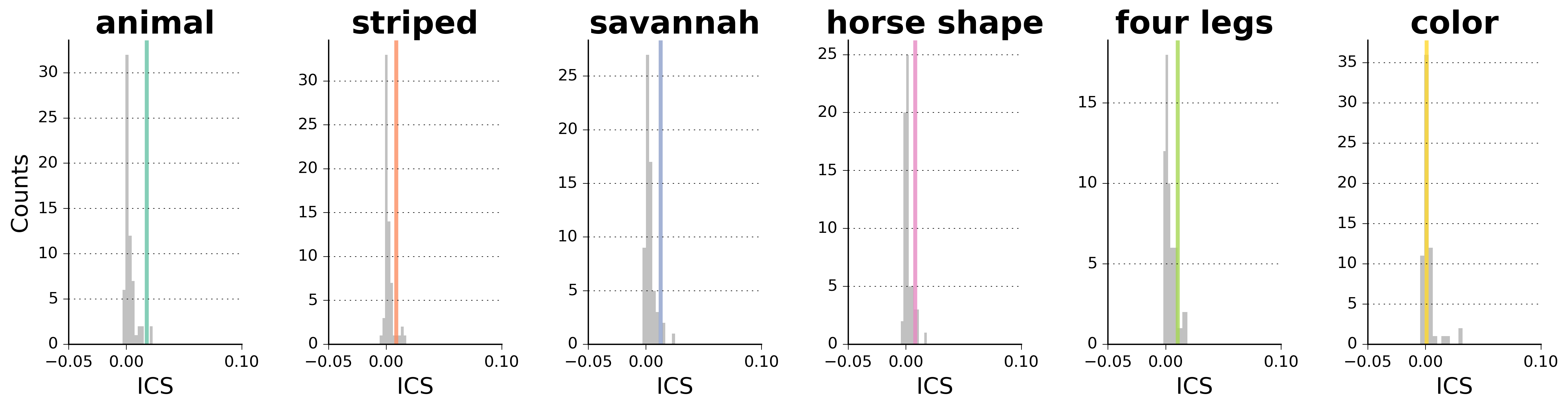} 
\end{subfigure} \\
\vspace{0.02\textwidth}
\begin{subfigure}[t]{0.14\textwidth}
\includegraphics[width=\linewidth,valign=t]{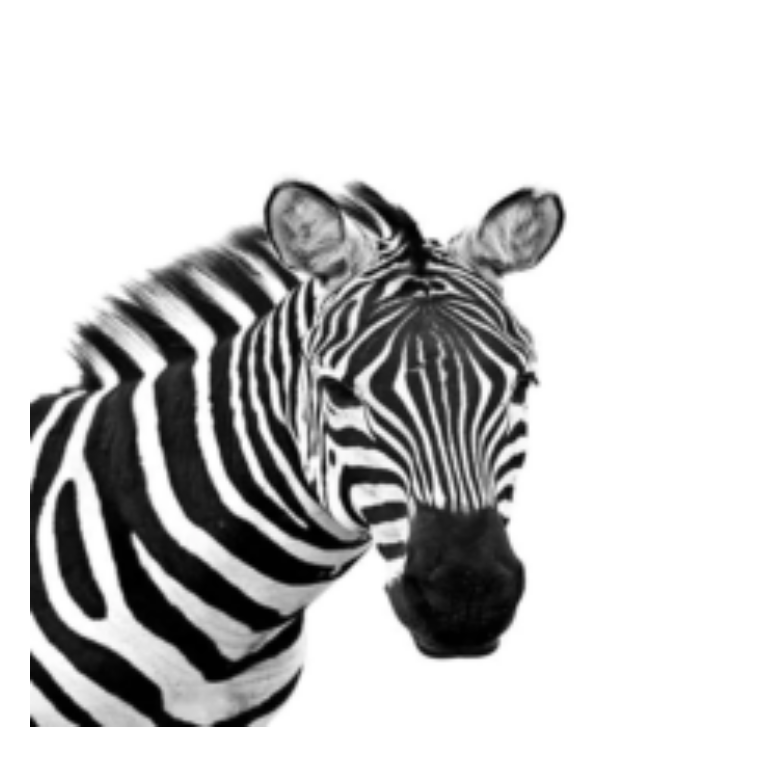} 
\end{subfigure}
\begin{subfigure}[t]{0.04\textwidth}
Zebra
\end{subfigure}
\hspace{0.01\textwidth}
\begin{subfigure}[t]{0.18\textwidth}
We can't see the full zebra or any background. However, the stripes are clearly visible and we know this is one of the most important concepts.
\end{subfigure}
\begin{subfigure}[t]{0.58\textwidth}
\includegraphics[width=\linewidth,valign=t]{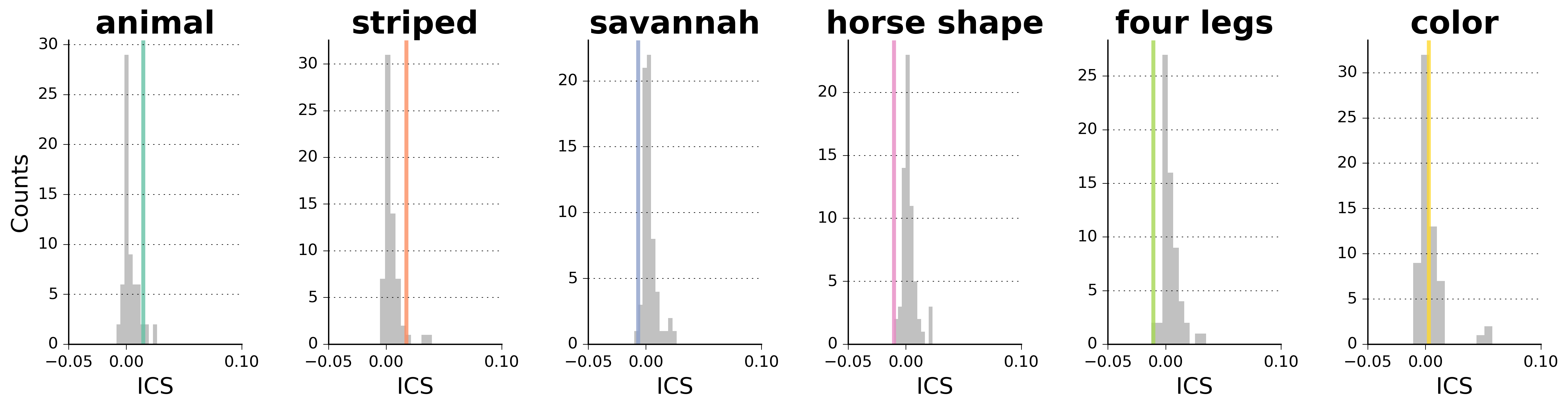} 
\end{subfigure} \\
\vspace{0.02\textwidth}
\begin{subfigure}[t]{0.14\textwidth}
\includegraphics[width=\linewidth,valign=t]{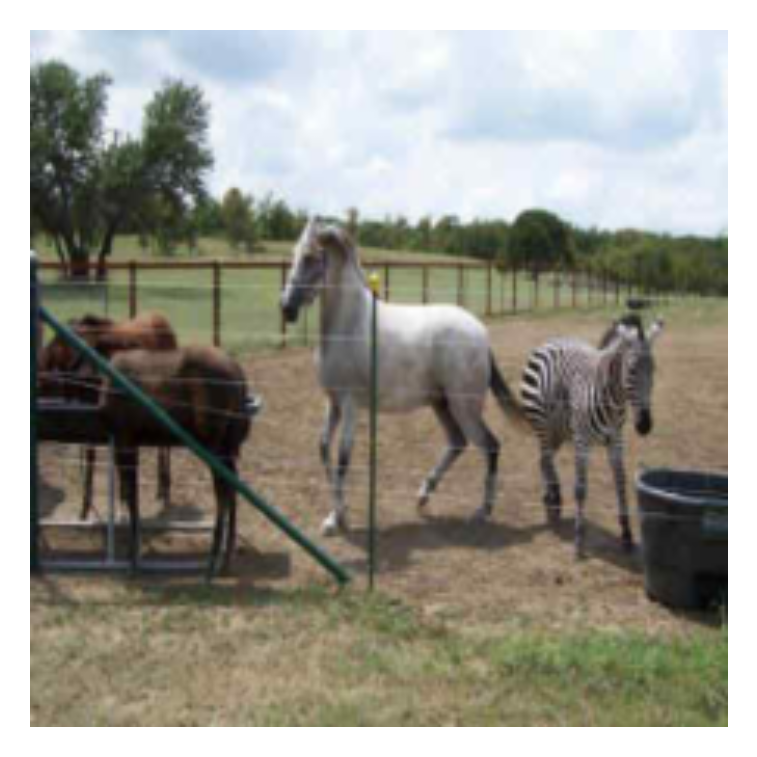} 
\end{subfigure}
\begin{subfigure}[t]{0.04\textwidth}
Hog
\end{subfigure}
\hspace{0.01\textwidth}
\begin{subfigure}[t]{0.18\textwidth}
There are multiple animals. The stripes are visible, as is the full animal shape, but not as the main element.
\end{subfigure}
\begin{subfigure}[t]{0.58\textwidth}
\includegraphics[width=\linewidth,valign=t]{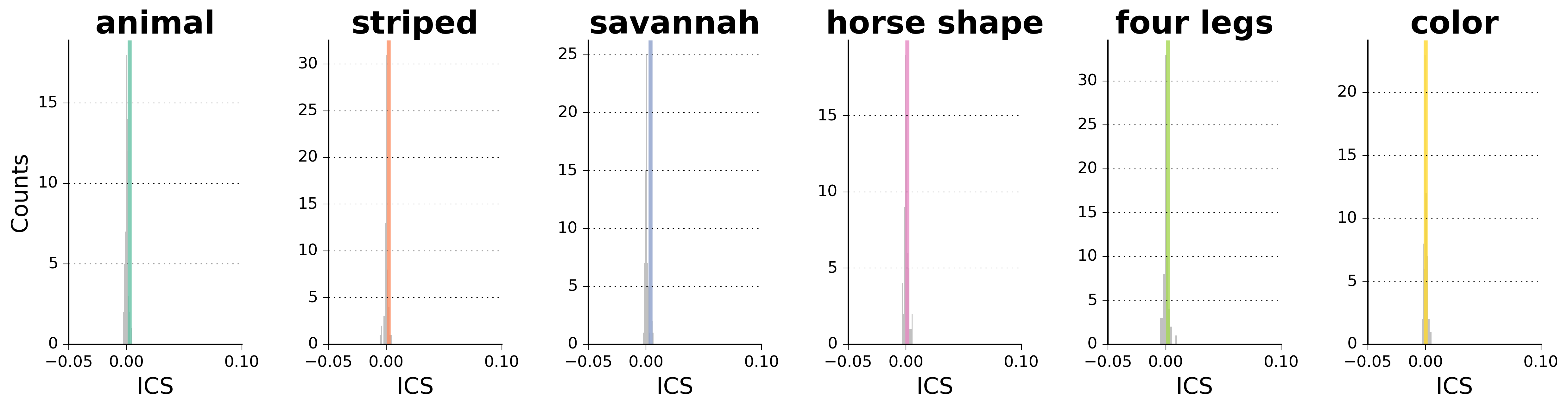} 
\end{subfigure} \\
\caption{Examples presented in the form: Group 1 sees the image and label, Group 2 also has the global explanations and a short text illustrating how to use the information, Group 3 also see the local explanations with an amended text. `Pred.' stands for the model prediction and `Global exp.' for global explanations.}
\label{fig:user_study_groups}
\end{figure}

To illustrate these local explanations, we present 3 images with their associated local explanations and predictions below (Figure \ref{fig:user_study_groups}, right column).

Each image was then displayed with the global and the local explanations along the 2 questions, as illustrated in Figure~\ref{app:fig_user_study_group2}b.

%% file: ICS_Schrouff_arxiv.bbl
\newcommand{\etalchar}[1]{$^{#1}$}
\begin{thebibliography}{COPA{\etalchar{+}}19}

\bibitem[AC{\"{O}}G18]{Ancona2018}
Marco Ancona, Enea Ceolini, Cengiz {\"{O}}ztireli, and Markus Gross.
\newblock {Towards better understanding of gradient-based attribution methods
  for Deep Neural Networks}.
\newblock In {\em Proceedings of the 2018 International Conference on Learning
  Representations (ICLR)}, 2018.

\bibitem[AGM{\etalchar{+}}18]{Adebayo2018}
Julius Adebayo, Justin Gilmer, Michael Muelly, Ian Goodfellow, Moritz Hardt,
  and Been Kim.
\newblock {Sanity checks for saliency maps}.
\newblock In {\em Advances in Neural Information Processing Systems}, volume
  2018-Decem, pages 9505--9515, 2018.

\bibitem[AJ18]{Alvarez2018}
David Alvarez{-}Melis and Tommi~S. Jaakkola.
\newblock On the robustness of interpretability methods.
\newblock {\em CoRR}, abs/1806.08049, 2018.

\bibitem[ALSA{\etalchar{+}}17]{Angelino2017}
Elaine Angelino, Nicholas Larus-Stone, Daniel Alabi, Margo Seltzer, and Cynthia
  Rudin.
\newblock Learning certifiably optimal rule lists for categorical data.
\newblock {\em J. Mach. Learn. Res.}, 18(1):8753–8830, January 2017.

\bibitem[A{\"{O}}G19]{Ancona2019}
Marco Ancona, Cengiz {\"{O}}ztireli, and Markus Gross.
\newblock {Explaining Deep Neural Networks with a Polynomial Time Algorithm for
  Shapley Values Approximation}.
\newblock In {\em Proceedings of the 36th International Conference on Machine
  Learning}, 2019.

\bibitem[AP08]{Artstein2008}
Ron Artstein and Massimo Poesio.
\newblock Inter-coder agreement for computational linguistics.
\newblock {\em Comput. Linguist.}, 34(4):555–596, dec 2008.

\bibitem[BBM{\etalchar{+}}15]{Bach2015}
Sebastian Bach, Alexander Binder, Gr{\'{e}}goire Montavon, Frederick Klauschen,
  Klaus~Robert M{\"{u}}ller, and Wojciech Samek.
\newblock {On pixel-wise explanations for non-linear classifier decisions by
  layer-wise relevance propagation}.
\newblock {\em PLoS ONE}, 10(7), jul 2015.

\bibitem[BCB16]{bahdanau2016neural}
Dzmitry Bahdanau, Kyunghyun Cho, and Yoshua Bengio.
\newblock Neural machine translation by jointly learning to align and
  translate, 2016.

\bibitem[BH20]{debiasingConcept2020}
Mohammad~Taha Bahadori and David~E. Heckerman.
\newblock Debiasing concept bottleneck models with instrumental variables.
\newblock {\em CoRR}, abs/2007.11500, 2020.

\bibitem[BOGO15]{Briol2015}
Fran\c{c}ois-Xavier Briol, Chris Oates, Mark Girolami, and Michael~A Osborne.
\newblock Frank-wolfe bayesian quadrature: Probabilistic integration with
  theoretical guarantees.
\newblock In C.~Cortes, N.~D. Lawrence, D.~D. Lee, M.~Sugiyama, and R.~Garnett,
  editors, {\em Advances in Neural Information Processing Systems 28}, pages
  1162--1170. Curran Associates, Inc., 2015.

\bibitem[BXS{\etalchar{+}}20]{Bhatt2020-mz}
Umang Bhatt, Alice Xiang, Shubham Sharma, Adrian Weller, Ankur Taly, Yunhan
  Jia, Joydeep Ghosh, Ruchir Puri, Jos{\'e} M~F Moura, and Peter Eckersley.
\newblock Explainable machine learning in deployment.
\newblock In {\em Proceedings of the 2020 Conference on Fairness,
  Accountability, and Transparency}, FAT* '20, pages 648--657, New York, NY,
  USA, January 2020. Association for Computing Machinery.

\bibitem[BZK{\etalchar{+}}17]{bau2017network}
David Bau, Bolei Zhou, Aditya Khosla, Aude Oliva, and Antonio Torralba.
\newblock Network dissection: Quantifying interpretability of deep visual
  representations.
\newblock In {\em Proceedings of the IEEE conference on computer vision and
  pattern recognition}, pages 6541--6549, 2017.

\bibitem[CDH{\etalchar{+}}20]{tfmodels2020github}
Chen Chen, Xianzhi Du, Le~Hou, Jaeyoun Kim, Jing Li, Yeqing Li, Abdullah
  Rashwan, Fan Yang, and Hongkun Yu.
\newblock Tensorflow official model garden, 2020.

\bibitem[CLT{\etalchar{+}}19]{Chen2019}
Chaofan Chen, Oscar Li, Daniel Tao, Alina Barnett, Cynthia Rudin, and
  Jonathan~K Su.
\newblock This looks like that: Deep learning for interpretable image
  recognition.
\newblock In H.~Wallach, H.~Larochelle, A.~Beygelzimer, F.~d'Alch\'{e} Buc,
  E.~Fox, and R.~Garnett, editors, {\em Advances in Neural Information
  Processing Systems}, volume~32. Curran Associates, Inc., 2019.

\bibitem[Coh60]{cohen1960}
Jacob Cohen.
\newblock A coefficient of agreement for nominal scales.
\newblock {\em Educational and Psychological Measurement}, 20(1):37--46, 1960.

\bibitem[COPA{\etalchar{+}}19]{Clough2019}
{James R.} Clough, Ilkay Oksuz, Esther Puyol-Ant{\'o}n, Bram Ruijsink, {Andrew
  P.} King, and {Julia A.} Schnabel.
\newblock {\em Global and local interpretability for cardiac MRI
  classification}, pages 656--664.
\newblock Lecture Notes in Computer Science (including subseries Lecture Notes
  in Artificial Intelligence and Lecture Notes in Bioinformatics). SPRINGER,
  January 2019.
\newblock 22nd International Conference on Medical Image Computing and
  Computer-Assisted Intervention, MICCAI 2019 ; Conference date: 13-10-2019
  Through 17-10-2019.

\bibitem[CRH{\etalchar{+}}19]{Cai2019}
Carrie~J Cai, Emily Reif, Narayan Hegde, Jason Hipp, Been Kim, Daniel Smilkov,
  Martin Wattenberg, Fernanda Viegas, Greg~S Corrado, Martin~C Stumpe, and
  Michael Terry.
\newblock {Human-centered tools for coping with imperfect algorithms during
  medical decision-making}.
\newblock In {\em Conference on Human Factors in Computing Systems -
  Proceedings}, page~14. ACM, 2019.

\bibitem[DAA{\etalchar{+}}19]{Dombrowski2019}
Ann-Kathrin Dombrowski, Maximilian Alber, Christopher~J. Anders, Marcel
  Ackermann, Klaus-Robert Müller, and Pan Kessel.
\newblock Explanations can be manipulated and geometry is to blame, 2019.

\bibitem[DDS{\etalchar{+}}09]{Deng09}
J.~Deng, W.~Dong, R.~Socher, L.-J. Li, K.~Li, and L.~Fei-Fei.
\newblock {ImageNet: A Large-Scale Hierarchical Image Database}.
\newblock In {\em CVPR09}, 2009.

\bibitem[DV17]{DoshiKim2017Interpretability}
Been Doshi-Velez, Finale;~Kim.
\newblock Towards a rigorous science of interpretable machine learning.
\newblock In {\em eprint arXiv:1702.08608}, 2017.

\bibitem[DVK18]{Doshi-Velez2018}
Finale Doshi-Velez and Been Kim.
\newblock {\em Considerations for Evaluation and Generalization in
  Interpretable Machine Learning}, pages 3--17.
\newblock Springer International Publishing, Cham, 2018.

\bibitem[EEG{\etalchar{+}}18]{Escalante2018}
Hugo~Jair Escalante, Sergio Escalera, Isabelle Guyon, Xavier Bar{\'{o}},
  Yağmur G{\"{u}}{\c{c}}l{\"{u}}t{\"{u}}rk, Umut G{\"{u}}{\c{c}}l{\"{u}}, and
  Marcel van Gerven.
\newblock {\em {Explainable and Interpretable Models in Computer Vision and
  Machine Learning}}.
\newblock Springer, 2018.

\bibitem[EJS{\etalchar{+}}20]{Erion2020}
Gabriel Erion, Joseph~D. Janizek, Pascal Sturmfels, Scott Lundberg, and Su-In
  Lee.
\newblock Improving performance of deep learning models with axiomatic
  attribution priors and expected gradients, 2020.

\bibitem[FCCS21]{Fel2021-ce}
Thomas Fel, Julien Colin, Remi Cadene, and Thomas Serre.
\newblock What {I} cannot predict, {I} do not understand: A {Human-Centered}
  evaluation framework for explainability methods.
\newblock December 2021.

\bibitem[GAM18]{Graziani2018}
Mara Graziani, Vincent Andrearczyk, and Henning M{\"u}ller.
\newblock Regression concept vectors for bidirectional explanations in
  histopathology.
\newblock In Danail Stoyanov, Zeike Taylor, Seyed~Mostafa Kia, Ipek Oguz,
  Mauricio Reyes, Anne Martel, Lena Maier-Hein, Andre~F. Marquand, Edouard
  Duchesnay, Tommy L{\"o}fstedt, Bennett Landman, M.~Jorge Cardoso, Carlos~A.
  Silva, Sergio Pereira, and Raphael Meier, editors, {\em Understanding and
  Interpreting Machine Learning in Medical Image Computing Applications}, pages
  124--132, Cham, 2018. Springer International Publishing.

\bibitem[GAZ17]{Ghorbani2017}
Amirata Ghorbani, Abubakar Abid, and James Zou.
\newblock Interpretation of neural networks is fragile, 2017.

\bibitem[GAZ19]{Ghorbani2019}
Amirata Ghorbani, Abubakar Abid, and James Zou.
\newblock {Interpretation of Neural Networks Is Fragile}.
\newblock {\em Proceedings of the AAAI Conference on Artificial Intelligence},
  33:3681--3688, jul 2019.

\bibitem[GLW{\etalchar{+}}20]{Goh2020}
Gary S.~W. Goh, Sebastian Lapuschkin, Leander Weber, Wojciech Samek, and
  Alexander Binder.
\newblock Understanding integrated gradients with smoothtaylor for deep neural
  network attribution.
\newblock {\em CoRR}, abs/2004.10484, 2020.

\bibitem[GORB21]{Ghassemi2021-ic}
Marzyeh Ghassemi, Luke Oakden-Rayner, and Andrew~L Beam.
\newblock The false hope of current approaches to explainable artificial
  intelligence in health care.
\newblock {\em Lancet Digit Health}, 3(11):e745--e750, November 2021.

\bibitem[GSK19]{Goyal2019}
Yash Goyal, Uri Shalit, and Been Kim.
\newblock Explaining classifiers with causal concept effect (cace).
\newblock {\em CoRR}, abs/1907.07165, 2019.

\bibitem[HPRPC20]{Hohman2020}
Fred Hohman, Haekyu Park, Caleb Robinson, and Duen~Horng Polo~Chau.
\newblock Summit: Scaling deep learning interpretability by visualizing
  activation and attribution summarizations.
\newblock {\em IEEE Transactions on Visualization and Computer Graphics},
  26(1):1096--1106, 2020.

\bibitem[HZRS15]{He2015}
Kaiming He, Xiangyu Zhang, Shaoqing Ren, and Jian Sun.
\newblock Deep residual learning for image recognition.
\newblock {\em CoRR}, abs/1512.03385, 2015.

\bibitem[ILMP19]{Ibrahim2019-hh}
Mark Ibrahim, Melissa Louie, Ceena Modarres, and John Paisley.
\newblock Global explanations of neural networks: Mapping the landscape of
  predictions.
\newblock In {\em Proceedings of the 2019 {AAAI/ACM} Conference on {AI},
  Ethics, and Society}, AIES '19, pages 279--287, New York, NY, USA, January
  2019. Association for Computing Machinery.

\bibitem[JW19]{jain2019attention}
Sarthak Jain and Byron~C Wallace.
\newblock Attention is not explanation.
\newblock In {\em Conference of the North American Chapter of the Association
  for Computational Linguistics: Human Language Technologies}, volume~1, pages
  3543--3556, 2019.

\bibitem[KB14]{Kingma2014}
Diederik~P. Kingma and Jimmy Ba.
\newblock Adam: A method for stochastic optimization, 2014.
\newblock cite arxiv:1412.6980Comment: Published as a conference paper at the
  3rd International Conference for Learning Representations, San Diego, 2015.

\bibitem[KBVT19]{Kapishnikov_2019_ICCV}
Andrei Kapishnikov, Tolga Bolukbasi, Fernanda Viegas, and Michael Terry.
\newblock Xrai: Better attributions through regions.
\newblock In {\em Proceedings of the IEEE/CVF International Conference on
  Computer Vision (ICCV)}, October 2019.

\bibitem[KHA{\etalchar{+}}19]{Kindermans2017}
Pieter-Jan Kindermans, Sara Hooker, Julius Adebayo, Maximilian Alber,
  Kristof~T. Sch{\"u}tt, Sven D{\"a}hne, Dumitru Erhan, and Been Kim.
\newblock {\em The (Un)reliability of Saliency Methods}, pages 267--280.
\newblock Springer International Publishing, Cham, 2019.

\bibitem[KNT{\etalchar{+}}20]{Koh2020_CBM}
Pang~Wei Koh, Thao Nguyen, Yew~Siang Tang, Stephen Mussmann, Emma Pierson, Been
  Kim, and Percy Liang.
\newblock Concept bottleneck models.
\newblock In Hal~Daumé III and Aarti Singh, editors, {\em Proceedings of the
  37th International Conference on Machine Learning}, volume 119 of {\em
  Proceedings of Machine Learning Research}, pages 5338--5348. PMLR, 13--18 Jul
  2020.

\bibitem[KVA{\etalchar{+}}21]{Kapishnikov2021-aa}
Andrei Kapishnikov, Subhashini Venugopalan, Besim Avci, Ben Wedin, Michael
  Terry, and Tolga Bolukbasi.
\newblock Guided integrated gradients: An adaptive path method for removing
  noise.
\newblock June 2021.

\bibitem[KWG{\etalchar{+}}18]{Kim2017}
Been Kim, Martin Wattenberg, Justin Gilmer, Carrie Cai, James Wexler, Fernanda
  Viegas, and Rory Sayres.
\newblock {Interpretability beyond feature attribution: Quantitative Testing
  with Concept Activation Vectors (TCAV)}.
\newblock In {\em 35th International Conference on Machine Learning, ICML
  2018}, volume~6, pages 4186--4195, 2018.

\bibitem[LEC{\etalchar{+}}19]{Lundberg19}
Scott~M. Lundberg, Gabriel~G. Erion, Hugh Chen, Alex DeGrave, Jordan~M.
  Prutkin, Bala Nair, Ronit Katz, Jonathan Himmelfarb, Nisha Bansal, and
  Su{-}In Lee.
\newblock Explainable {AI} for trees: From local explanations to global
  understanding.
\newblock {\em CoRR}, abs/1905.04610, 2019.

\bibitem[LHK19]{Linden2019GlobalAO}
I.~V.~D. Linden, H.~Haned, and E.~Kanoulas.
\newblock Global aggregations of local explanations for black box models.
\newblock {\em ArXiv}, abs/1907.03039, 2019.

\bibitem[Lip18]{Lipton2016}
Zachary~C. Lipton.
\newblock The mythos of model interpretability: In machine learning, the
  concept of interpretability is both important and slippery.
\newblock {\em Queue}, 16(3):31–57, June 2018.

\bibitem[LJAJ19]{Lee2019}
Guang{-}He Lee, Wengong Jin, David Alvarez{-}Melis, and Tommi~S. Jaakkola.
\newblock Functional transparency for structured data: a game-theoretic
  approach.
\newblock In Kamalika Chaudhuri and Ruslan Salakhutdinov, editors, {\em
  Proceedings of the 36th International Conference on Machine Learning, {ICML}
  2019, 9-15 June 2019, Long Beach, California, {USA}}, volume~97 of {\em
  Proceedings of Machine Learning Research}, pages 3723--3733. {PMLR}, 2019.

\bibitem[LL17]{Lundberg2017_SHAP}
Scott~M Lundberg and Su-In Lee.
\newblock A unified approach to interpreting model predictions.
\newblock In I.~Guyon, U.~V. Luxburg, S.~Bengio, H.~Wallach, R.~Fergus,
  S.~Vishwanathan, and R.~Garnett, editors, {\em Advances in Neural Information
  Processing Systems 30}, pages 4765--4774. Curran Associates, Inc., 2017.

\bibitem[LMB{\etalchar{+}}14]{Lin2014}
Tsung{-}Yi Lin, Michael Maire, Serge~J. Belongie, Lubomir~D. Bourdev, Ross~B.
  Girshick, James Hays, Pietro Perona, Deva Ramanan, Piotr Doll{\'{a}}r, and
  C.~Lawrence Zitnick.
\newblock Microsoft {COCO:} common objects in context.
\newblock {\em CoRR}, abs/1405.0312, 2014.

\bibitem[MLH{\etalchar{+}}21]{Mincu2021}
Diana Mincu, Eric Loreaux, Shaobo Hou, Sebastien Baur, Ivan Protsyuk, Martin
  Seneviratne, Anne Mottram, Nenad Tomasev, Alan Karthikesalingam, and Jessica
  Schrouff.
\newblock {\em Concept-Based Model Explanations for Electronic Health Records},
  page 36–46.
\newblock Association for Computing Machinery, New York, NY, USA, 2021.

\bibitem[MWZ{\etalchar{+}}19]{Mitchell2019-ma}
Margaret Mitchell, Simone Wu, Andrew Zaldivar, Parker Barnes, Lucy Vasserman,
  Ben Hutchinson, Elena Spitzer, Inioluwa~Deborah Raji, and Timnit Gebru.
\newblock Model cards for model reporting.
\newblock In {\em Proceedings of the Conference on Fairness, Accountability,
  and Transparency}, FAT* '19, pages 220--229, New York, NY, USA, January 2019.
  Association for Computing Machinery.

\bibitem[PPPP20]{Panigutti2020}
Cecilia Panigutti, Alan Perotti, Dino Pedreschi, and Dino~2020 Pedreschi.
\newblock {An ontology-based approach to black-box sequential data
  classification explanations}.
\newblock 2020.

\bibitem[RMP{\etalchar{+}}20]{Reyes2020-ym}
Mauricio Reyes, Raphael Meier, S{\'e}rgio Pereira, Carlos~A Silva,
  Fried-Michael Dahlweid, Hendrik von Tengg-Kobligk, Ronald~M Summers, and
  Roland Wiest.
\newblock On the interpretability of artificial intelligence in radiology:
  Challenges and opportunities.
\newblock {\em Radiol Artif Intell}, 2(3):e190043, May 2020.

\bibitem[RSG16]{Ribeiro2016}
Marco~Tulio Ribeiro, Sameer Singh, and Carlos Guestrin.
\newblock {"Why should i trust you?" Explaining the predictions of any
  classifier}.
\newblock In {\em Proceedings of the ACM SIGKDD International Conference on
  Knowledge Discovery and Data Mining}, volume 13-17-Augu, pages 1135--1144,
  2016.

\bibitem[SGK17]{Shrikumar2017}
Avanti Shrikumar, Peyton Greenside, and Anshul Kundaje.
\newblock {Learning important features through propagating activation
  differences}.
\newblock In {\em 34th International Conference on Machine Learning, ICML
  2017}, volume~7, pages 4844--4866. International Machine Learning Society
  (IMLS), apr 2017.

\bibitem[SGL19]{sixt2019explanations}
Leon Sixt, Maximilian Granz, and Tim Landgraf.
\newblock When explanations lie: Why modified bp attribution fails.
\newblock {\em arXiv preprint arXiv:1912.09818}, 2019.

\bibitem[SGSK16]{Shrikumar2016}
Avanti Shrikumar, Peyton Greenside, Anna Shcherbina, and Anshul Kundaje.
\newblock {Not Just a Black Box: Learning Important Features Through
  Propagating Activation Differences}.
\newblock {\em arXiv}, 1:0--5, may 2016.

\bibitem[SHZ{\etalchar{+}}18]{Sandler2018}
Mark Sandler, Andrew~G. Howard, Menglong Zhu, Andrey Zhmoginov, and
  Liang{-}Chieh Chen.
\newblock Inverted residuals and linear bottlenecks: Mobile networks for
  classification, detection and segmentation.
\newblock {\em CoRR}, abs/1801.04381, 2018.

\bibitem[SLJ{\etalchar{+}}15]{Szegedy2015}
Christian Szegedy, Wei Liu, Yangqing Jia, Pierre Sermanet, Scott Reed, Dragomir
  Anguelov, Dumitru Erhan, Vincent Vanhoucke, and Andrew Rabinovich.
\newblock Going deeper with convolutions.
\newblock In {\em Computer Vision and Pattern Recognition (CVPR)}, 2015.

\bibitem[SLL20]{Sturmfels2020}
Pascal Sturmfels, Scott Lundberg, and Su-In Lee.
\newblock {Visualizing the Impact of Feature Attribution Baselines}.
\newblock {\em Distill}, 5(1), jan 2020.

\bibitem[STK{\etalchar{+}}17]{Smilkov2017-vc}
Daniel Smilkov, Nikhil Thorat, Been Kim, Fernanda Vi{\'e}gas, and Martin
  Wattenberg.
\newblock {SmoothGrad}: removing noise by adding noise.
\newblock June 2017.

\bibitem[STR{\etalchar{+}}19]{Sayres2019}
Rory Sayres, Ankur Taly, Ehsan Rahimy, Katy Blumer, David Coz, Naama Hammel,
  Jonathan Krause, Arunachalam Narayanaswamy, Zahra Rastegar, Derek Wu, Shawn
  Xu, Scott Barb, Anthony Joseph, Michael Shumski, Jesse Smith, Arjun~B Sood,
  Greg~S Corrado, Lily Peng, and Dale~R Webster.
\newblock {Using a Deep Learning Algorithm and Integrated Gradients Explanation
  to Assist Grading for Diabetic Retinopathy}.
\newblock {\em Ophthalmology}, 126(4):552--564, 2019.

\bibitem[STY17]{Sundararajan2017}
Mukund Sundararajan, Ankur Taly, and Qiqi Yan.
\newblock {Axiomatic attribution for deep networks}.
\newblock In {\em 34th International Conference on Machine Learning, ICML
  2017}, volume~7, pages 5109--5118, 2017.

\bibitem[SVZ14]{Simonyan2013}
Karen Simonyan, Andrea Vedaldi, and Andrew Zisserman.
\newblock Deep inside convolutional networks: Visualising image classification
  models and saliency maps.
\newblock In {\em Workshop at International Conference on Learning
  Representations}, 2014.

\bibitem[TJMG19]{Tonekaboni2019}
Sana Tonekaboni, Shalmali Joshi, Melissa~D McCradden, and Anna Goldenberg.
\newblock {What Clinicians Want: Contextualizing Explainable Machine Learning
  for Clinical End Use}.
\newblock In {\em Proceedings of Machine Learning Research}, pages 1 -- 21,
  2019.

\bibitem[TL20]{Tan2020}
Mingxing Tan and Quoc~V. Le.
\newblock Efficientnet: Rethinking model scaling for convolutional neural
  networks, 2020.

\bibitem[UCS17]{Ubaru2017}
Shashanka Ubaru, Jie Chen, and Yousef Saad.
\newblock Fast estimation of tr(f(a)) via stochastic lanczos quadrature.
\newblock {\em SIAM Journal on Matrix Analysis and Applications},
  38(4):1075--1099, January 2017.

\bibitem[vdLHK19]{Van_der_Linden2019-vf}
Ilse van~der Linden, Hinda Haned, and Evangelos Kanoulas.
\newblock Global aggregations of local explanations for black box models.
\newblock July 2019.

\bibitem[WSC{\etalchar{+}}20]{Wu2020}
Weibin Wu, Yuxin Su, Xixian Chen, Shenglin Zhao, Irwin King, Michael~R. Lyu,
  and Yu-Wing Tai.
\newblock Towards global explanations of convolutional neural networks with
  concept attribution.
\newblock In {\em 2020 IEEE/CVF Conference on Computer Vision and Pattern
  Recognition (CVPR)}, pages 8649--8658, 2020.

\bibitem[YHS{\etalchar{+}}19]{Chih-Kuan2019}
Chih-Kuan Yeh, Cheng-Yu Hsieh, Arun Suggala, David~I Inouye, and Pradeep~K
  Ravikumar.
\newblock On the (in)fidelity and sensitivity of explanations.
\newblock In H.~Wallach, H.~Larochelle, A.~Beygelzimer, F.~d'Alch\'{e} Buc,
  E.~Fox, and R.~Garnett, editors, {\em Advances in Neural Information
  Processing Systems 32}, pages 10967--10978. Curran Associates, Inc., 2019.

\bibitem[YK19]{Yang2019}
Mengjiao Yang and Been Kim.
\newblock {BIM:} towards quantitative evaluation of interpretability methods
  with ground truth.
\newblock {\em CoRR}, abs/1907.09701, 2019.

\bibitem[YKA{\etalchar{+}}20]{Yeh2020-xq}
Chih-Kuan Yeh, Been Kim, Sercan Arik, Chun-Liang Li, Tomas Pfister, and Pradeep
  Ravikumar.
\newblock On completeness-aware {Concept-Based} explanations in deep neural
  networks.
\newblock {\em Adv. Neural Inf. Process. Syst.}, 33:20554--20565, 2020.

\bibitem[ZF14]{Zeiler2014}
Matthew~D. Zeiler and Rob Fergus.
\newblock Visualizing and understanding convolutional networks.
\newblock In David Fleet, Tomas Pajdla, Bernt Schiele, and Tinne Tuytelaars,
  editors, {\em Computer Vision -- ECCV 2014}, pages 818--833, Cham, 2014.
  Springer International Publishing.

\bibitem[ZH05]{Zou05}
Hui Zou and Trevor Hastie.
\newblock Regularization and variable selection via the elastic net.
\newblock {\em Journal of the Royal Statistical Society, Series B},
  67:301--320, 2005.

\bibitem[ZLK{\etalchar{+}}18]{Zhou2018}
B.~{Zhou}, A.~{Lapedriza}, A.~{Khosla}, A.~{Oliva}, and A.~{Torralba}.
\newblock Places: A 10 million image database for scene recognition.
\newblock {\em IEEE Transactions on Pattern Analysis and Machine Intelligence},
  40(6):1452--1464, 2018.

\end{thebibliography}
